\documentclass[10pt,twocolumn,letterpaper]{article}

\usepackage{cvpr}

\usepackage{colortbl}
\colorlet{colorFst}{Green!25}
\colorlet{colorSnd}{SpringGreen!45}
\newcommand{\fs}{\cellcolor{colorFst}\bf}
\newcommand{\rd}{\cellcolor{colorSnd}\underline}

\definecolor{cvprblue}{rgb}{0.21,0.49,0.74}
\usepackage[pagebackref,breaklinks,colorlinks,allcolors=cvprblue]{hyperref}
\usepackage{multirow}
\usepackage{graphicx}
\usepackage{booktabs}
\usepackage{tabularx}
\usepackage{adjustbox}
\usepackage[accsupp]{axessibility}

\newcommand\blfootnote[1]{%
  \begingroup
  \renewcommand\thefootnote{}\footnote{#1}%
  \addtocounter{footnote}{-1}%
  \endgroup
}

\def\paperID{8818}
\def\confName{CVPR}
\def\confYear{2025}

\title{Generated Reality: Human-centric World Simulation using Interactive Video Generation with Hand and Camera Control}

\author{Linxi Xie$^{1,2*\dagger}$~~~Lisong C. Sun$^{1*}$~~~Ashley Neall$^{1,3*\dagger}$~~~
Tong Wu$^1$~~~Shengqu Cai$^1$~~~Gordon Wetzstein$^1$\\
\\[-0.3cm]
$\phantom{}^1$Stanford University~~~~~$\phantom{}^2$NYU Shanghai~~~~~$\phantom{}^3$UNC Chapel Hill
\vspace{-1cm}
}

\begin{document}

\twocolumn[{
\maketitle
\centering
\includegraphics[width=0.999\textwidth]{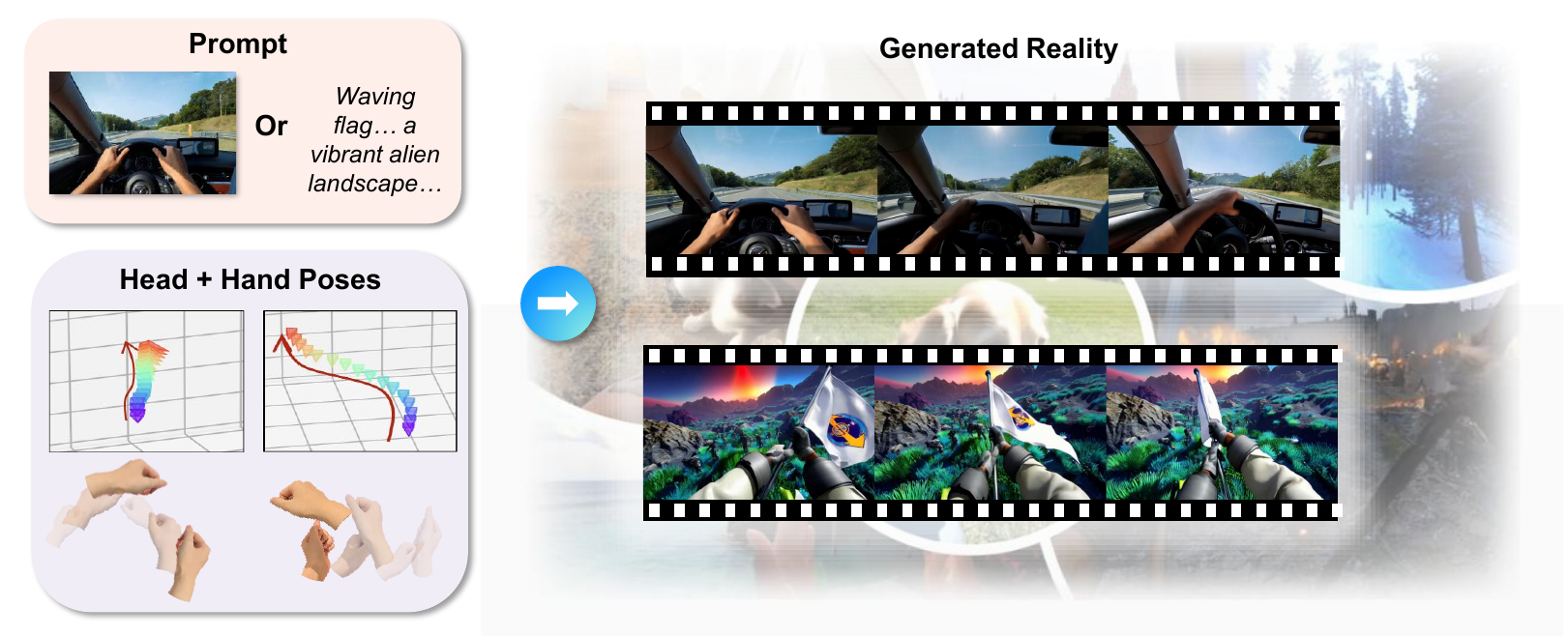}
\captionof{figure}{
Generated reality is a concept that incorporates human-tracked data (left) into an autoregressive video generation model to enable immersive experiences (right). These generated virtual environments do not rely on laboriously designed 3D assets but are created in a zero-shot manner by the video generator. We explore diffusion transformer conditioning strategies for joint-level hand and head poses, identifying a hybrid 2D--3D strategy as the most effective approach. Our bidirectional attention-based video generator is distilled into a few-step autoregressive model, enabling interactive, human-centric experiences supporting dexterous hand--object interactions.}
\label{fig:teaser}
\vspace*{0.5cm}
}]
\blfootnote{${}^*$Equal Contribution.}
\blfootnote{${}^\dagger$Work done as a visiting researcher at Stanford.}

\begin{abstract}
Extended reality (XR) demands generative models that respond to users’ tracked real-world motion, yet current video world models accept only coarse control signals such as text or keyboard input, limiting their utility for embodied interaction. We introduce a human-centric video world model that is conditioned on both tracked head pose and joint-level hand poses. 
For this purpose, we evaluate existing diffusion transformer  conditioning strategies and propose an effective mechanism for 3D head and hand control, enabling dexterous hand--object interactions. 
We train a bidirectional video diffusion model teacher using this strategy and distill it into a causal, interactive system that generates egocentric virtual environments. We evaluate this generated reality system with human subjects and demonstrate improved task performance as well as a significantly higher level of perceived amount of control over the performed actions compared with relevant baselines. 
The project website is at \url{https://codeysun.github.io/generated-reality/}.
\end{abstract}
    
\section{Introduction}
\label{sec:intro}

Extended reality~(XR)---encompassing virtual, augmented, and mixed reality---is crucial in healthcare and rehabilitation, education and professional training, design and engineering, as well as entertainment and media.
Despite its transformative potential across these domains, the creation of XR content remains difficult, laborious, and expensive due to the need for specialized expertise, complex development tools, and high production costs.

Emerging video world models offer a powerful platform to address the challenge of content creation for immersive technologies. These large generative AI models are able to autoregressively generate close-to-photorealistic video at interactive framerates conditioned on actions or other signals~\cite{gameNGen,cosmos,parkerholder2024genie2,decart2024oasis,mineworld}. 

Current video world models, however, remain limited in the types of conditioning signals they accept, often restricted to simple keyboard controls or text prompts~\cite{cosmos,mineworld, animegamer, worldmem, wu2025video, parkerholder2024genie2, yu2025gamefactory}. The limited control makes current world models ineffective as human-centric content generation tools for XR applications. Recent works have focused on conditioning on camera motion~\cite{cameractrlii, AC3D, recammaster} or full-body pose~\cite{PlayerOne, PEVA}, showing promise in modeling interactive egocentric dynamics. However, these approaches lack the precision required to represent the detailed wrist and finger movements involved in dexterous hand--object interactions. As a result, it remains an open question how to effectively incorporate joint-level hand pose conditioning into video diffusion models. Furthermore, it is unclear which conditioning strategies best preserve hand fidelity, realism, and temporal coherence in video generation.

We hypothesize that next-generation world models could support truly embodied interactivity by effectively incorporating rich streams of tracked user data, including head and gaze direction, body pose, foot placement, hand and finger articulation, and full-body movement.
To this end, we develop a human-centric video world model that enables interactive content generation across both existing and yet-unimagined applications, with a focus on effective head and hand control.
Specifically, we present the first systematic study of hand pose conditioning strategies in video diffusion models. We compare several representative approaches, including token concatenation, addition, cross-attention, ControlNet-style conditioning, and adaptive layer normalization, using metrics that evaluate visual quality and hand-pose fidelity. We find that a combination of 2D ControlNet-style conditioning and a 3D joint-level representation of hand poses injected via token addition is the most effective. Finally, we distill our head- and hand-conditioned video generation model into a causal, real-time architecture, achieving 11 frames per second with a latency of 1.4 seconds on a remotely streamed H100. We conduct a user study with this system, demonstrating significantly improved task performance on three different tasks and a substantially larger perceived sense of control by human subjects compared to relevant baselines.

Our vision of generated reality could enable immersive learning, training, and exploration by allowing users to acquire skills, practice complex tasks without detailed models, and experience real or imagined environments in a zero-shot manner. It could support novel interactive media and real-time generative guidance through smart eyewear for diverse applications.

Our key technical contributions include:
\begin{itemize}
    \item We conduct a \textbf{comprehensive ablation} study comparing hand pose conditioning strategies for video diffusion models, identifying a combination of 2D ControlNet-style conditioning and 3D joint conditioning as the most effective strategy. \textbf{Our method outperforms baselines} on video quality, camera pose accuracy, and hand pose accuracy metrics. 
    \item We distill our camera- and hand-conditioned bidirectional teacher model into an \textbf{interactive, autoregressive student model} that runs at interactive frame rates. Using this model, we demonstrate \textbf{improved task accuracy} and \textbf{increased perceived control} in our user studies.
\end{itemize}

\section{Related Work}
\label{sec:formatting}

\subsection{From Video Generation to World Simulation}
Recent progress in diffusion models has significantly advanced video generation. Transformer-based bidirectional models~\cite{hunyuanvideo,ltxvideo,wan2025,sora,veo3techreport} leverage full spatio-temporal attention to produce realistic and temporally coherent sequences. 
However, their bidirectional denoising requires access to the full sequence, limiting their use in interactive scenarios. To support causal prediction and long-horizon rollouts, autoregressive video models have been introduced~\cite{genie3, selfforcing, framePack, causVid}. These methods generate frames sequentially in a manner more consistent with real-world dynamics.
These advances in video generation have motivated the development of \emph{world simulators}, whose goal is to predict the visual consequences of actions given the current state~\cite{learninginteractiverealworldsimulators}. 
Recent advancements~\cite{theMatrixMovingControl, mineworld, animegamer, wonderland, worldmem, cosmos, wu2025video, parkerholder2024genie2, decart2024oasis,yu2025gamefactory,yu2025cam} illustrate how actions can be applied to guide visual outcomes. However, most of these existing approaches rely on coarse action vocabularies such as keyboard and mouse inputs or raw camera poses, which describe scene-level information adequately but do not enable dexterous hand--object interactions. This highlights the need for fine-grained embodied control signals in interactive egocentric video generation.

\begin{figure*}[t!]
    \centering
    \includegraphics[width=\linewidth]{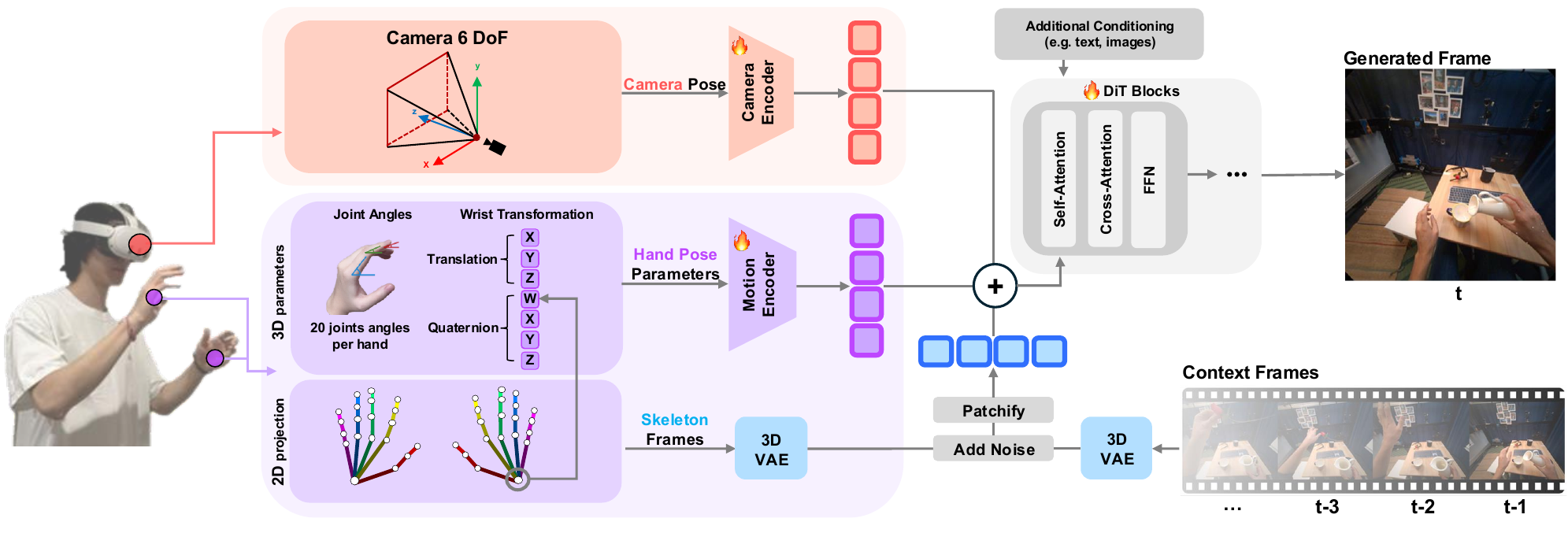}
    \caption{\textbf{Pipeline of generated reality system.} We track the head and hand poses of the user with a commercial headset. Hands are represented using the UmeTrack hand model~\cite{UmeTrack}, which includes translation and rotation of the wrist as well as rotation angles for 20 finger joints per hand. Our conditioning strategy employs a hybrid 2D--3D mechanism, combining a 2D image of the rendered hand skeleton~(purple box, bottom) and the 3D model parameters~(purple box, top). Features extracted from these modules are combined with the head pose features via token addition and fed into the diffusion transformer~(DiT). The diffusion model autoregressively generates new frames at time $t$ using the last few generated frames as context in addition to the user-tracked conditioning signals.}
    \label{fig:pipeline}
\end{figure*}
 
\subsection{Camera- and Hand-conditioned Generation}
In generated virtual environments, camera and hand motions jointly determine how people perceive and interact with their surroundings, making both modalities essential control signals for egocentric world simulators. Camera-conditioned video generation has been extensively explored with various condition-injection strategies~\cite{wang2024motionctrl,he2025cameractrl,cameractrlii,recammaster,AC3D}. For instance, ReCamMaster~\cite{recammaster} injects camera extrinsic parameters through a dedicated camera encoder; CameraCtrl2~\cite{cameractrlii} encodes Plücker rays and adds them element-wise to visual features before the DiT module; and AC3D~\cite{AC3D} adopts a more dynamic design by introducing camera embeddings via a ControlNet-style feedback branch.
In contrast, hand-conditioned video generation remains relatively underexplored. PlayerOne~\cite{PlayerOne} adds body pose embeddings to visual tokens before the DiT backbone, while PEVA~\cite{PEVA} extends adaptive layer normalization~(AdaLN) to inject pose information. However, both methods treat hands merely as part of the full-body pose, thereby limiting the granularity of hand control. InterDyn~\cite{akkerman2025interdyn} employs binary masks instead of pose parameters as conditioning signals, which, however, increases the ambiguity between hand size and depth. In this work, we systematically compare various joint-level hand-conditioning strategies and identify a novel hybrid 2D--3D strategy that outperforms baselines on relevant metrics. We then incorporate this strategy into a camera-controlled video generation model, distill it into an autoregressive video generator, and evaluate this with users in an immersive format.

\section{Conditional Video Generation with Tracked Head and Hands}
\label{sec:method}

In this section, we briefly review preliminaries on video diffusion models (Sec.~\ref{sec:method:preliminaries}). We then discuss hand pose representations and video model conditioning strategies, proposing a novel hybrid 2D--3D conditioning strategy (Sec.~\ref{sec:method:hands}). We describe how to extend this framework to jointly condition on tracked head/camera poses, as well as joint-level hand signals (Sec.~\ref{sec:method:cameraandhands}).

\subsection{Preliminaries}
\label{sec:method:preliminaries}

Our study builds upon the Wan family of video generation models~\cite{wan2025}, a latent video diffusion transformer capable of generating temporally coherent video from a single input image or text prompt. The model consists of a 3D variational autoencoder $(\mathcal{E}, \mathcal{D})$ and a transformer-based diffusion model parameterized by $\Theta$. Given an input latent $z_0 = \mathcal{E}(V_0)$, the forward process follows the rectified flow formulation~\cite{esser2024scalingrectifiedflowtransformers}, where the noised latent is generated by linear interpolation:
\begin{equation}
z_t = (1 - t)\, z_0 + t\,\epsilon, \quad \epsilon \sim \mathcal{N}(0, I)
\label{eq:zt_interp}
\end{equation}
with timestep $t \in [0, 1]$. The denoising process learns a velocity field $v_\Theta(z_t, t)$ that guides the transformation of noise back to data. The model is trained using a conditional flow matching~\cite{lipman2023flowmatching}, with objective:
\begin{equation}
\mathcal{L}_{\text{CFM}} =
\mathbb{E}_{t, z_0, \epsilon}\!\left[\left\lVert
v_\Theta(z_t, t) - u_t(z_0 \mid \epsilon)
\right\rVert_2^2\right]
\label{eq:l_cfm}
\end{equation}
where $u_t$ is the target velocity derived analytically from the forward process. At inference, a sequence of latent frames is recovered by integrating $v_\Theta$ over time.

In the image-to-video (I2V) setting, the model is conditioned on an initial image $I_0$ encoded as $z_{\text{img}} = \mathcal{E}(I_0)$. The transformer-based denoiser $\mathcal{F}_\Theta$ autoregressively predicts video latents $\{z^{(f)}\}_{f=1}^F$, starting from $z_{\text{img}}$ and producing temporally consistent sequences. In the text-to-video (T2V) setting, the model is instead conditioned on a text prompt $p$ encoded as $z_{\text{text}}=\mathcal{T}(p)$ and starts the autoregressive generation from noise. The final video is reconstructed as $\hat{V} = \mathcal{D}(z^{(1)}, \dots, z^{(F)})$.

\subsection{Hand Pose-conditioned Video Generation}
\label{sec:method:hands}

Conditioning strategies for video diffusion models have been widely explored, yet joint-level hand poses remain a challenging modality due to their high dimensionality and complex articulation.
We systematically study how to integrate hand poses into video diffusion transformers~(DiT), focusing on two design choices:  
(1) the hand pose \textbf{representation}, i.e., how to represent tracked user hands, and  
(2) the \textbf{conditioning strategy}, i.e., how the conditioning information is injected into the generative model.

\paragraph{Hand Pose Representation.}
One option for the hand pose representation is a ControlNet-style pose video~\cite{controlnet}. This representation is essentially a sequence of images that visualize the positions of human body joints and corresponding bones in the 2D pixel image space. In the context of egocentric hand conditioning, the video encodes a 2D hand skeleton rendered from the user's viewpoint.

While a skeleton video representation serves as a control signal spatially aligned with the image space, it inherently lacks 3D information. In immersive applications, a hand pose representation with 3D information is crucial for interactive video generation: in isolation, a 2D skeleton video exhibits depth ambiguity and suffers from self-occlusion as overlapping components of the skeleton make the position of certain hand joints ambiguous.

A 3D-aware hand representation is required for dexterous manipulation without ambiguities. Relevant parametric hand models are well known~\cite{UmeTrack, MANO} and usually model a hand pose as a 6 degree-of-freedom (DoF) transformation of the wrist along with rotation angles of each finger joint.
We refer to the wrist pose and local joint rotations collectively as hand pose parameters~(HPP). Applying standard forward kinematics to the HPP analytically yields the full set of 3D poses for all joints.

For compatibility with our training data~\cite{hot3d}, we adopt the UmeTrack hand model, whose HPP consist of 20 joint angles describing hand articulation together with the wrist pose.
These HPP provide metric precision in depth and hand articulation, complementing the coarse but spatially grounded skeleton video representation.

\paragraph{Hand Pose Conditioning.}
To effectively incorporate hand pose parameters into the generative backbone, we examine four widely used condition injection strategies:  
(1) \emph{token concatenation}, (2) \emph{token addition}, (3) \emph{adaptive layer normalization (AdaLN)}, and (4) \emph{cross-attention fusion}.  
A pretrained variational autoencoder with encoder $\mathcal{E}$ projects a hand-contained raw video $V_r$ into the latent space,  
$z_r = \mathcal{E}(V_r)$, where $z_r \in \mathbb{R}^{b \times f \times c \times h \times w}$ is the latent of the raw video and $b$, $f$, $c$, and $h \times w$ denote batch size, frame count, channel dimension, and spatial size, respectively. We additionally extract the hand pose parameters of the same video, denoted as $H \in \mathbb{R}^{b \times f \times d}$, where $d$ is the dimensionality of the HPP.
For token concatenation~(1), we add additional input channels to the input convolutional layer and concatenate the embedded HPP features with the video latents along the channel dimension before patchification:
\begin{equation}
x = \operatorname{patchify}\!\left([\; z_r,\; \mathcal{E}_{\text{conv}}(H) \;]_{\text{channel-dim}}\right)
\label{eq:patchify_concat_hpp}
\end{equation}
where $\mathcal{E}_{\text{conv}}$ denotes a lightweight motion encoder composed of 1D convolutional layers.
For token addition~(2), conditioning is applied through element-wise addition of HPP embeddings to patch tokens:
\begin{equation}
x = \operatorname{patchify}(z_r) \;+\; \mathcal{E}_{\text{conv}}(H)
\label{eq:patchify_add_hpp}
\end{equation}
For AdaLN~(3), the hand features modulate the activations within each DiT block through adaptive scale and shift vectors,  
a method inspired by adaptive normalization in conditional transformers~\cite{dit}:
\begin{equation}
x = \alpha(H) \odot v_r + \beta(H)
\label{eq:film}
\end{equation}
where $\alpha(H)$ and $\beta(H)$ are learned from $H$, and $\odot$ denotes the Hadamard product.
Finally, for cross-attention fusion~(4), HPP embeddings serve as keys and values in motion-conditioned cross-attention layers injected after selected Transformer blocks,  
following the after-block cross-attention design of recent works~\cite{wans2v}:
\begin{equation}
x^{(l+1)} = x^{(l)} + \operatorname{CrossAttn}\!\big(x^{(l)},\, \mathcal{E}_{\text{conv}}(H)\big)
\label{eq:crossattn_update}
\end{equation}

\paragraph{Hybrid 2D--3D Hand Pose Conditioning.}
We propose a hybrid conditioning scheme that combines ControlNet-style 2D skeleton videos with the 3D-aware HPP. This strategy combines the efficiency of ControlNet with the spatial awareness of HPP. 
As shown in Sec.~\ref{sec:exp}, \emph{token addition} yields the best performance among the evaluated pose injection approaches. We therefore incorporate HPP into the skeleton-based video control branch via element-wise token addition.  
Specifically, a hand-contained raw video $V_r$ and its corresponding skeleton video $V_c$ are encoded by the same VAE encoder $\mathcal{E}$ to obtain $z_r$ and $z_c$, respectively.  
We then concatenate the two latents in a channel-wise manner, and inject the HPP features using token addition:
\begin{equation}
x = \operatorname{patchify}\!\left([\; z_r,\; z_c \;]_{\text{channel-dim}}\right)
\;+\;
\mathcal{E}_{\text{conv}}(H)
\label{eq:joint_conditioning_simple}
\end{equation}
This design allows the model to resolve depth and self-occlusion ambiguity while maintaining strong spatial grounding from the skeleton representation.

\subsection{Joint Camera and Hand Control}
\label{sec:method:cameraandhands}
In head-mounted display~(HMD) formats, visual content must be generated dynamically based on user interaction. Therefore, the user's viewpoint~(camera), left hand, and right hand are foundational control signals for interactive video generation. Hand interaction enables intent-driven movement of generated objects, and viewpoint interaction enables the user to view the generated content from new perspectives. To support these interactions, we introduce a framework for \textbf{joint hand and camera conditioning}, enabling realistic egocentric video generation driven by natural user interactions.

\begin{table*}[htbp]
  \centering
  \caption{\textbf{Quantitative comparison of hand-motion conditioning strategies.}
We perform an ablation study on the Wan2.2 14B model, evaluating hand pose parameters (HPP), binary mask, skeleton video, and hybrid conditioning schemes. Results are reported for both video quality as well as 3D and 2D hand pose accuracy, where best results are highlighted as  \colorbox{colorFst}{\bf first} and  \colorbox{colorSnd}{\underline{second}}. Our hybrid strategy using both 2D skeleton projection and 3D HPPs achieves the best accuracy while maintaining a competitive video quality. Note that the position errors here are in millimeters and Procrustes aligned. ControlNet* represents the use of pixel-level image conditioning, but we do not copy the DiT blocks as done in the original ControlNet implementation.
}
\small
    \begin{tabular}{p{5.5em}p{14.5em}|cccc|ccc}
    \toprule
    \multicolumn{2}{c|}{\multirow{2}[4]{*}{\textbf{Method}}} & \multicolumn{4}{c|}{\textbf{Video Quality}} & \multicolumn{3}{c}{\textbf{Hand Pose Accuracy}} \\
\cmidrule{3-9}    \multicolumn{2}{c|}{} & \multicolumn{1}{p{3em}}{\textbf{PSNR↑}} & \multicolumn{1}{p{3em}}{\textbf{LPIPS↓}} & \multicolumn{1}{p{3em}}{\textbf{SSIM↑}} & \multicolumn{1}{p{3em}|}{\textbf{FVD↓}} & \multicolumn{1}{p{3.5em}}{\textbf{MPJPE↓}} & \multicolumn{1}{p{3.5em}}{\textbf{MPVPE↓}} & \multicolumn{1}{p{3.5em}}{\textbf{L2Err↓}} \\
    \midrule
    No Cond. & Baseline (Wan 2.2 Video 14B) & 14.59 & 0.4872 & 0.4855 & 601.55 & 17.86 & 12.29 & 67.50 \\
    \midrule
    \multirow{4}[2]{*}{HPP Cond.} & TokenConcat (PlayerOne~\cite{PlayerOne}) & 15.09 & 0.4633 & 0.4983 & 560.34 & 18.02 & 12.34 & 65.43 \\
          & AdaLN (PEVA~\cite{PEVA}) & 15.02 & 0.4591 & 0.4906 & 677.26 & 18.49 & 12.53 & 65.97 \\
          & CrossAttention & 14.71 & 0.4686 & 0.4840 & 662.22 & 17.56	& 12.04 & 63.23 \\
          & TokenAddition (ReCamMaster~\cite{recammaster}) & 15.19 & 0.4520 & 0.4975 & 601.15 & 17.84 & 12.14 & 56.66 \\
    \midrule
    \multirow{2}[2]{*}{Video Cond.} & Binary Mask (InterDyn~\cite{akkerman2025interdyn}) & 16.58 & 0.3947 & 0.5533 & \fs{356.11} & 12.83 & 9.56 & 35.64 \\
          & Skeleton Video (ControlNet*~\cite{controlnet}) & \fs{16.89} & \fs{0.3837} & \fs{0.5601} & 389.26 & \rd{12.38} & \rd{9.25} & \rd{11.72} \\
    \midrule
    Hybrid Cond. & Skeleton Video + HPP Cond. & \rd{16.85} & \rd{0.3874} & \rd{0.5574} & \rd{383.69} & \fs 12.23 & \fs 9.10 & \fs 11.50 \\
    \bottomrule
    \end{tabular}%
  \label{tab:ablation_hand}%
\end{table*}%

\begin{table*}[htbp]
  \centering
  \caption{\textbf{Quantitative comparison of joint hand and camera conditioning strategies.}
   Compared with the camera-only and hand-only baselines, JointCtrl achieves the best overall performance across video quality, hand pose, and camera pose metrics. It maintains the highest visual quality while delivering competitive control accuracy for both hand and camera signals, relative to models specialized in a single modality. Translation and rotation errors are reported in meters and degrees, respectively.
   }
\small
    \begin{tabular}{lp{6.3em}|cccc|ccc|cc}
    \toprule 
    \multicolumn{2}{c|}{\multirow{2}[4]{*}{\textbf{Method}}} & \multicolumn{4}{c|}{\textbf{Video Quality}} & \multicolumn{3}{c|}{\textbf{Hand Pose Accuracy}} & \multicolumn{2}{c}{\textbf{Camera Pose Accuracy}} \\ 
\cmidrule{3-11}    \multicolumn{2}{c|}{} & \multicolumn{1}{p{3em}}{\textbf{PSNR↑}} & \multicolumn{1}{p{3em}}{\textbf{LPIPS↓}} & \multicolumn{1}{p{3em}}{\textbf{SSIM↑}} & \multicolumn{1}{p{3em}|}{\textbf{FVD↓}} & \multicolumn{1}{p{3.5em}}{\textbf{MPJPE↓}} & \multicolumn{1}{p{3.5em}}{\textbf{MPVPE↓}} & \multicolumn{1}{p{3.5em}|}{\textbf{L2Err↓}} & \multicolumn{1}{p{3.5em}}{\textbf{TransErr↓}} & \multicolumn{1}{p{3em}}{\textbf{RotErr↓}} \\
    \midrule
    CamCtrl & CameraCtrl~\cite{he2025cameractrl} & \rd{18.58}  &  \rd{0.2943}  & \rd{0.6099}  & 558.94     & 18.37     & 12.72      & 50.33     & \fs{0.23}     & \fs{2.77} \\
    \midrule
    HandCtrl & Best in Tab. 1  & 16.85 & 0.3874 & 0.5574 & \fs{383.69} & \fs{12.23} & \fs{9.10} & \fs{11.50} & 2.27 & 13.40 \\
    \midrule
    JointCtrl & Ours  &  \fs{18.60} & \fs{0.2800} & \fs{0.6173} & \rd{396.93}   & \rd{12.81} & \rd{9.66} & \underline{13.42}  & \rd{0.25} & \rd{2.79} \\
    \bottomrule
    \end{tabular}%
  \label{tab:ablation_w_cam}%
\end{table*}%

\paragraph{Camera Pose Representation.}
Previous works on pose-conditioned video generation often infer camera poses implicitly from body kinematics. For example, PlayerOne~\cite{PlayerOne} estimates rotation-only camera trajectories from head pose with exocentric videos, while
PEVA~\cite{PEVA} models viewpoint change via body joint signals without explicitly modeling camera extrinsics.
In contrast, we directly exploit the built-in inertial sensors and egocentric cameras of modern HMD, which provide a 6-DoF camera pose in world space, including both rotation ($r \in \mathbb{R}^{3 \times 3}$) and translation ($t \in \mathbb{R}^{3}$). 
This explicit camera representation enables the accurate modeling of the camera (or head) pose, making the generated video responsive to a user's head motion.

\paragraph{Joint Conditioning Strategy.}
We transform the 6-DoF camera poses into per-frame Plücker embeddings $P\in\mathbb{R}^{b \times f \times 6 \times h \times w}$~\cite{sitzmann2022lightfieldnetworksneural}, which are then projected into the same shape as the patch tokens with encoder $\mathcal{E}_{\text{cam}}$. We then apply element-wise addition over three components in the latent space: (a) video latents, (b) HPP embeddings, and (c) camera embeddings:
\begin{equation}
\begin{aligned}
x &= \operatorname{patchify}\!\left([\; z_r,\; z_c \;]_{\text{channel-dim}}\right) \\
  &\quad + \mathcal{E}_{\text{conv}}(H)
  + \mathcal{E}_{\text{cam}}(P)
\end{aligned}
\label{eq:joint_conditioning}
\end{equation}
The fused representation $x$ is then passed into the DiT blocks for generation. During training, both hand and camera signals are jointly optimized under a unified conditioning schema, ensuring coherent motion alignment between user actions and egocentric viewpoint changes. An overview of this joint conditioning architecture is shown in Figure~\ref{fig:pipeline}.

\paragraph{Iterative Encoder Training.}
In practice, we find jointly training both encoders from scratch to be unstable. We attribute this to (1) both camera and HPP embeddings being added in the same operation and (2) ambiguity between motion caused by hand interaction and camera movement. Thus, we adopt an iterative training approach: camera and HPP encoders are first trained independently, with the camera encoder weights initialized from the FUN model~\cite{wan2025}. Then, both encoders are trained jointly in a final fine-tuning step to merge the conditionings.

\begin{figure}[t!]
    \centering
    \caption{\textbf{Qualitative comparison of hand-pose conditioning strategies.}
    Ground-truth conditioning hand input is shown in red. Predicted hands are orange; overlap is green. Our hybrid conditioning strategy is most accurate among these baselines, especially when hands are partly occluded at the boundaries of the frame.}
    \label{fig:qualitative_results}

    \newcommand{\figpath}{figs/qualitative_results}

    \begin{tabularx}{\columnwidth}{r >{\centering\arraybackslash}X}
        & \\
        
        \rotatebox[origin=c]{90}{\footnotesize \textbf{Baseline}} &
        \adjustbox{valign=c}{\includegraphics[width=\linewidth]{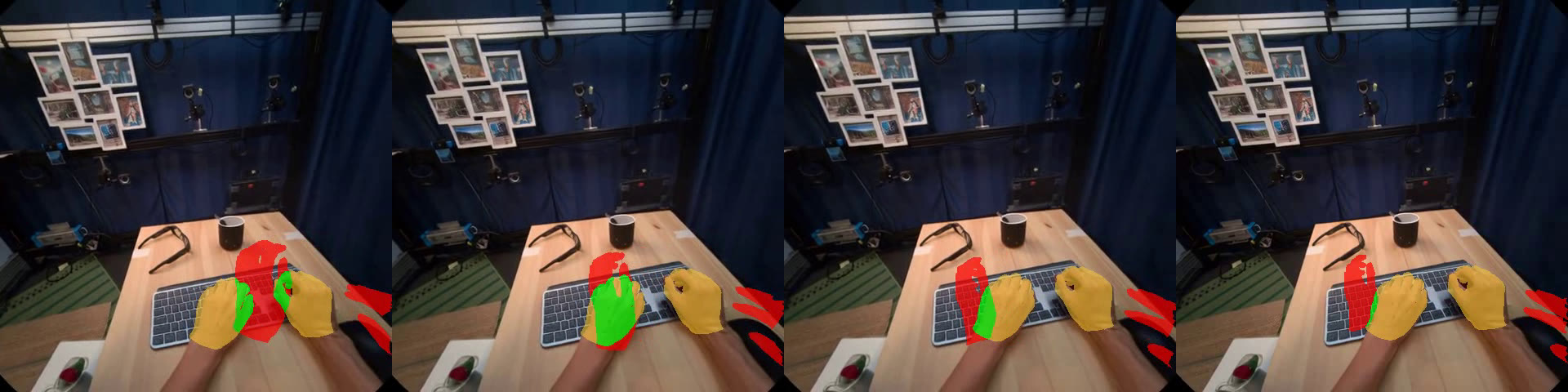}} \\
        \addlinespace

        \rotatebox[origin=c]{90}{\footnotesize \textbf{HPP Cond.}} &
        \adjustbox{valign=c}{\includegraphics[width=\linewidth]{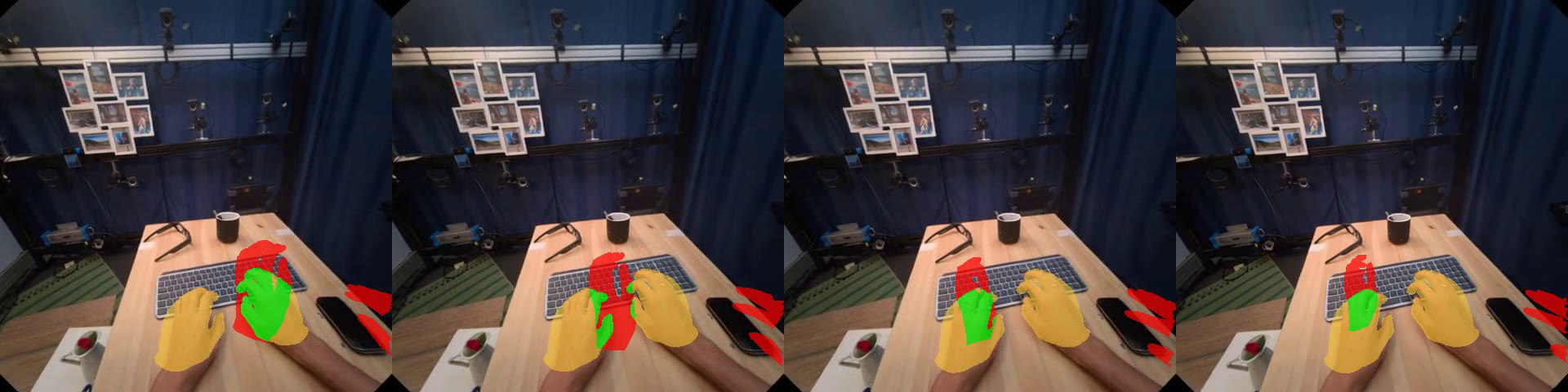}} \\
        \addlinespace

        \rotatebox[origin=c]{90}{\footnotesize \textbf{Video Cond.}} &
        \adjustbox{valign=c}{\includegraphics[width=\linewidth]{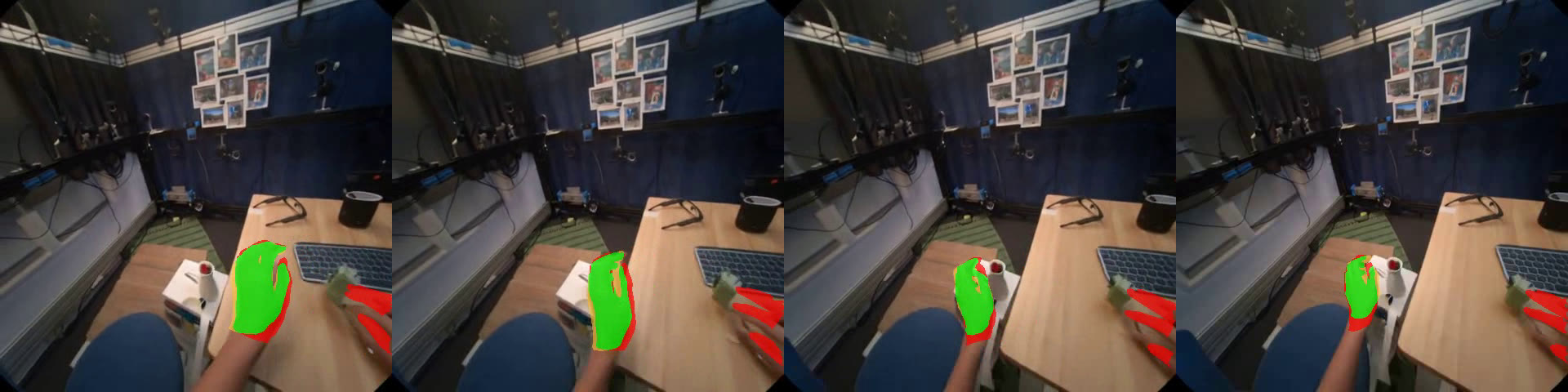}} \\
        \addlinespace

        \rotatebox[origin=c]{90}{\footnotesize \textbf{Hybrid}} &
        \adjustbox{valign=c}{\includegraphics[width=\linewidth]{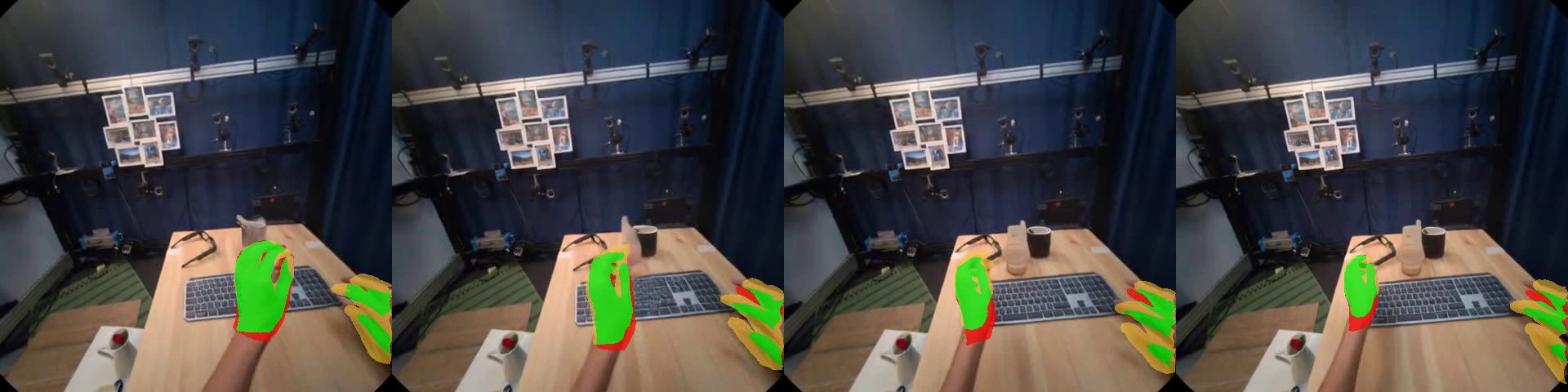}} \\
    \end{tabularx}

\end{figure}

\section{Experiments}
\label{sec:exp}

\paragraph{Implementation details.} 
Building upon the Wan2.2 14B image-to-video (I2V) generation model~\cite{wan2025}, we first conduct a systematic study to determine the most effective hand motion conditioning strategy. 
Experiments are performed on the HOT3D dataset~\cite{hot3d}, which captures hand–object interactions with precise 3D hand annotations obtained via optical-marker motion capture and synchronized camera pose annotations.
We segment each video into 5-second clips, yielding 5824 training samples, and reserve an unseen sequence of 45 clips for evaluation. 
For each of the conditioning strategies described in Sec.~\ref{sec:method:hands}, we train LoRA~\cite{hu2022lora} modules with rank~32 on both low-noise and high-noise experts for over 1K steps at a resolution of $480\times480$, using a learning rate of $1\times10^{-5}$ and a batch size of~16. 

\paragraph{Metrics.} We evaluate our model along three dimensions: overall video quality, hand pose accuracy, and camera pose accuracy. For video quality, we report PSNR for pixel-level accuracy, LPIPS~\cite{zhang2018unreasonableeffectivenessdeepfeatures} for perceptual similarity, SSIM for structural consistency, and Fréchet Video Distance (FVD) for distribution-level realism. 
For hand pose accuracy, we use  WiLoR~\cite{wilor} to evaluate Procrustes Aligned Mean Per-Joint Position Error (PA-MPJPE) computed over 20 joints to measure 3D pose accuracy, and Procrustes Aligned Mean Per-Vertex Position Error (PA-MPVPE) computed over 778 vertices to measure 3D hand shape accuracy. We further compute the average L2 distance between ground truth and generated hand landmarks in the pixel space of each 2D frame~\cite{handPoseEstimate}. Camera pose accuracy is evaluated by extracting estimated trajectories from generated clips using GLOMAP~\cite{glomap} and computing rotation error (RotErr) and translation error (TransErr) following previous work~\cite{recammaster}.

\paragraph{Evaluating Hand-pose Conditioning.} 
Among the four injection strategies evaluated for conditioning on hand pose parameters (HPP), the \emph{token addition} method achieves the best performance across hand pose accuracy metrics, as shown in Table~\ref{tab:ablation_hand}.  
In contrast, \emph{cross-attention} and \emph{AdaLN} struggle to establish a stable mapping between HPP and visual features, likely due to the limited scale of the HOT3D dataset and the high dimensionality of the HPP, performing worse than the unconditioned baseline.

\begin{figure}[t!]
    \centering
    \caption{\textbf{Qualitative comparison of joint hand--camera control.}
    Ground-truth (GT), camera-only, hand-only, and joint-control results.
    Camera-Ctrl and Hand-Ctrl are effective at controlling one of these modalities but not the other.
    Our Joint-Ctrl mechanism enables simultaneous control of camera and hands.}
    \label{fig:qualitative_joint_cond}

    \newcommand{\figpath}{figs/qualitative_w_cam}

    \begin{tabularx}{\columnwidth}{r >{\centering\arraybackslash}X}
        \rotatebox[origin=c]{90}{\footnotesize \textbf{GT}} &
        \adjustbox{valign=c}{\includegraphics[width=\linewidth]{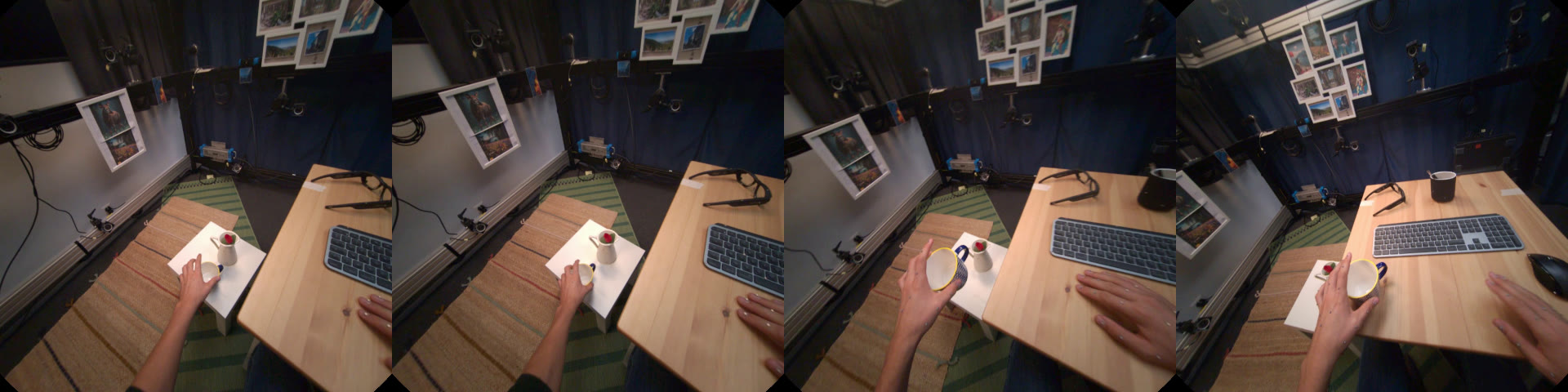}} \\
        \addlinespace[4pt]

        \rotatebox[origin=c]{90}{\footnotesize \textbf{Camera-Ctrl}} &
        \adjustbox{valign=c}{\includegraphics[width=\linewidth]{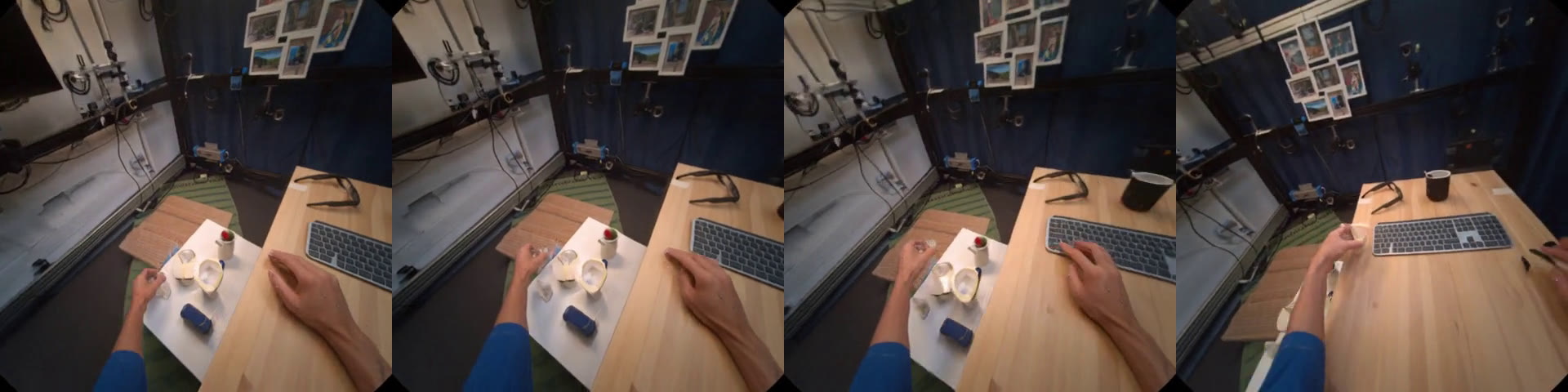}} \\
        \addlinespace[4pt]

        \rotatebox[origin=c]{90}{\footnotesize \textbf{Hand-Ctrl}} &
        \adjustbox{valign=c}{\includegraphics[width=\linewidth]{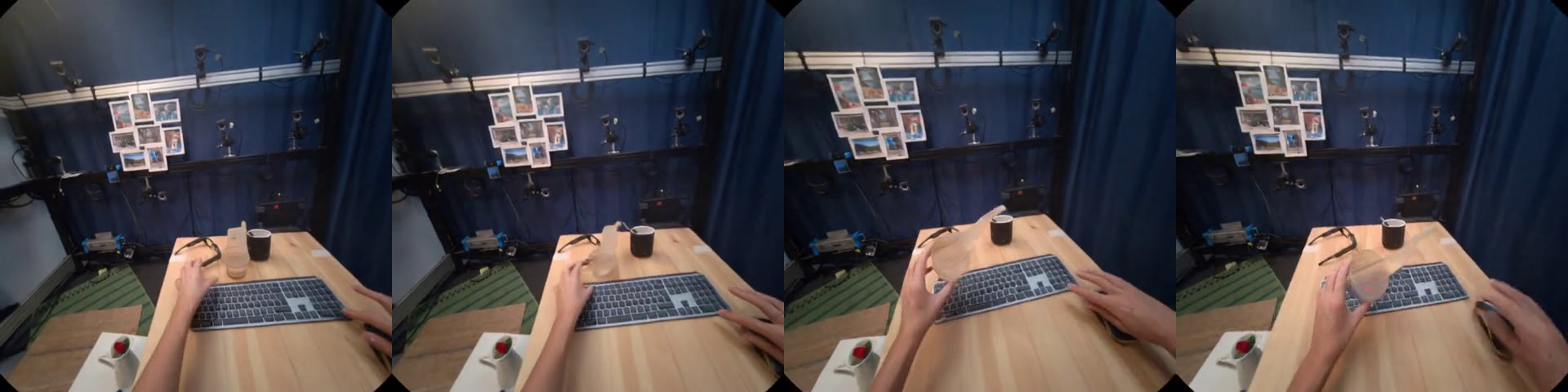}} \\
        \addlinespace[4pt]

        \rotatebox[origin=c]{90}{\footnotesize \textbf{Joint-Ctrl}} &
        \adjustbox{valign=c}{\includegraphics[width=\linewidth]{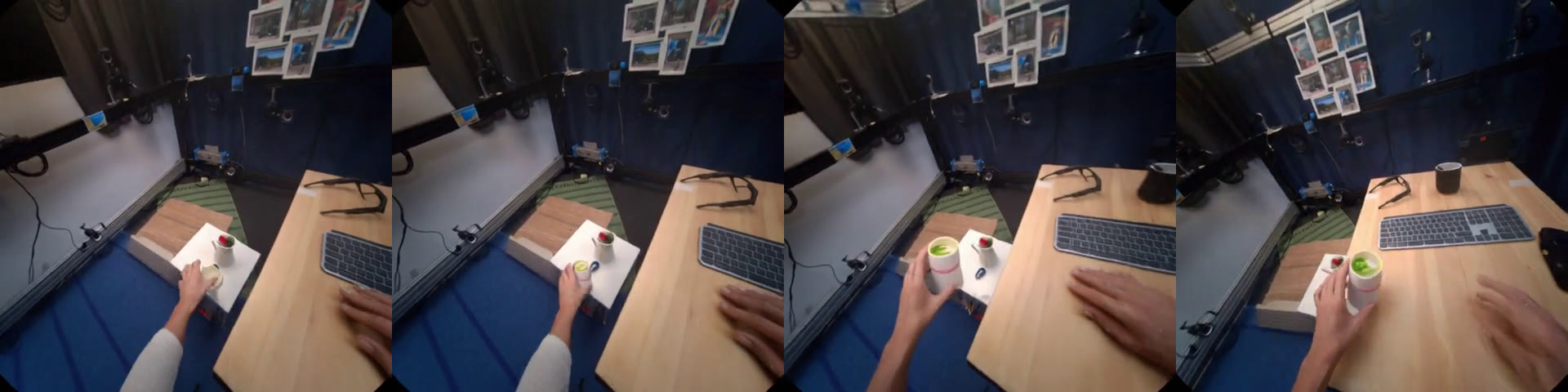}}
    \end{tabularx}
\end{figure}

We further evaluate hybrid conditioning that integrates both skeleton video and HPP information.  
As shown in Table~\ref{tab:ablation_hand}, the hybrid approach achieves the best performance across all hand accuracy metrics. 
Although the numerical gains over the ControlNet-style 2D skeleton-image conditioning strategy are moderate, likely due to the relatively simple hand motions in HOT3D, the hybrid 2D--3D method still produces more stable and anatomically faithful hand reconstructions qualitatively.

To contextualize the quantitative results for hand pose accuracy, we estimate lower bounds for the different metrics by evaluating the HOT3D test annotations under the same protocol, i.e., by fitting a 3D hand model using WiLoR~\cite{wilor} to the ground truth test images and evaluating our hand pose accuracy metrics.
This yields MPJPE of 9.42, MPVPE of 7.74, and an L2 landmark error of 9.08, representing the inherent accuracy and uncertainty of the WiLoR-based hand pose estimator we use for all generated frames. Table~\ref{tab:ablation_hand} (right) shows that our hybrid conditioning method approaches this lower bound. To further validate robustness beyond HOT3D, we evaluate on the larger GigaHands dataset~\cite{gigahands} and observe consistent improvements over 2D-only conditioning.

Qualitative comparisons in Figure~\ref{fig:qualitative_results} highlight these improvements, where predicted hands are shown in orange, ground truth in red, and their overlap in green.
In the challenging case shown in this figure, ControlNet conditioning fails to reconstruct hands near the image boundary due to incomplete skeleton inputs, whereas the hybrid model generates complete and spatially consistent hand structures even when hands are close to the frame edge.

\paragraph{Evaluating Joint Head- and Hand-pose Conditioning.} 
We compare the proposed joint hand–camera conditioning framework against hand-only (HandCtrl) and camera-only (CameraCtrl~\cite{he2025cameractrl}) baselines. 
As shown in Table~\ref{tab:ablation_w_cam}, the joint-control model achieves the best video quality and balanced performance across hand and camera pose metrics. 
Specifically, CameraCtrl achieves the lowest rotation and translation errors in camera pose but fails to maintain accurate hand alignment, whereas HandCtrl produces precise hand poses but lacks camera control. 
Our joint-control model bridges this gap, achieving coherent coordination between hand motion and head dynamics. 
Figure~\ref{fig:qualitative_joint_cond} further illustrates that, without camera control, the hand-only model often interacts with incorrect objects. 
In this example, the hand-only model incorrectly predicts user intent by reaching toward an object on the table instead of the cup on the left.

\section{The Generated Reality System}

Using our detailed analysis of joint-level hand- and head--conditioned video generation, we next develop our generated reality system. This is a variant of the aforementioned video diffusion model, rolled out in a causal, i.e., autoregressive, manner and distilled to achieve interactive frame rates. The user's head and hand poses are dynamically tracked with a commercial VR system and used to condition the video generation model, whose output is streamed directly to the headset worn by the user.

\paragraph{Autoregressive Distillation.}
Following the self-forcing strategy, we distill a bidirectional Wan2.2 5B teacher model that is trained with our head- and hand-conditioning strategy into a causal 5B student model~\cite{selfforcing}. Autoregressive videos are generated in 12-frame chunks, complete with per-frame hand and head conditioning as outlined. The model supports both image-to-video (I2V) and text-to-video (T2V) settings. The resulting system provides a closed-loop generative experience---users can continuously move their hands and head, and the model renders the corresponding virtual response.

\paragraph{Integration with VR System.}
A real-time generative VR system is implemented with Unity on the Meta Quest 3. We use the captured head and hand poses from the Quest as our conditioning. This conditioning is streamed to a server hosting the distilled autoregressive model. For each video chunk, conditionings are read from a circular frame buffer with the most recent tracked data. Generated video chunks are then streamed back to the Quest 3 for interactive viewing in VR. We achieve 11 FPS in real-time with 1.4 seconds of latency on a single H100 GPU. The latency is bottlenecked by the time to generate and decode a 12-frame chunk. The added conditioning adds only an additional 0.002~s of latency.

\paragraph{User Study Design.}

To evaluate our generated reality system, we conducted two user studies. For this purpose, we recruited 11 subjects (age range = 22--30 years). The cohort consisted of 4 female and 7 male participants; 6 of them wore glasses. All participants reported normal or corrected-to-normal vision.

\begin{figure}[t!]
    \centering
    \includegraphics[width=\linewidth]{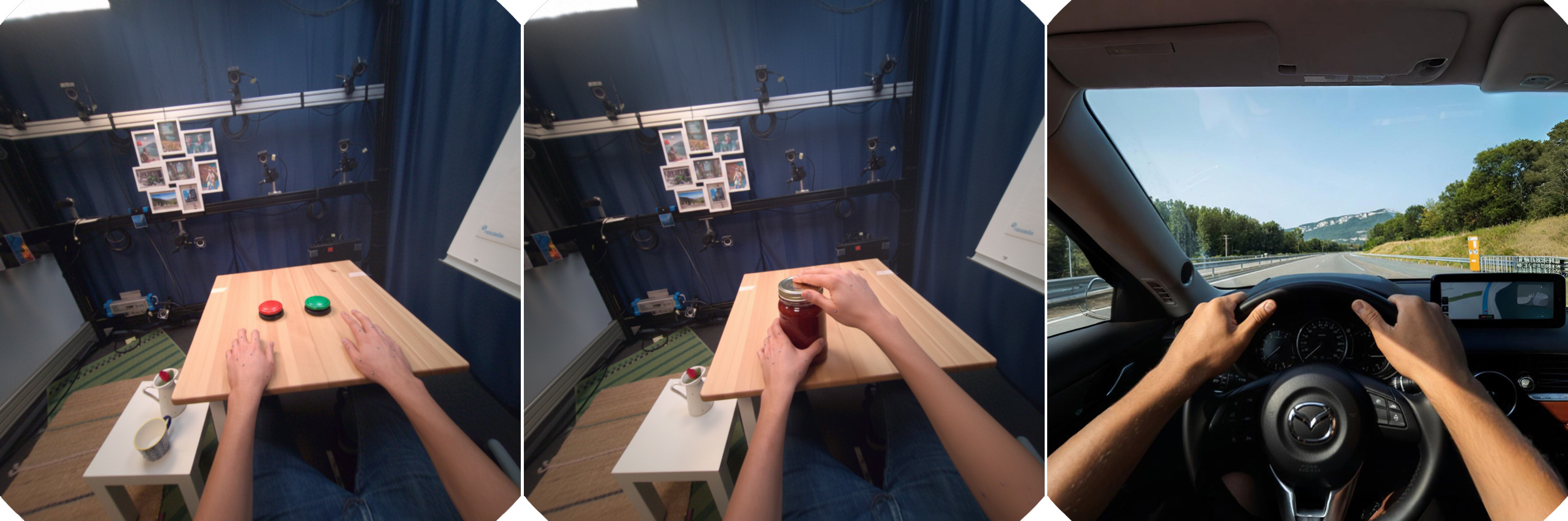}\\
    \vspace{5pt}
    \includegraphics[width=\linewidth]{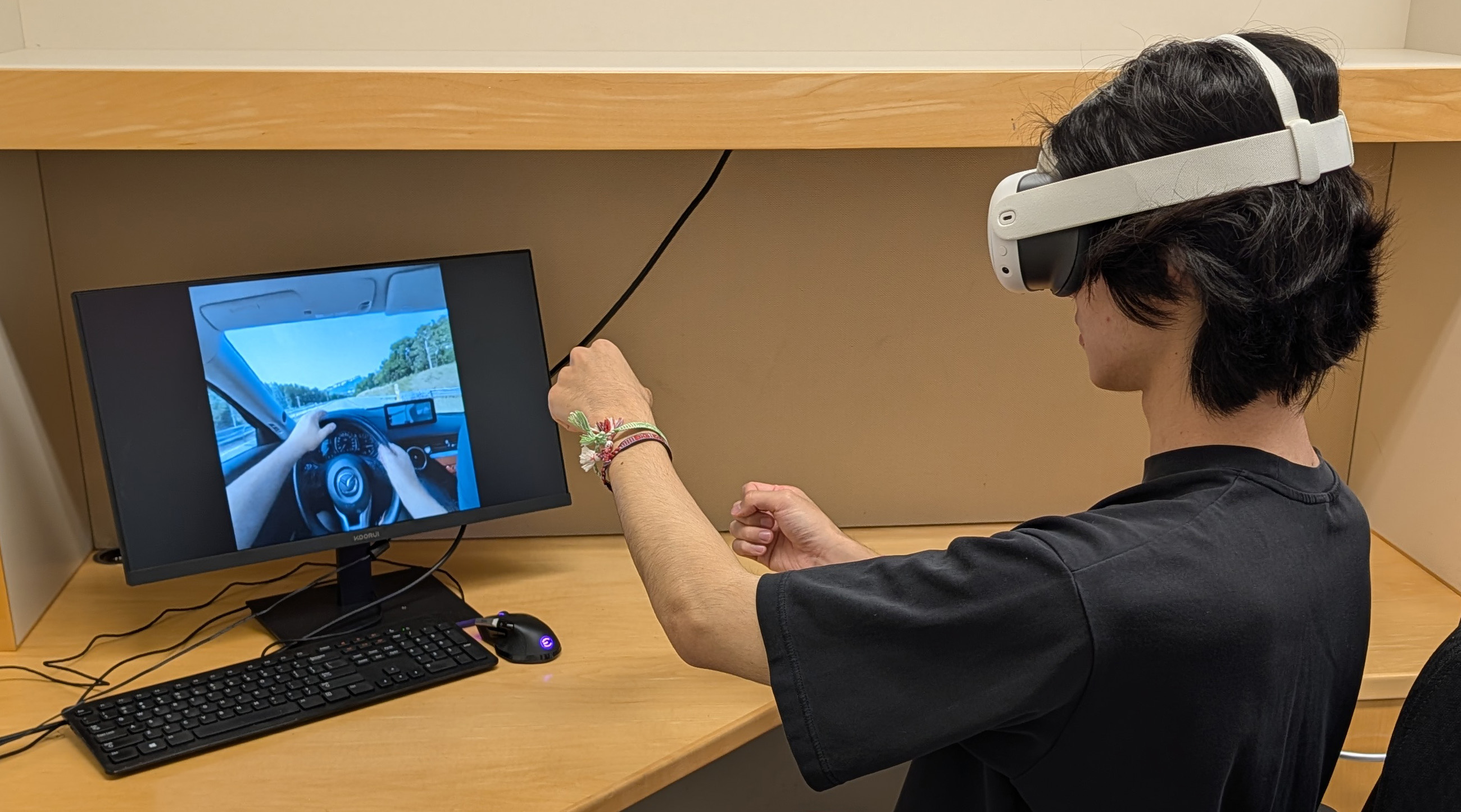}
    \caption{\textbf{User study tasks and setup}. Our subjects completed three tasks using a commercial virtual reality headset: ``push the green button'', ``open the jar'', and ``turn the steering wheel''. Representative screenshots of all three tasks from the perspective seen by the user (top). A photo of our setup, in which the generated video's hands reflect the user's in real-time (bottom).}
    \label{fig:userstudy:tasks}
\end{figure}

We designed the three different environments shown in Figure~\ref{fig:userstudy:tasks} for our studies. Observing these in the Quest headset, we ask users to perform the following tasks: ``push the green button'', ``open the jar'', and ``turn the steering wheel'', respectively. Users had a total of 8 seconds to complete a task. We tested two conditions for each task: one using our hand- and head-pose conditioned model and one baseline model that uses only head-pose conditioning. The relative difference between these conditions, therefore, demonstrates the effectiveness of hand control in our application. The baseline relies purely on the text-conditioned video model to complete the task without the user directly controlling the generated rendering of their hands. Users completed each of the three tasks four times~(twice for each of the conditions), all in random order. Before starting each run, we asked users to roughly align their hands with the input image; we overlay their real-time hand pose to assist with this process. Once they indicated alignment, we disabled the hand pose overlay and began the interactive experience, so users saw only the environment, the generated hands, and the results of hand-object interactions. Users were allowed two practice runs to familiarize themselves with the process before we began recording results. More details are outlined in Appendix.

\begin{figure}[t!]
    \centering
    \includegraphics[width=\linewidth]{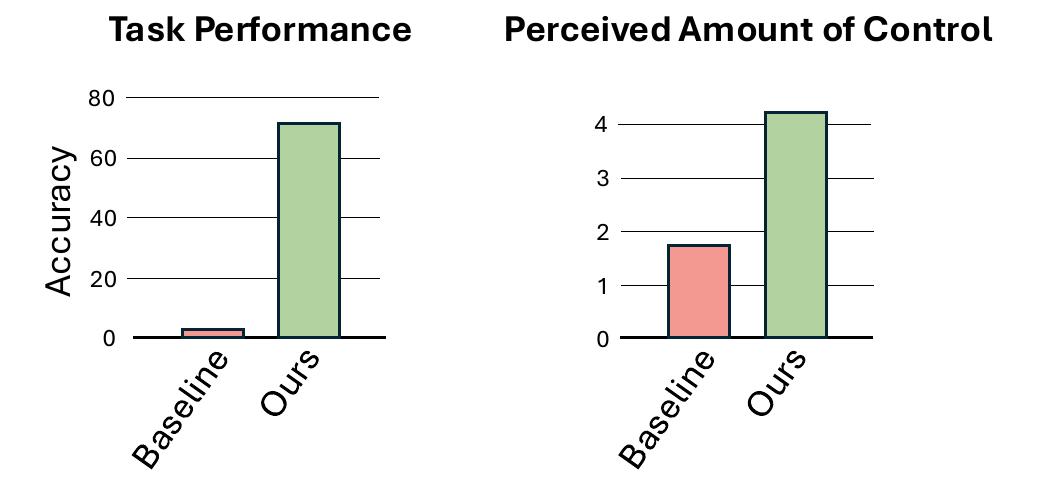}
    \caption{\textbf{User evaluation}. We show human subjects interactive videos without (baseline) and with (ours) tracked hand conditioning signals. For the baseline, we prompt the video model to complete the task using the same instructions provided to human subjects in our setting. In our tracked hand conditioning setting, users can more accurately complete the task than a video model can with just text conditioning (left). Moreover, users report a significantly higher level of perceived control over the interaction in our setting compared to the baseline, measured using a 7-point Likert scale (right). }
    \label{fig:userstudy:taskperformance}
\end{figure}

\paragraph{Evaluating Task Efficiency.} As shown in Figure~\ref{fig:userstudy:taskperformance} (left), the baseline achieved an average of 3.0\% for task accuracy, demonstrating that text prompts alone are insufficient for reliably completing tasks that require fine-grained hand--object interaction. Under identical conditions, our hand-controlled model achieved 71.2\% task accuracy on average, highlighting the substantial improvement in task success provided by explicit hand controls.

\paragraph{Evaluating User Experience.}
After each trial, participants rated their perceived amount of control on a 7-point Likert scale (1 = worst, 7 = best). Shown in Figure~\ref{fig:userstudy:taskperformance} (right), our hand-controlled model received a mean score of 4.21, compared to 1.74 for the baseline. These results indicate that participants experienced markedly greater control over hand pose and movements with explicit hand conditioning than with text prompts alone, aligning with the observed improvements in task success.

\section{Discussion}

We present crucial first steps towards a vision of human-centric world simulation. Specifically, we identify and evaluate efficient and effective mechanisms for conditioning video diffusion models on tracked head and joint-level hand data. Moreover, we present a first version of an interactive generated reality system and demonstrate its efficacy with user studies. 

\paragraph{Limitations.} The resolution, latency, stereo rendering capabilities, image quality, and computing efficiency of our system lag far behind those of modern virtual reality systems. As with all current autoregressive video models, drift significantly degrades the image quality after a few seconds of rollout. Yet, the promise of generating an interactive and immersive virtual environment in a zero-shot manner is unprecedented and motivates future research on solving these issues. 

\paragraph{Future Work.} Improving the aforementioned limitations towards retinal image resolution in stereo with imperceptible (i.e., $<20$~ms) latency and long rollouts on a wearable computer embedded in a headset is an enormous challenge. Yet, most of these problems are well aligned with ongoing research and development efforts on autoregressive video diffusion models across the computer vision and AI communities.

\paragraph{Conclusion.}
Generated reality could enable immersive learning and exploration, letting users acquire skills and practice complex tasks in a zero-shot manner, without the need for laborious modeling of 3D virtual environments. 

\section*{Acknowledgements}
We thank the NVIDIA Academic Grant Program for providing part of the GPU resources for this project.

{
    \small
    \bibliographystyle{ieeenat_fullname}
    \bibliography{main}
}

\maketitlesupplementary
\appendix
\section{Experiment Details}
\subsection{Initialization of the Motion Encoder}

Our experiments are based on the Wan2.2 14B model family, which uses a mixture-of-experts (MoE) architecture with two DiT experts: one specialized for high-noise steps and one for low-noise steps. To train the motion encoder effectively under this design, we adopt a continual training scheme.

During high-noise DiT training, we zero-initialize the motion encoder. After convergence, the trained encoder is transferred and used as the initialization for low-noise training. This two-stage setup provides a stronger starting point for the low-noise model and mitigates the impact of the limited HOT3D dataset, resulting in more stable training and improved motion alignment.

\subsection{Continual Training of DiT Experts}

For hybrid conditioning, we aim to emphasize fine-grained alignment during training. 
To achieve this, we initialize the DiT with the LoRA weights learned from skeleton-video conditioning and continue training from this point. 
This provides the model with a well-structured spatial prior and allows the hybrid training stage to focus on refining articulation and depth cues introduced by the hand pose parameters.

Similarly, for joint hand–camera conditioning, we initialize the DiT with the LoRA weights obtained from the hybrid model and then train with both hand and camera inputs. 
This continual training strategy gives the joint model a strong initialization and leads to more stable convergence and improved motion consistency. Furthermore, it helps the model decouple the conditionings, which are both applied in the same token addition operation.

\subsection{Lower bounds}
\label{subsec:lower_bounds}

We estimate the lower bound of our evaluation pipeline by running the same
metrics on the HOT3D validation annotations themselves.  
Hand poses are obtained from WiLoR~\cite{wilor}, and camera trajectories are
computed using GLOMAP~\cite{glomap}.  
This provides the inherent error level of the annotation and reconstruction
process under our evaluation protocol.

\begin{table}[htbp]
  \centering
  \caption{\textbf{Lower bound for hand and camera pose evaluation metrics.}}
  \small
  \begin{tabular}{ccccc}
    \toprule
    \textbf{MPJPE↓} &
    \textbf{MPVPE↓} &
    \textbf{L2Err↓} &
    \textbf{TransErr↓} &
    \textbf{RotErr↓} \\
    \midrule
    9.42 & 7.74 & 9.08 & 0.0191 & 0.44$^\circ$ \\
    \bottomrule
  \end{tabular}
  \label{tab:lower_bound}
\end{table}

\section{Additional Evaluation}

\subsection{Alternative Datasets}
\label{sec:giga_datasets}
\begin{figure*}[t!]
    \centering
    \newcommand{\figpath}{figs/qualitative_gigahands}
    \setlength{\tabcolsep}{2pt}
    \begin{tabularx}{\textwidth}{r >{\centering\arraybackslash}X}
        \multicolumn{2}{c}{\small\textbf{Scene 1}} \\[2pt]
        \rotatebox[origin=c]{90}{\footnotesize \textbf{Baseline}} &
        \adjustbox{valign=c}{\includegraphics[width=\linewidth]{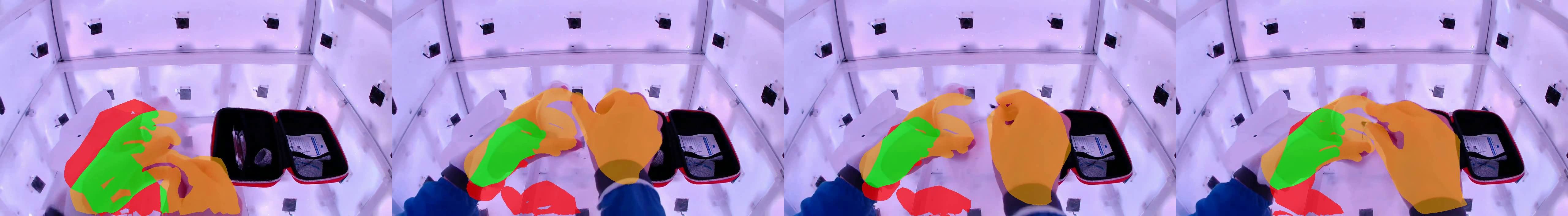}} \\[4pt]
        \rotatebox[origin=c]{90}{\footnotesize \textbf{HPP Cond.}} &
        \adjustbox{valign=c}{\includegraphics[width=\linewidth]{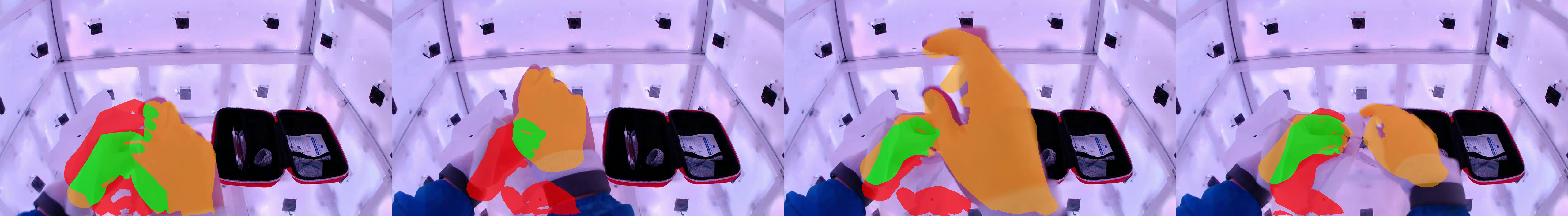}} \\[4pt]
        \rotatebox[origin=c]{90}{\footnotesize \textbf{Video Cond.}} &
        \adjustbox{valign=c}{\includegraphics[width=\linewidth]{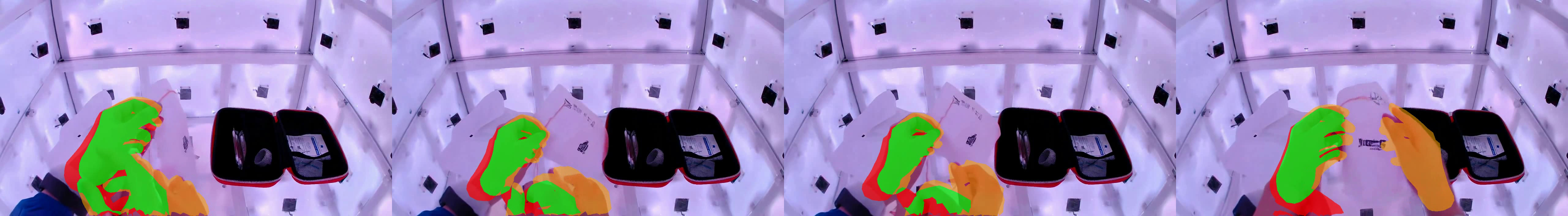}} \\[4pt]
        \rotatebox[origin=c]{90}{\footnotesize \textbf{Hybrid}} &
        \adjustbox{valign=c}{\includegraphics[width=\linewidth]{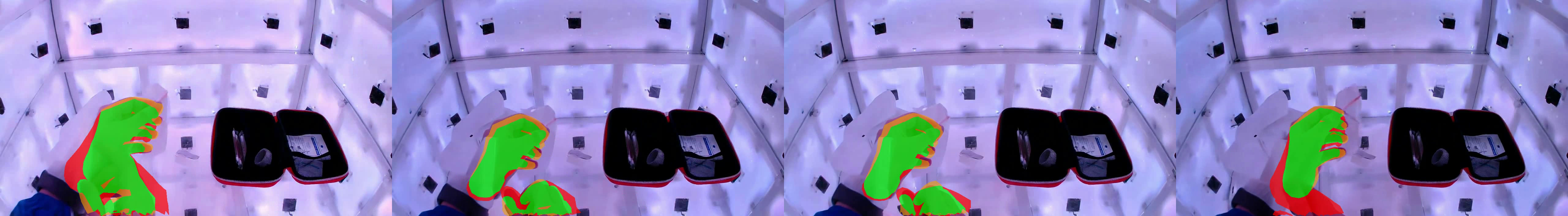}} \\
    \end{tabularx}

    \vspace{10pt}

    \begin{tabularx}{\textwidth}{r >{\centering\arraybackslash}X}
        \multicolumn{2}{c}{\small\textbf{Scene 2}} \\[2pt]
        \rotatebox[origin=c]{90}{\footnotesize \textbf{Baseline}} &
        \adjustbox{valign=c}{\includegraphics[width=\linewidth]{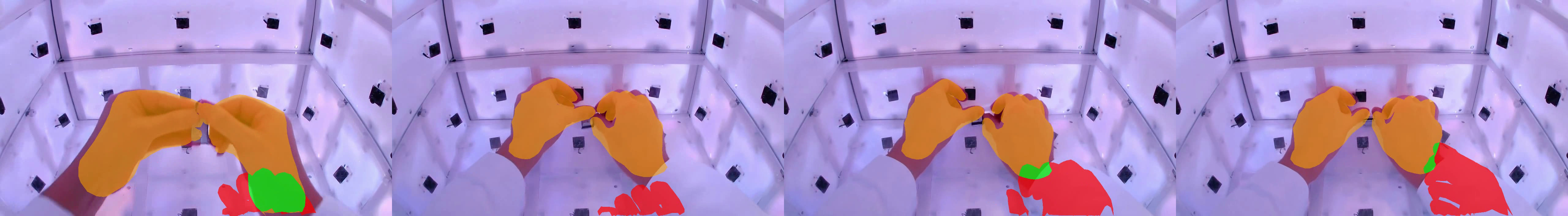}} \\[4pt]
        \rotatebox[origin=c]{90}{\footnotesize \textbf{HPP Cond.}} &
        \adjustbox{valign=c}{\includegraphics[width=\linewidth]{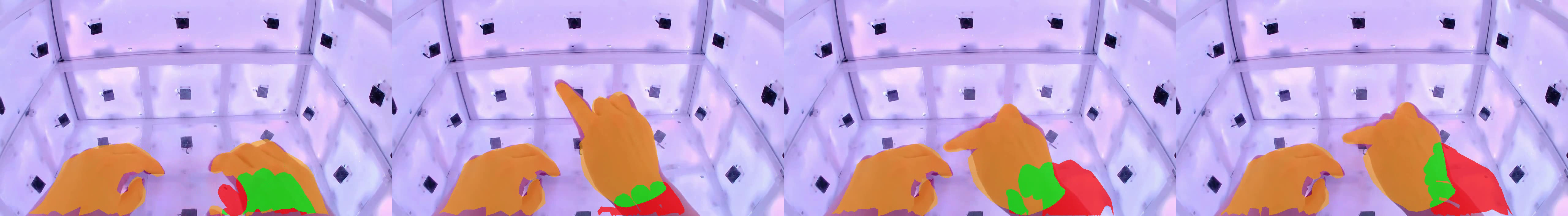}} \\[4pt]
        \rotatebox[origin=c]{90}{\footnotesize \textbf{Video Cond.}} &
        \adjustbox{valign=c}{\includegraphics[width=\linewidth]{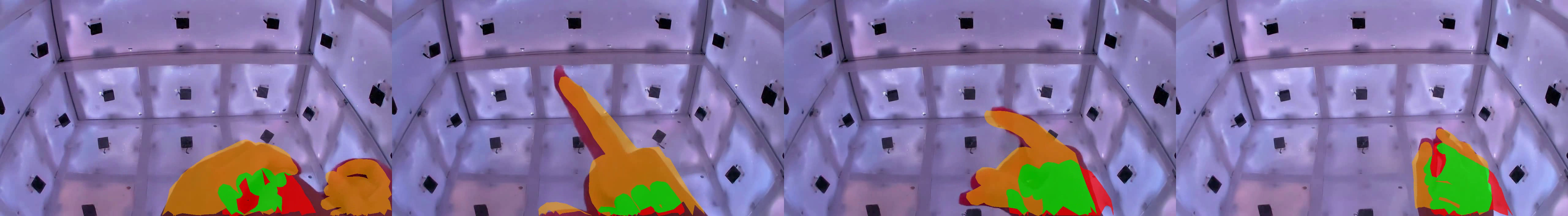}} \\[4pt]
        \rotatebox[origin=c]{90}{\footnotesize \textbf{Hybrid}} &
        \adjustbox{valign=c}{\includegraphics[width=\linewidth]{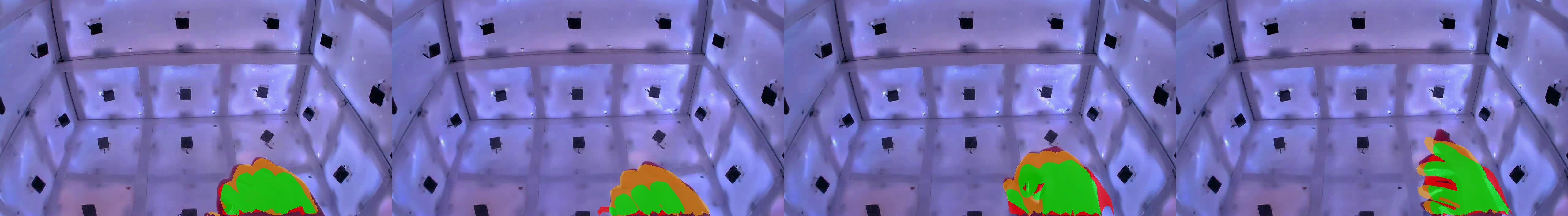}} \\
    \end{tabularx}

    \caption{\textbf{GigaHands qualitative comparison (1/2).}
    Qualitative comparison of hand-pose conditioning strategies on the GigaHands dataset. Ground-truth conditioning hand input is shown in red. Predicted hands are orange; overlap is green. Our hybrid conditioning strategy continues to outperform.}
    \label{fig:qualitative_gigahands_1}
\end{figure*}

\begin{figure*}[t!]
    \centering
    \newcommand{\figpath}{figs/qualitative_gigahands}
    \setlength{\tabcolsep}{2pt}
    \begin{tabularx}{\textwidth}{r >{\centering\arraybackslash}X}
        \multicolumn{2}{c}{\small\textbf{Scene 3}} \\[2pt]
        \rotatebox[origin=c]{90}{\footnotesize \textbf{Baseline}} &
        \adjustbox{valign=c}{\includegraphics[width=\linewidth]{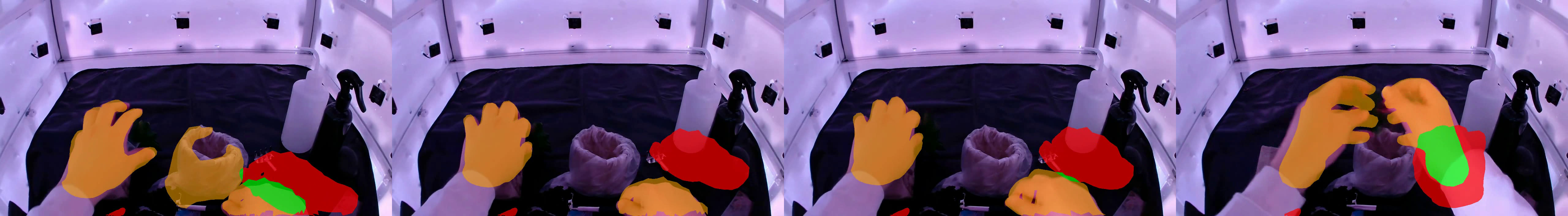}} \\[4pt]
        \rotatebox[origin=c]{90}{\footnotesize \textbf{HPP Cond.}} &
        \adjustbox{valign=c}{\includegraphics[width=\linewidth]{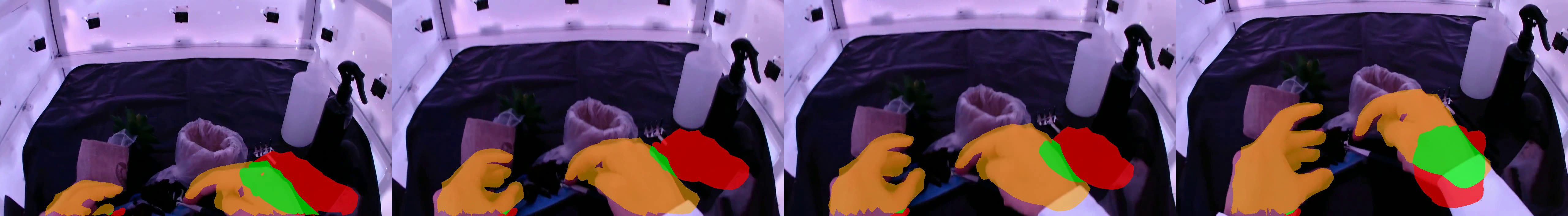}} \\[4pt]
        \rotatebox[origin=c]{90}{\footnotesize \textbf{Video Cond.}} &
        \adjustbox{valign=c}{\includegraphics[width=\linewidth]{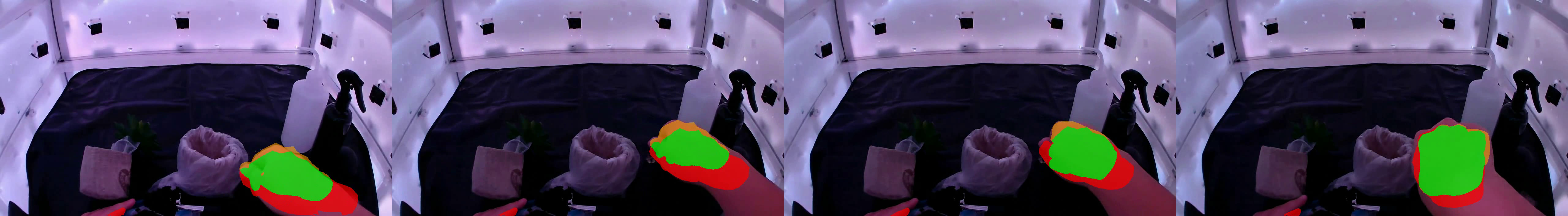}} \\[4pt]
        \rotatebox[origin=c]{90}{\footnotesize \textbf{Hybrid}} &
        \adjustbox{valign=c}{\includegraphics[width=\linewidth]{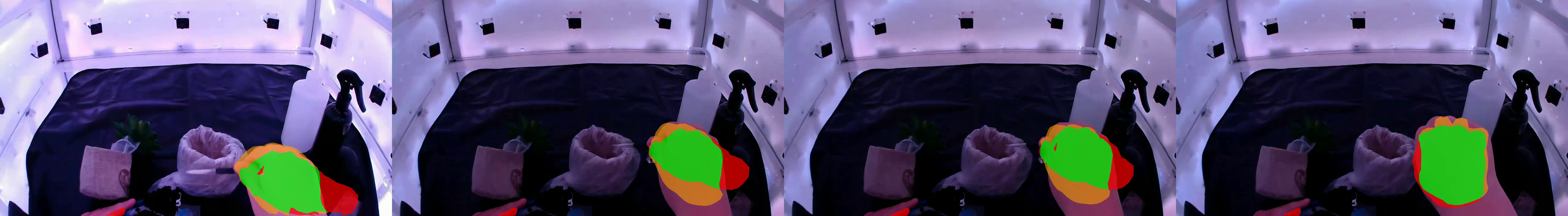}} \\
    \end{tabularx}

    \vspace{10pt}

    \begin{tabularx}{\textwidth}{r >{\centering\arraybackslash}X}
        \multicolumn{2}{c}{\small\textbf{Scene 4}} \\[2pt]
        \rotatebox[origin=c]{90}{\footnotesize \textbf{Baseline}} &
        \adjustbox{valign=c}{\includegraphics[width=\linewidth]{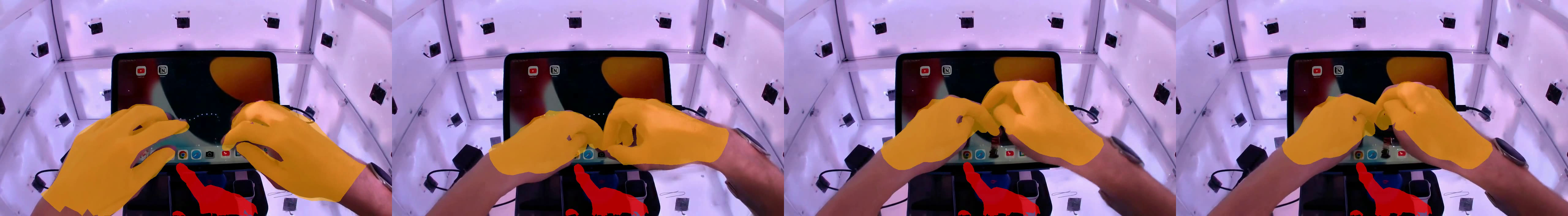}} \\[4pt]
        \rotatebox[origin=c]{90}{\footnotesize \textbf{HPP Cond.}} &
        \adjustbox{valign=c}{\includegraphics[width=\linewidth]{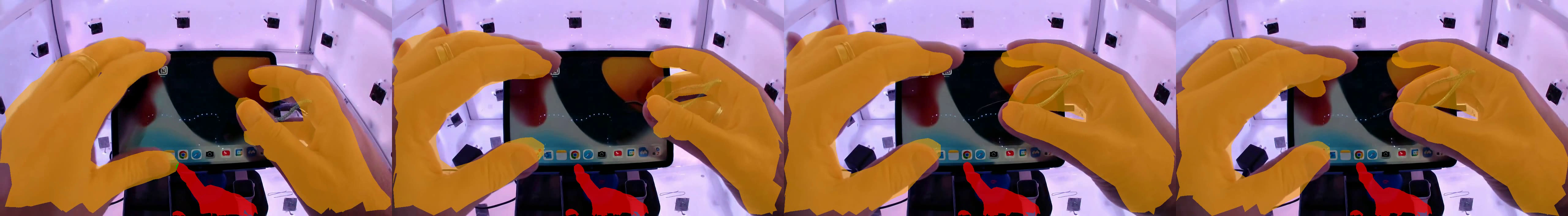}} \\[4pt]
        \rotatebox[origin=c]{90}{\footnotesize \textbf{Video Cond.}} &
        \adjustbox{valign=c}{\includegraphics[width=\linewidth]{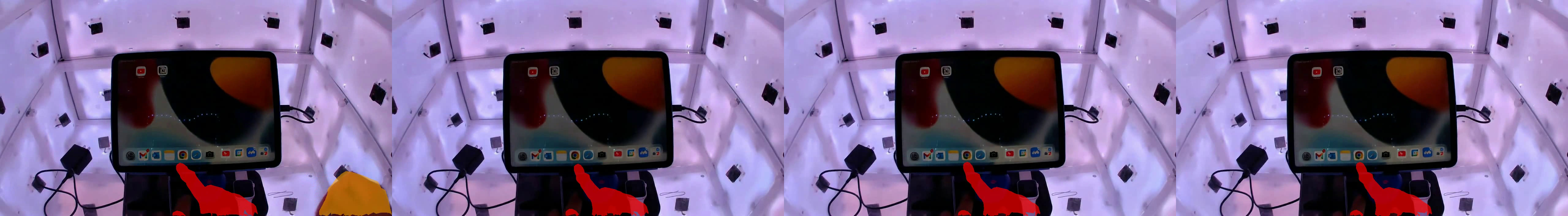}} \\[4pt]
        \rotatebox[origin=c]{90}{\footnotesize \textbf{Hybrid}} &
        \adjustbox{valign=c}{\includegraphics[width=\linewidth]{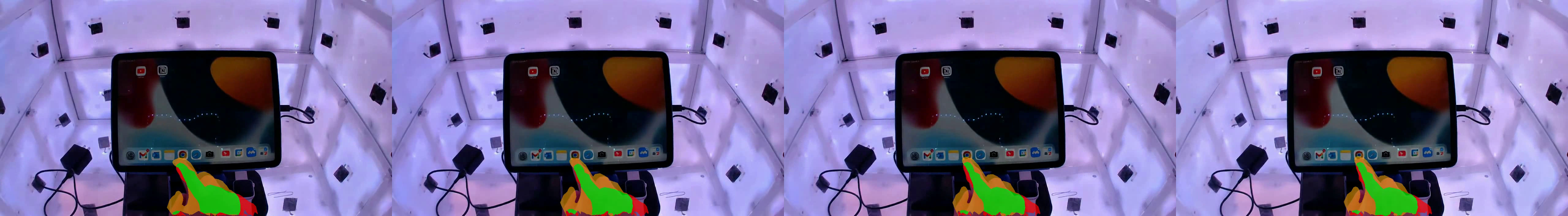}} \\
    \end{tabularx}

    \caption{\textbf{GigaHands qualitative comparison (2/2).}
    Qualitative comparison continued. Ground-truth conditioning hand input is shown in red. Predicted hands are orange; overlap is green.}
    \label{fig:qualitative_gigahands_2}
\end{figure*}

In addition to HOT3D, we evaluate our method on the larger GigaHands~\cite{gigahands} dataset ($8\times$ larger than HOT3D) with the Wan2.2 5B model. As shown in Table~\ref{tab:gigahands}, we continue to yield consistent improvements over the baselines; particularly, our 2D--3D hybrid conditioning outperforms 2D only conditioning, reducing MPJPE by 10\%, MPVPE by 11\%, and 2D error by 34\%. These results indicate scalability to larger, more complex data and richer hand motions. Fig.~\ref{fig:qualitative_gigahands_1} and~\ref{fig:qualitative_gigahands_2} provide additional qualitative comparisons across four scenes from the GigaHands dataset.

\begin{table}
    \centering
    \small
    \caption{\textbf{GigaHands ablation.} Additional hand pose accuracy ablations with Wan2.2 5B, trained on the GigaHands dataset. Hybrid conditioning continues to improve over 2D-only conditioning as dataset scale increases.}
    \begin{tabular}{c|ccc}
        \toprule
        \textbf{Method} & \textbf{MPJPE$\downarrow$} & \textbf{MPVPE$\downarrow$} & \textbf{L2Err$\downarrow$} \\
        \midrule
        Ground-truth            & 16.41 & 11.03 & 59.38 \\
        \midrule
        Baseline      & 20.86 & 15.08 & 268.49 \\
        3D Cond.      & 20.63 & 14.90 & 250.79 \\
        2D Cond.      & \rd{19.67} & \rd{14.03} & \rd{134.77} \\
        Hybrid Cond.  & \fs{17.78} & \fs{12.48} & \fs{89.59} \\
        \bottomrule
    \end{tabular}
    
    \label{tab:gigahands}
\end{table}
\subsection{Text-to-Video Generation}

\begin{figure*}[t!]
    \centering

    \begin{minipage}[t]{\linewidth}
      \includegraphics[width=0.2\linewidth]{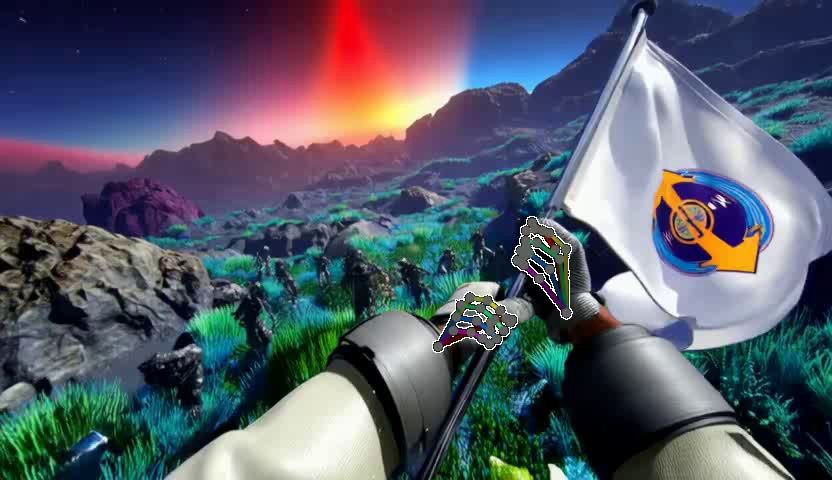}%
      \includegraphics[width=0.2\linewidth]{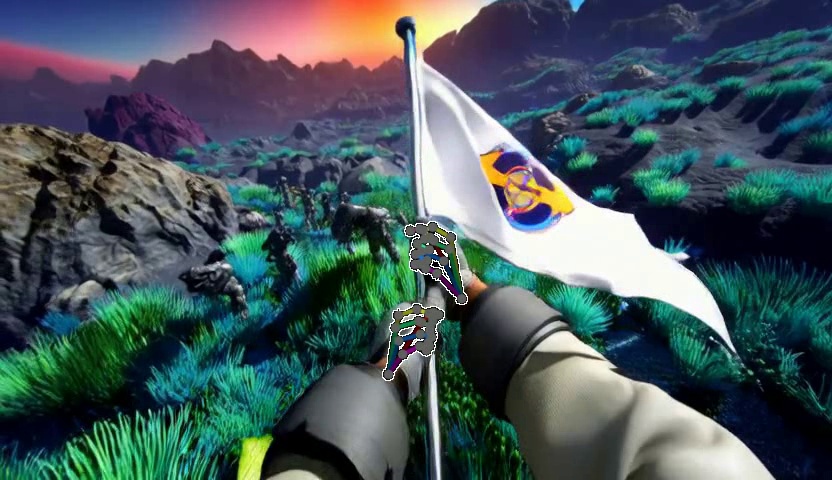}%
      \includegraphics[width=0.2\linewidth]{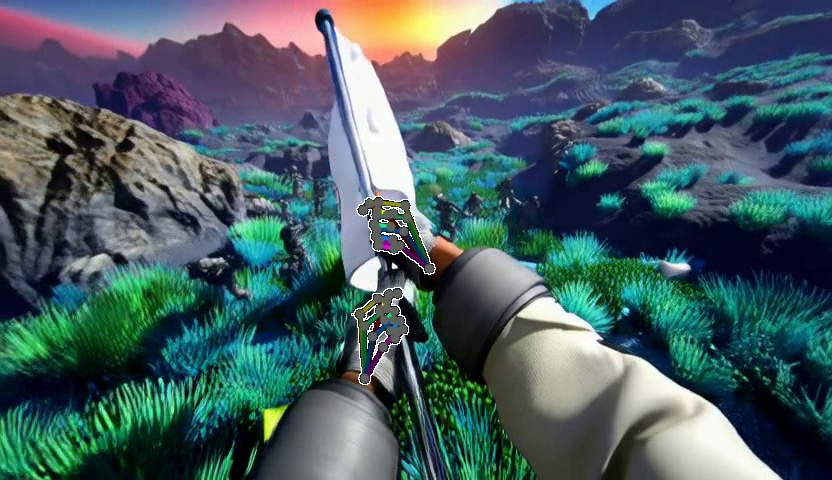}%
      \includegraphics[width=0.2\linewidth]{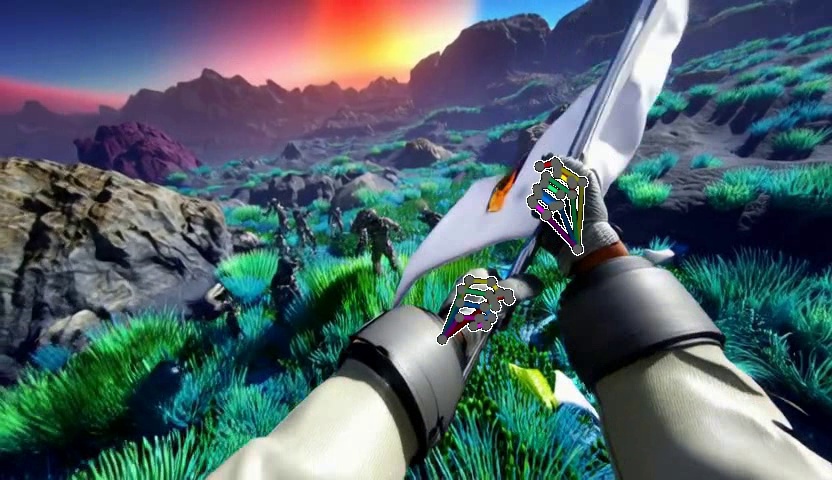}%
      \includegraphics[width=0.2\linewidth]{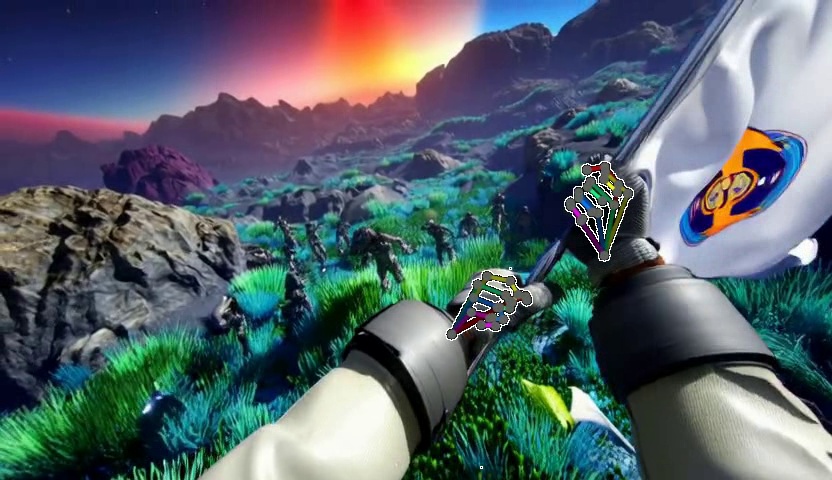}%
      
    \end{minipage}

    Wearing \textbf{astronaut gloves}, grips the shaft of \textbf{a waving flag}... a vibrant \textbf{alien landscape} under a colorful sky...
    \smallskip

    \begin{minipage}[t]{\linewidth}
      \includegraphics[width=0.2\linewidth]{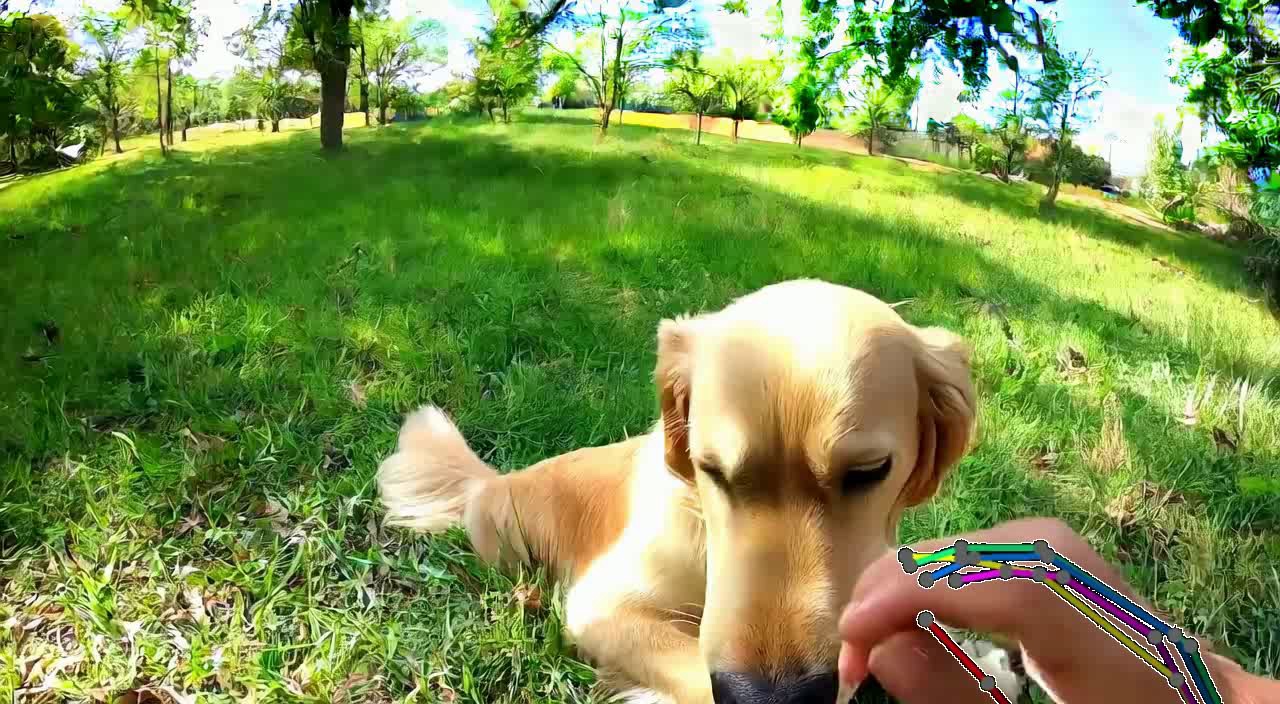}%
      \includegraphics[width=0.2\linewidth]{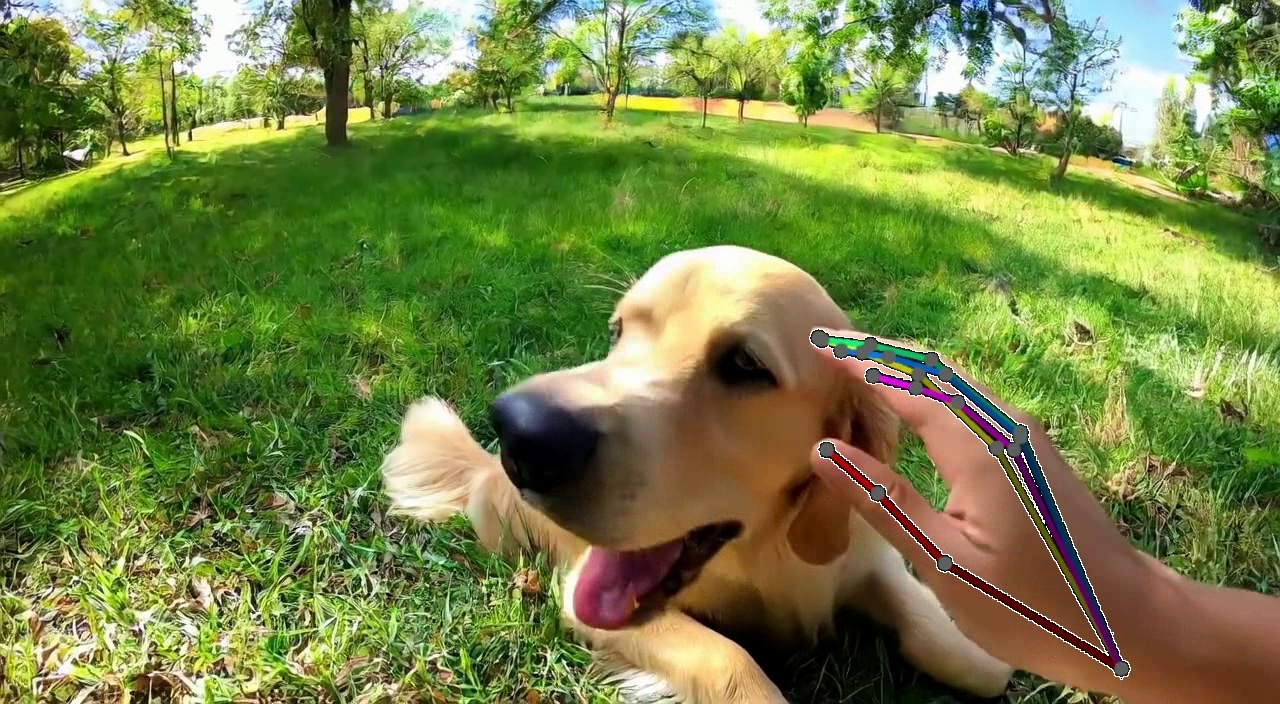}%
      \includegraphics[width=0.2\linewidth]{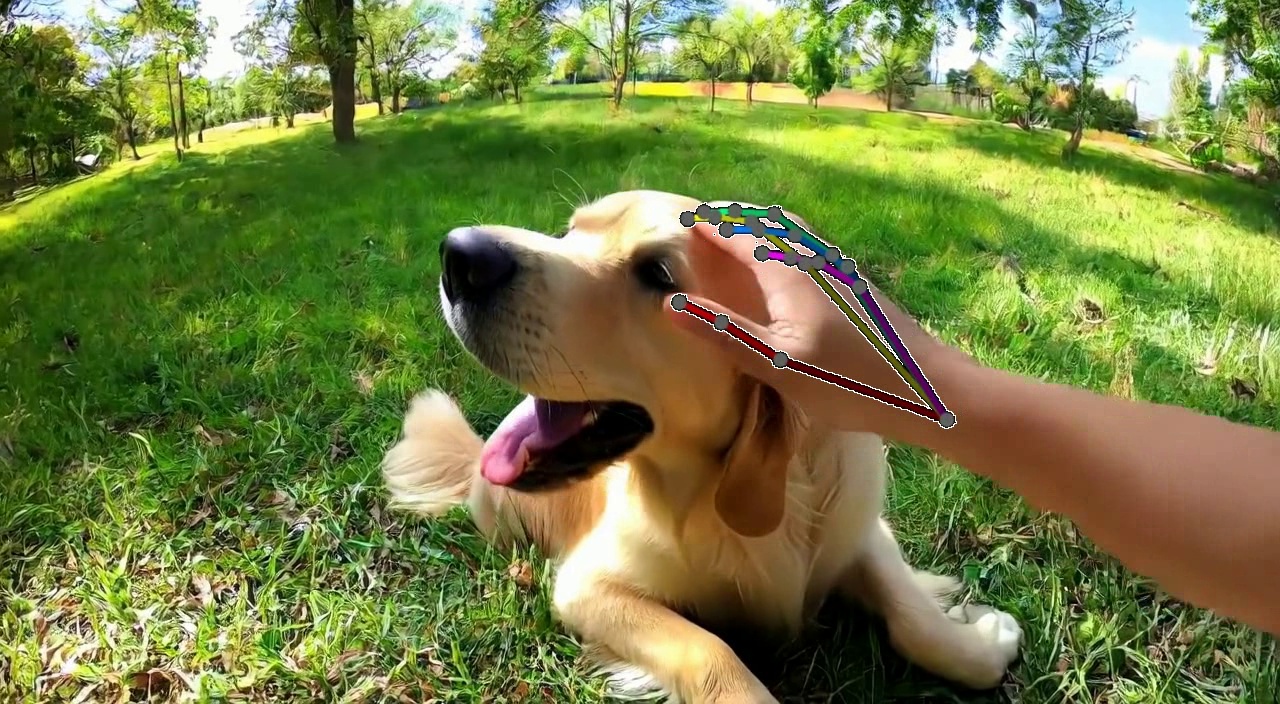}%
      \includegraphics[width=0.2\linewidth]{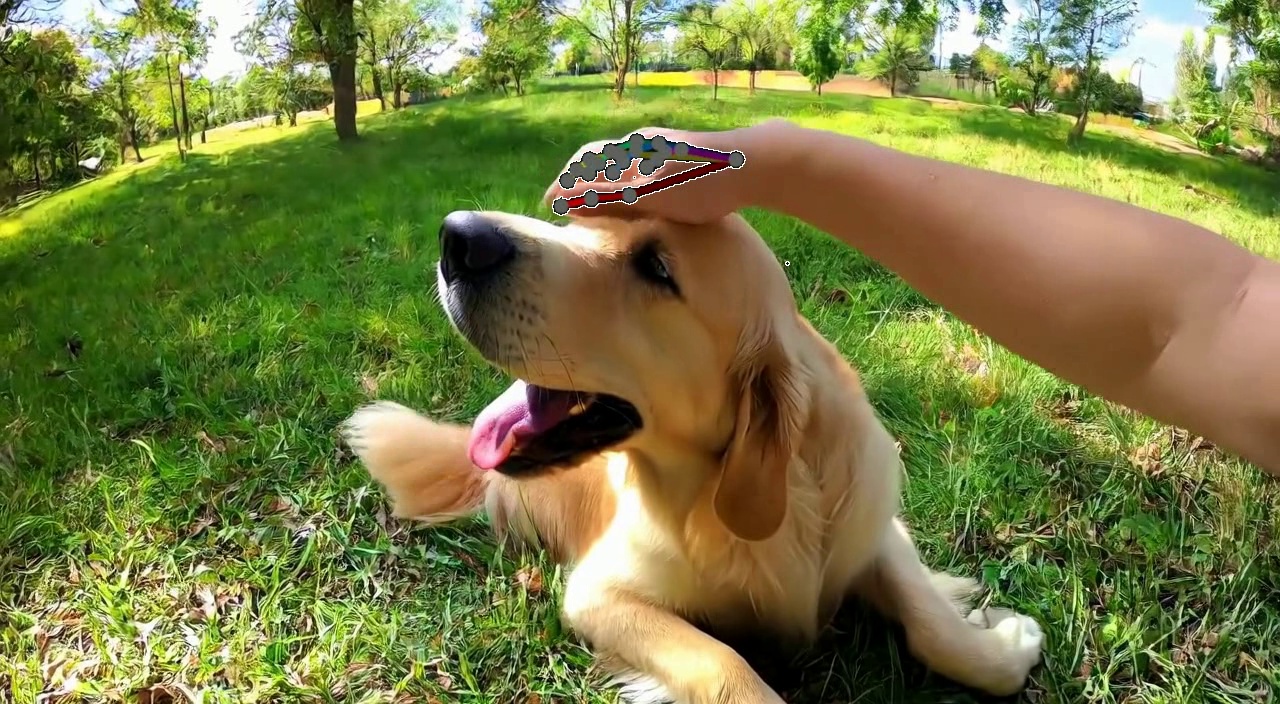}%
      \includegraphics[width=0.2\linewidth]{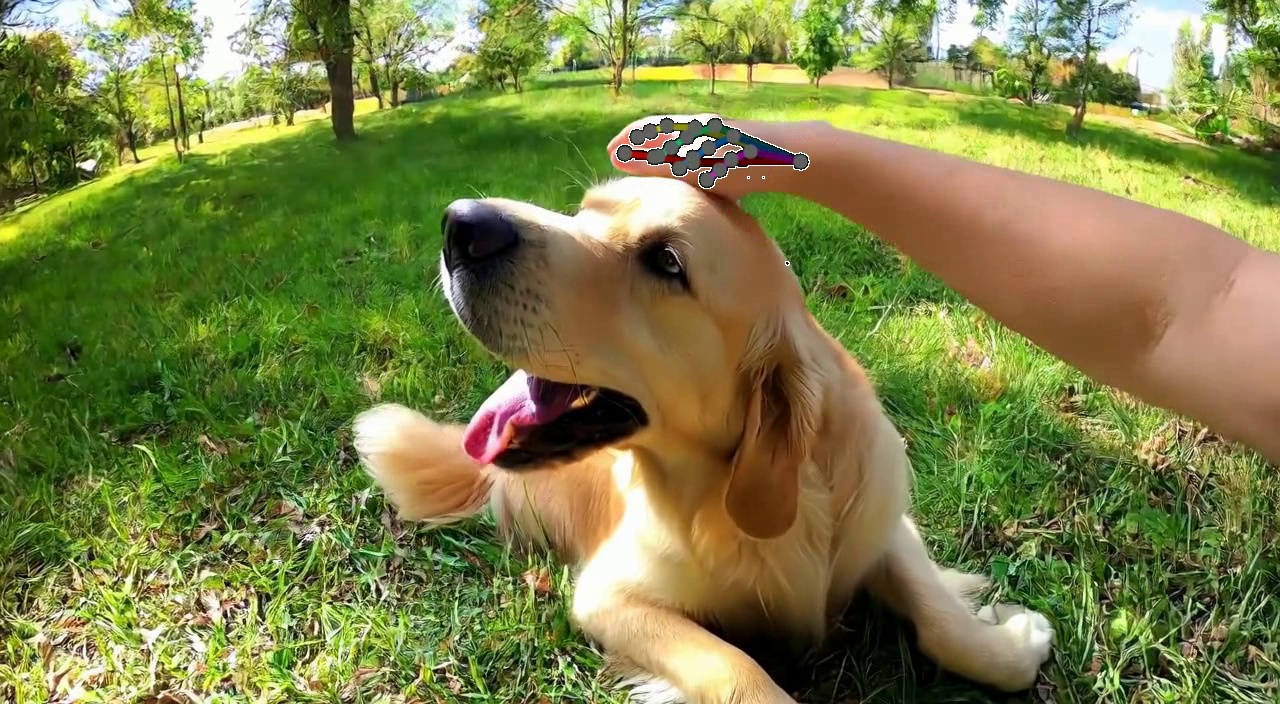}%
    \end{minipage}

    A bright \textbf{outdoor park} on a clear day... a friendly \textbf{golden retriever} sits obediently...
    \smallskip

    \begin{minipage}[t]{\linewidth}
      \includegraphics[width=0.2\linewidth]{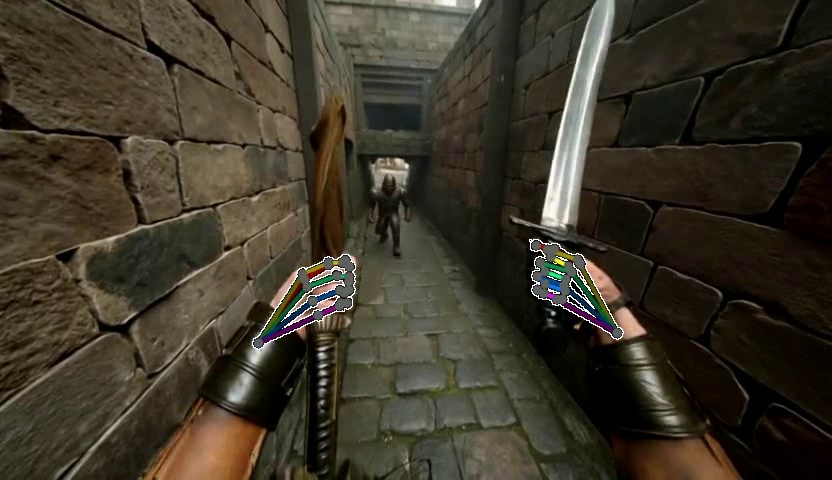}%
      \includegraphics[width=0.2\linewidth]{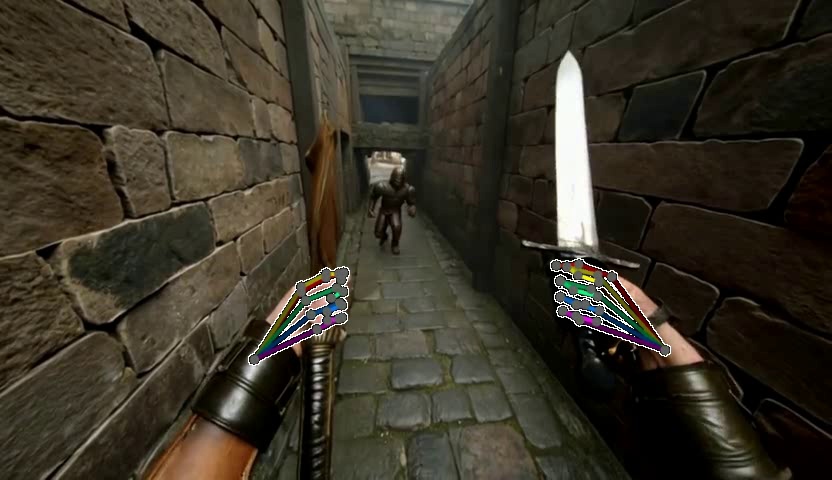}%
      \includegraphics[width=0.2\linewidth]{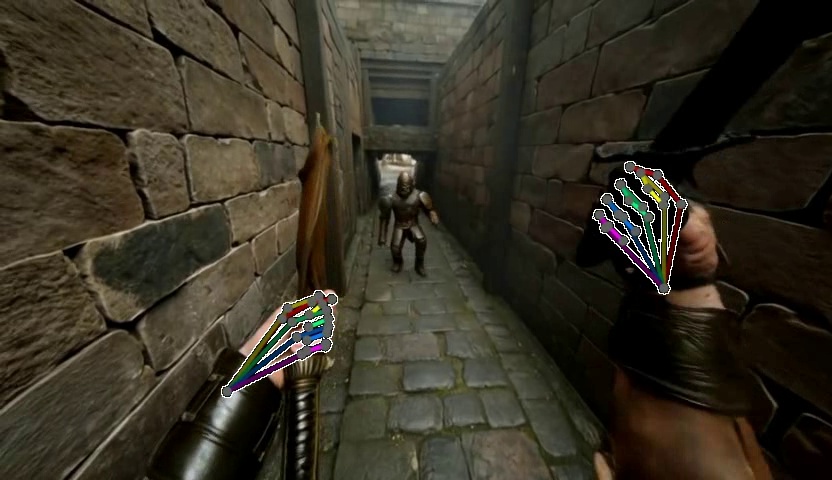}%
      \includegraphics[width=0.2\linewidth]{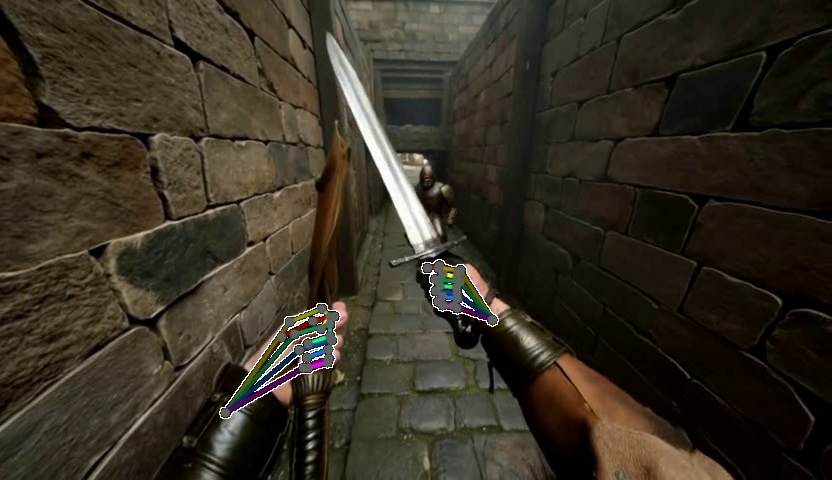}%
      \includegraphics[width=0.2\linewidth]{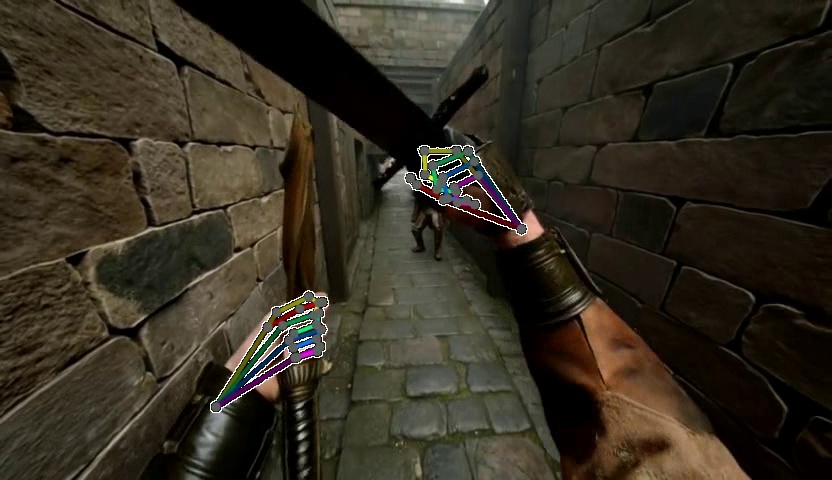}%
    \end{minipage}

    A gritty \textbf{medieval dungeon}... the right hand wields a \textbf{steel longsword}... an \textbf{armored soldier} charges towards the viewer...
    \smallskip
    
    \begin{minipage}[t]{\linewidth}
      \includegraphics[width=0.2\linewidth]{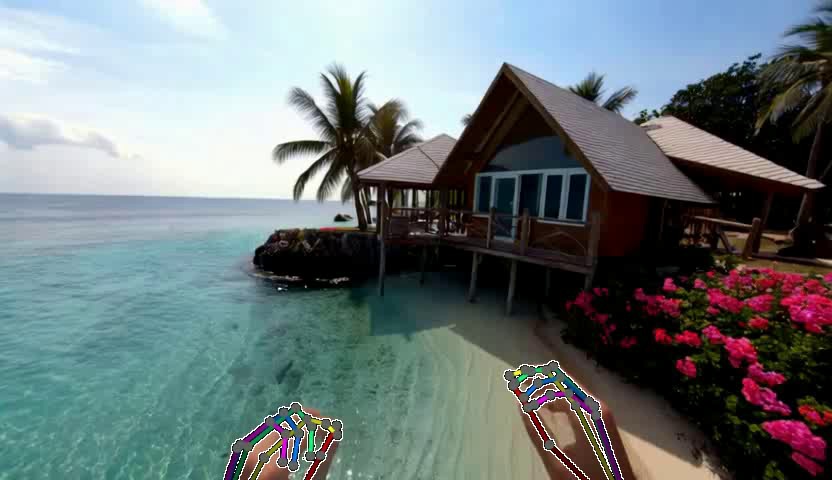}%
      \includegraphics[width=0.2\linewidth]{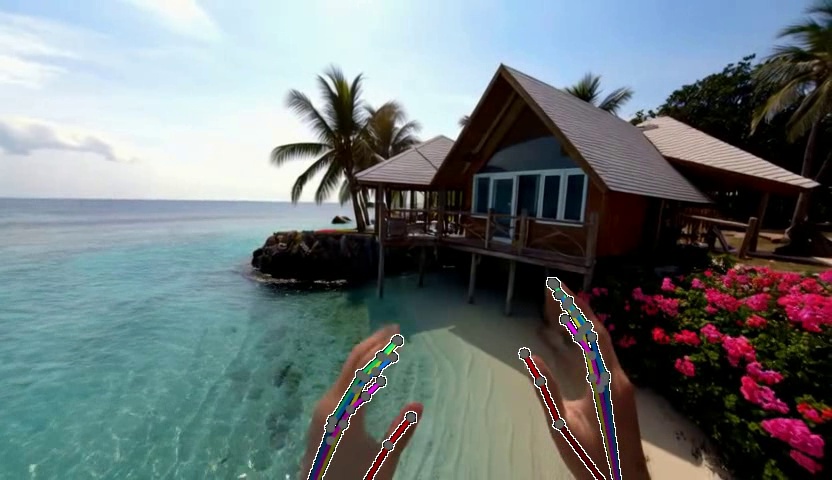}%
      \includegraphics[width=0.2\linewidth]{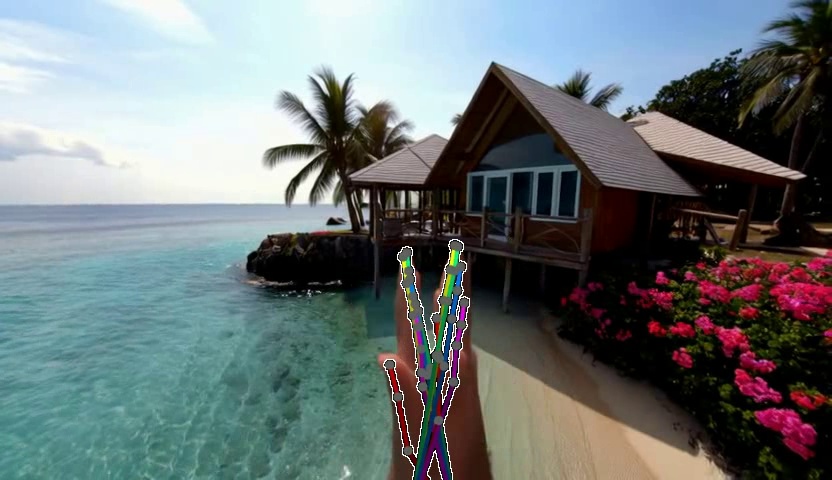}%
      \includegraphics[width=0.2\linewidth]{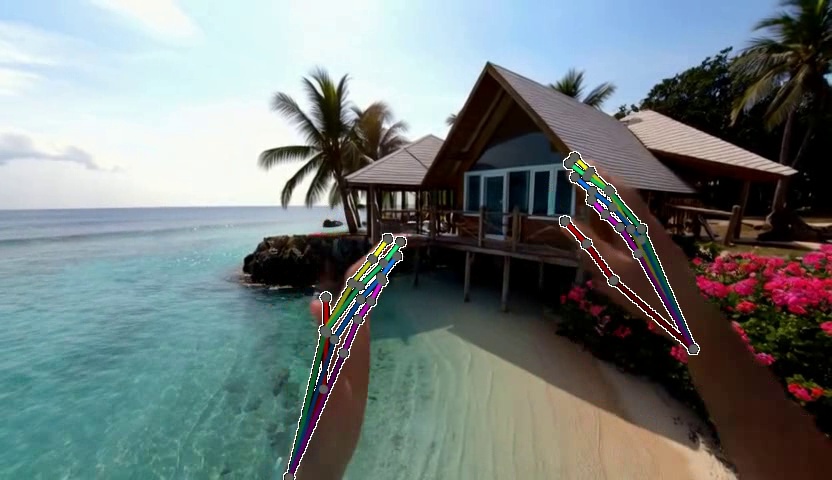}%
      \includegraphics[width=0.2\linewidth]{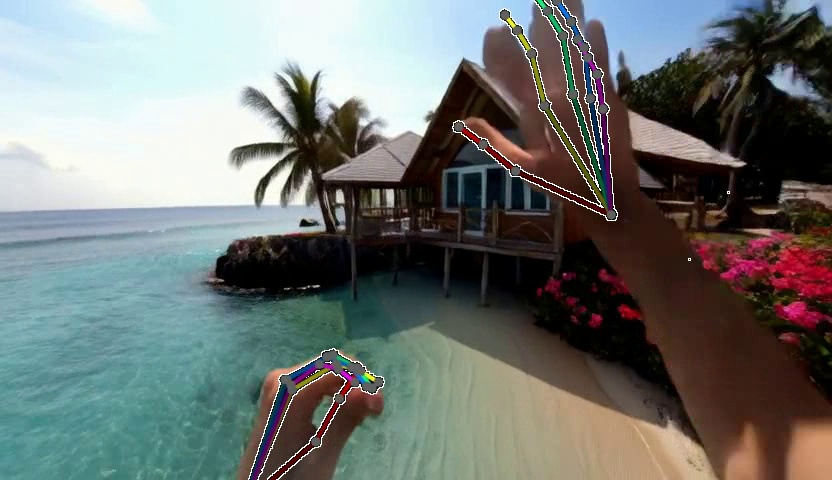}%
    \end{minipage}

    A quaint \textbf{A-frame cottage}... surrounded by turquoise waters and sandy beaches... palm trees sway in the background...
    \smallskip

    \begin{minipage}[t]{\linewidth}
      \includegraphics[width=0.2\linewidth]{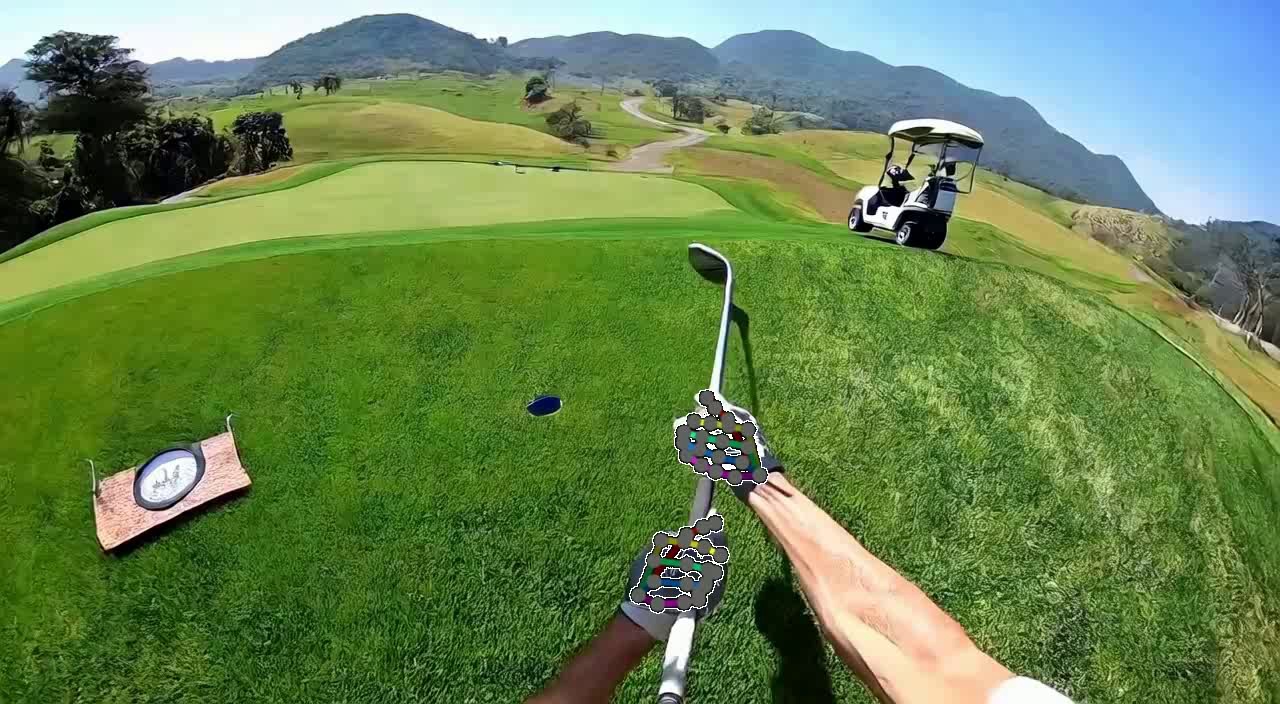}%
      \includegraphics[width=0.2\linewidth]{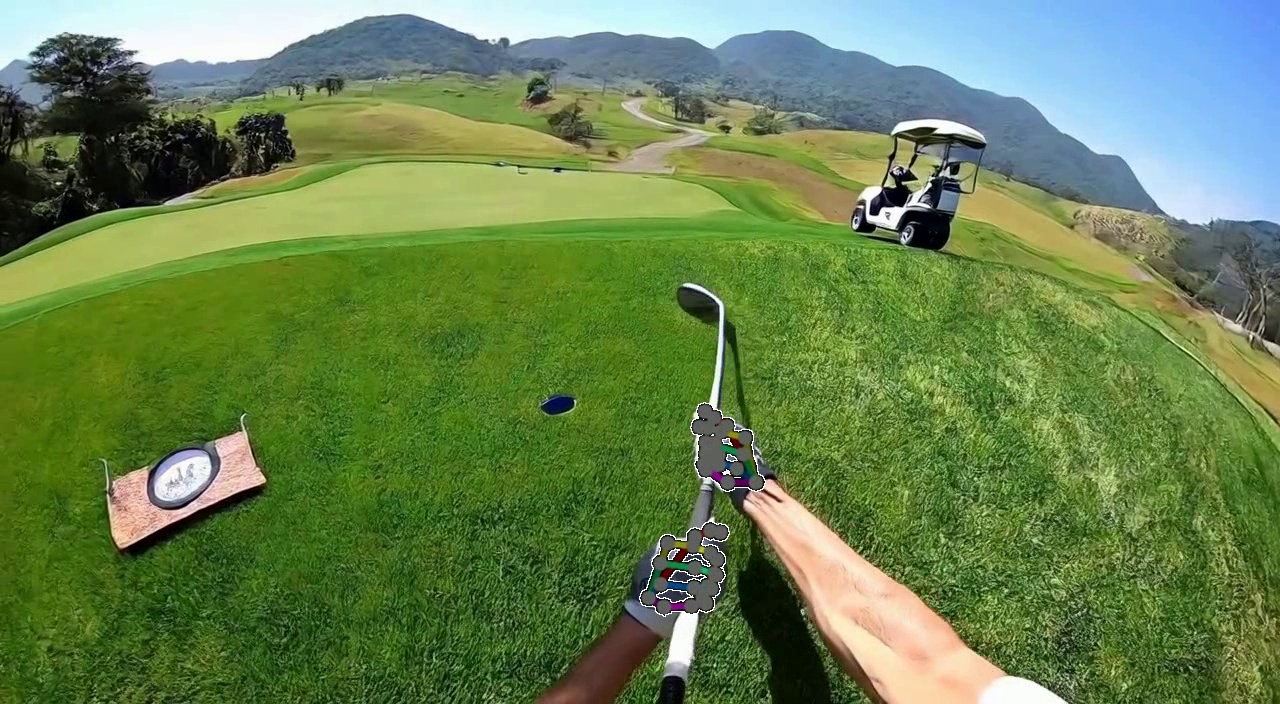}%
      \includegraphics[width=0.2\linewidth]{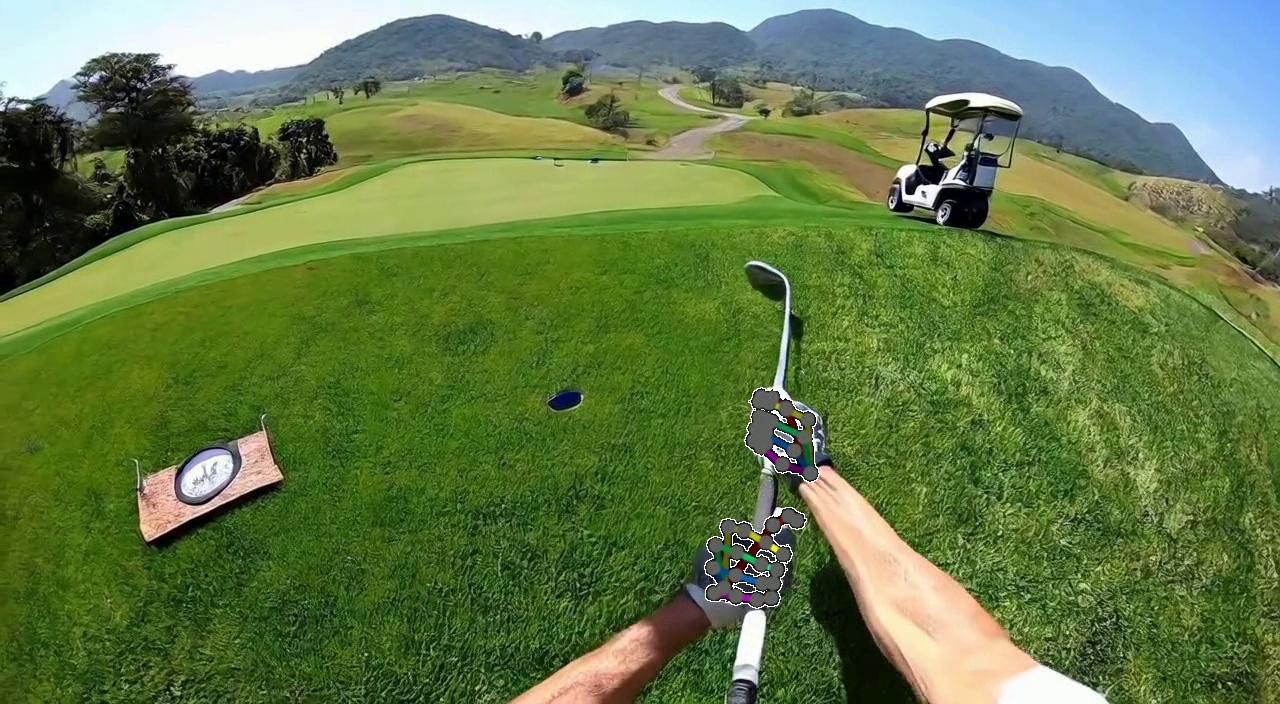}%
      \includegraphics[width=0.2\linewidth]{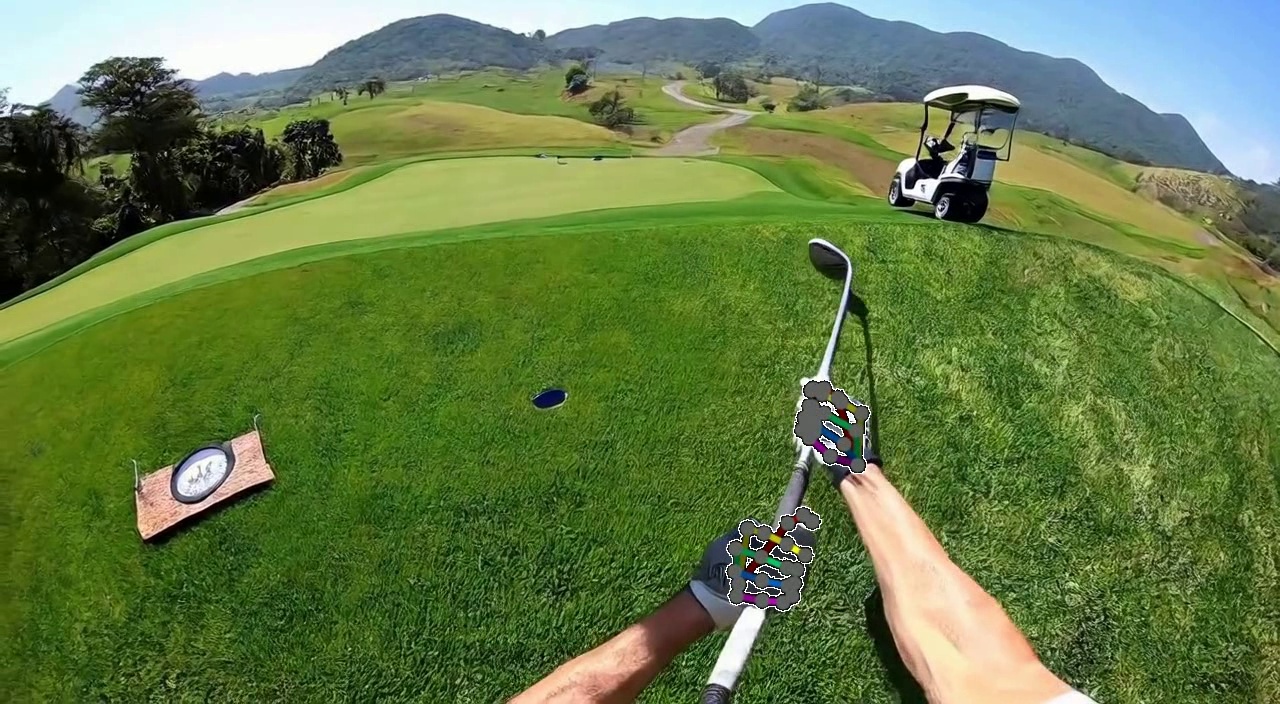}%
      \includegraphics[width=0.2\linewidth]{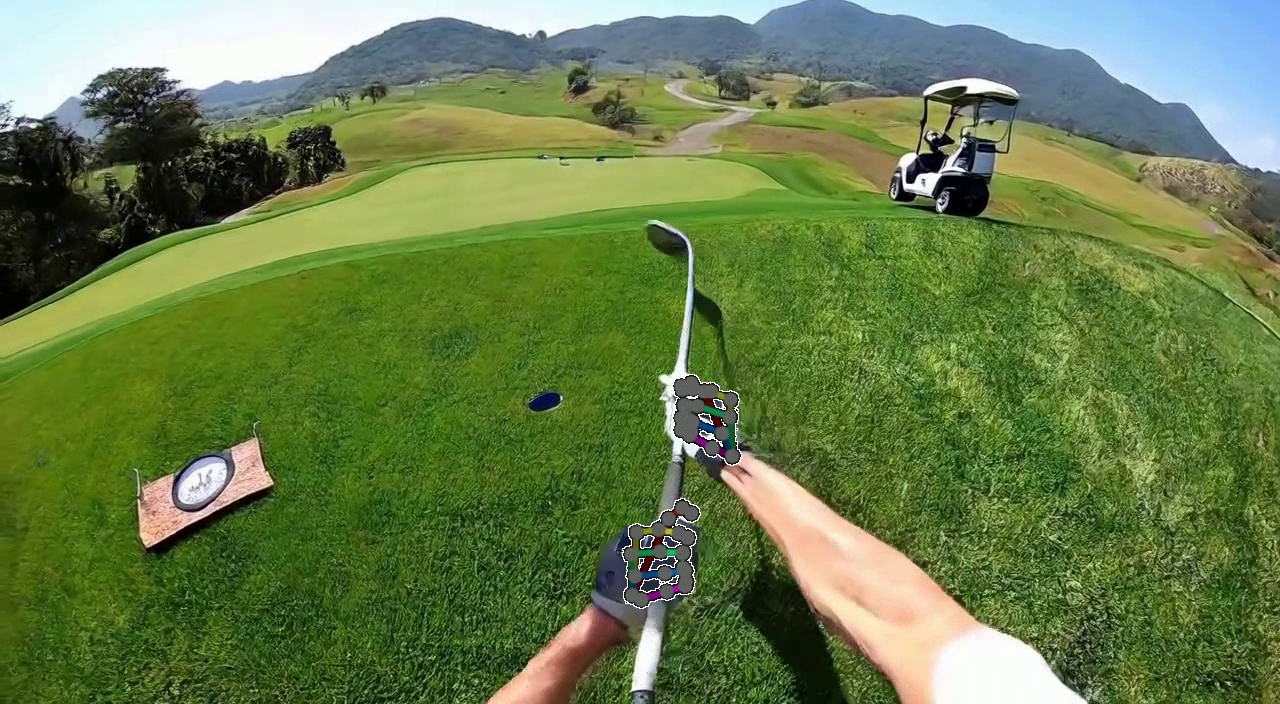}%
    \end{minipage}

    A lush green \textbf{golf course} on a sunny day... hands are \textbf{swinging a golf club}.... a golf buggy and a caddy stand ready...
    \smallskip
    
    \begin{minipage}[t]{\linewidth}
      \includegraphics[width=0.2\linewidth]{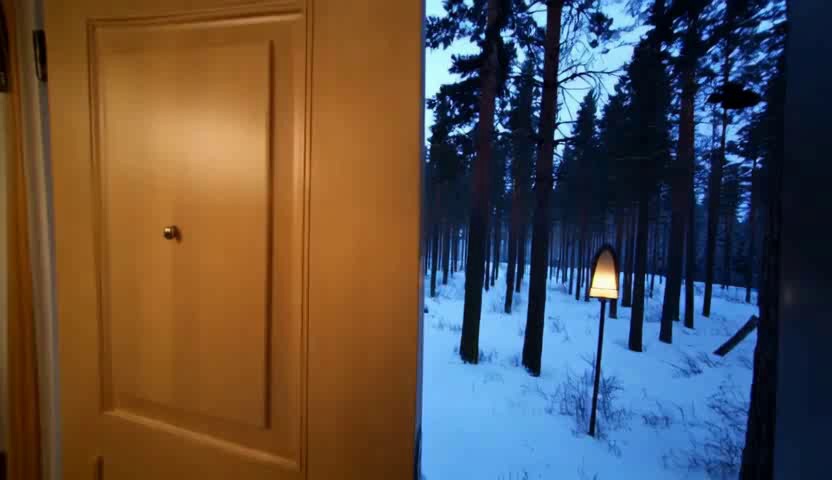}%
      \includegraphics[width=0.2\linewidth]{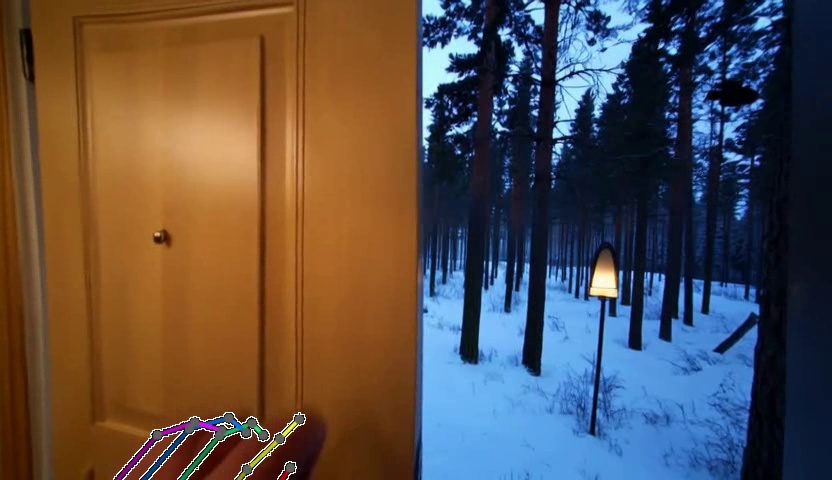}%
      \includegraphics[width=0.2\linewidth]{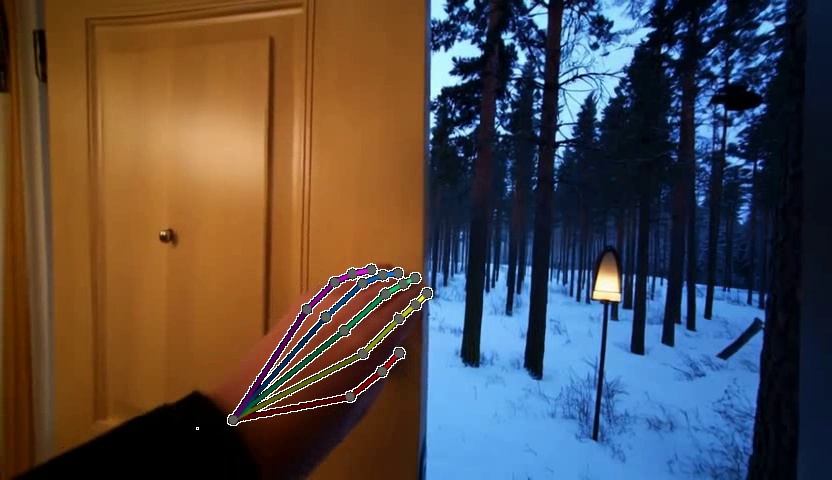}%
      \includegraphics[width=0.2\linewidth]{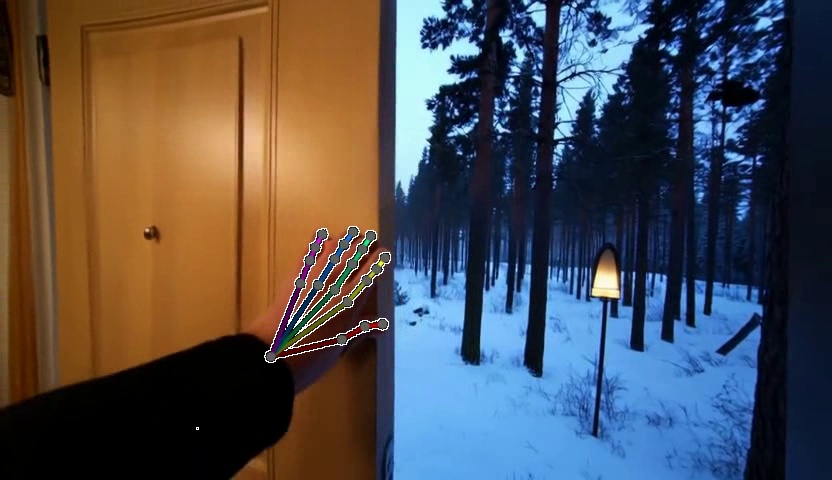}%
      \includegraphics[width=0.2\linewidth]{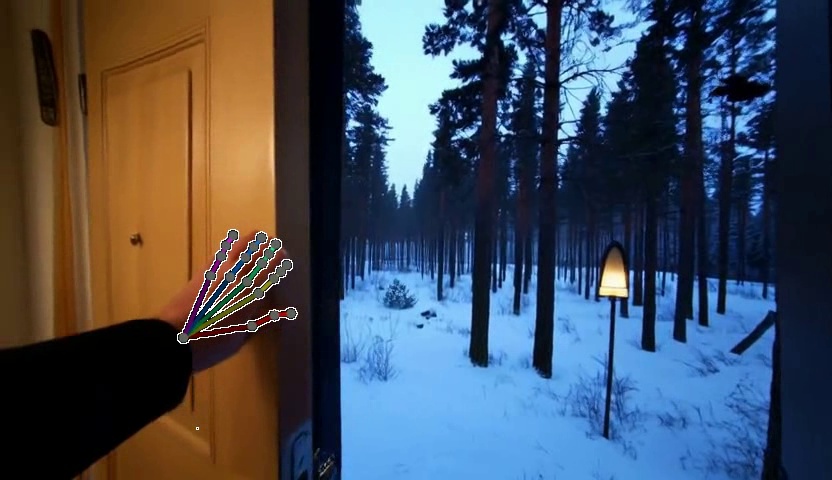}%
    \end{minipage}

    \textbf{Pushes the wooden door} open... revealing a magical \textbf{winter forest}.. a vintage lamppost glows warmly...
    \smallskip

    \begin{minipage}[t]{\linewidth}
      \includegraphics[width=0.2\linewidth]{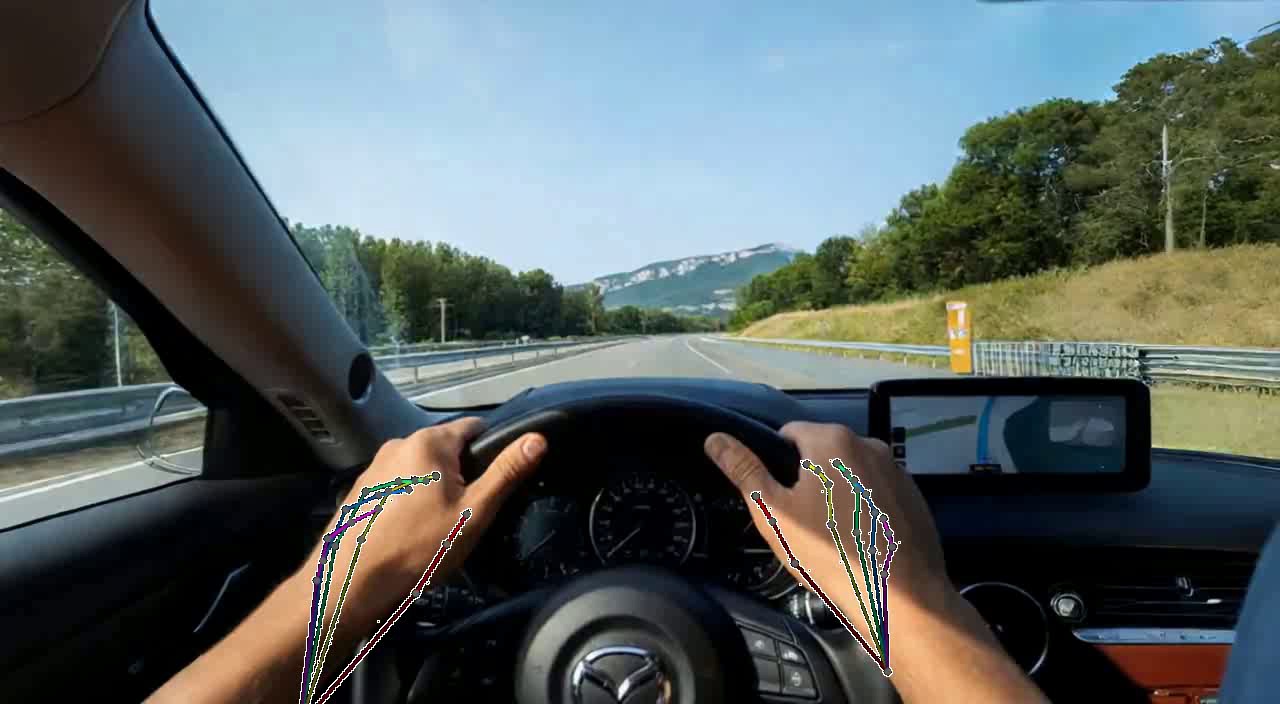}%
      \includegraphics[width=0.2\linewidth]{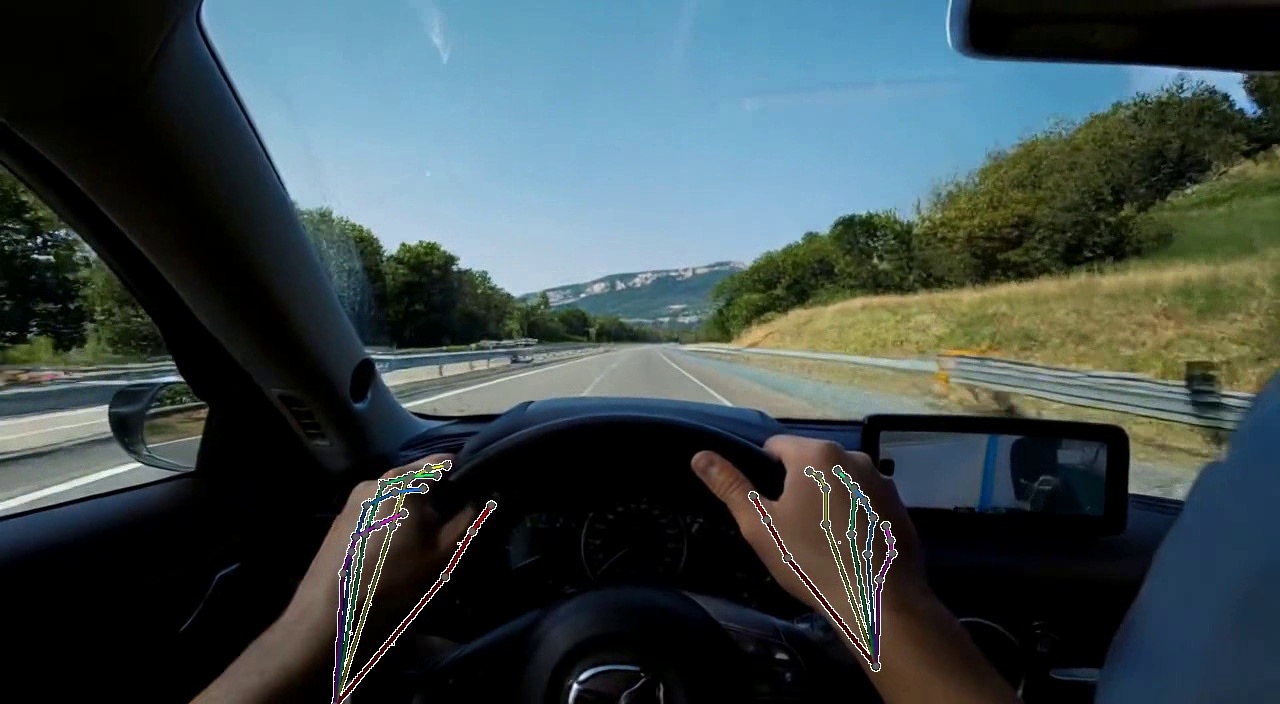}%
      \includegraphics[width=0.2\linewidth]{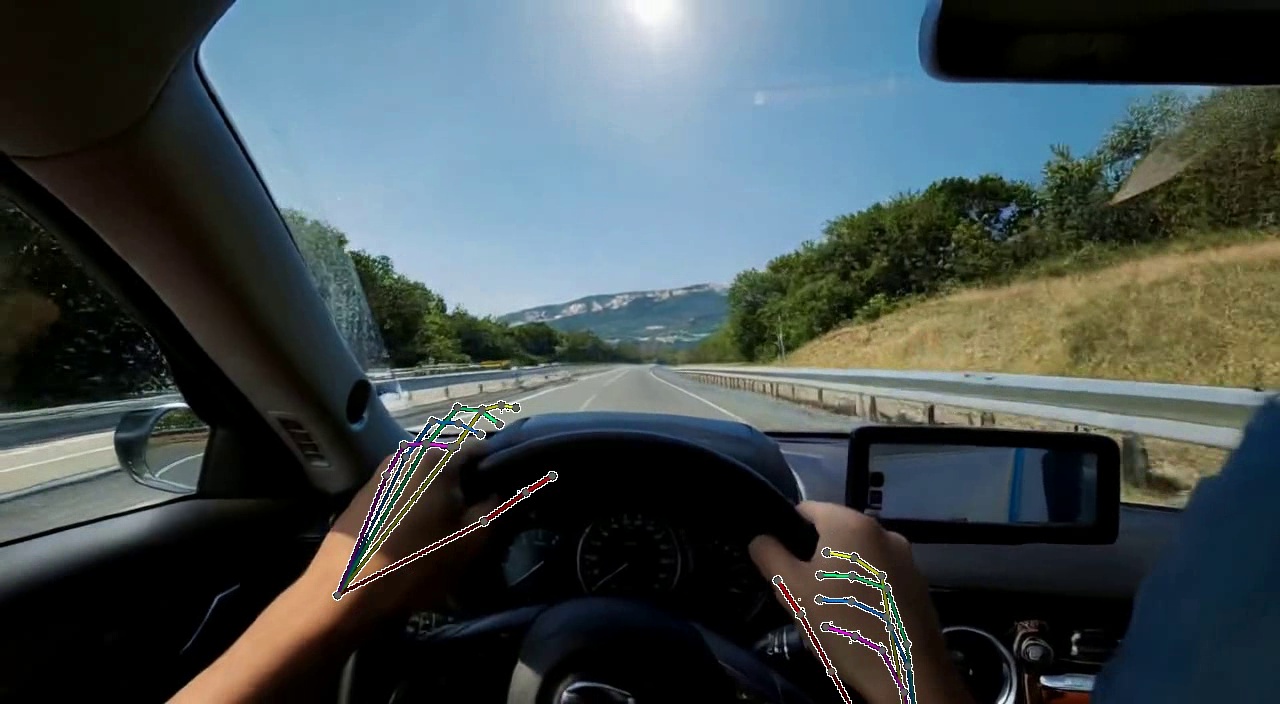}%
      \includegraphics[width=0.2\linewidth]{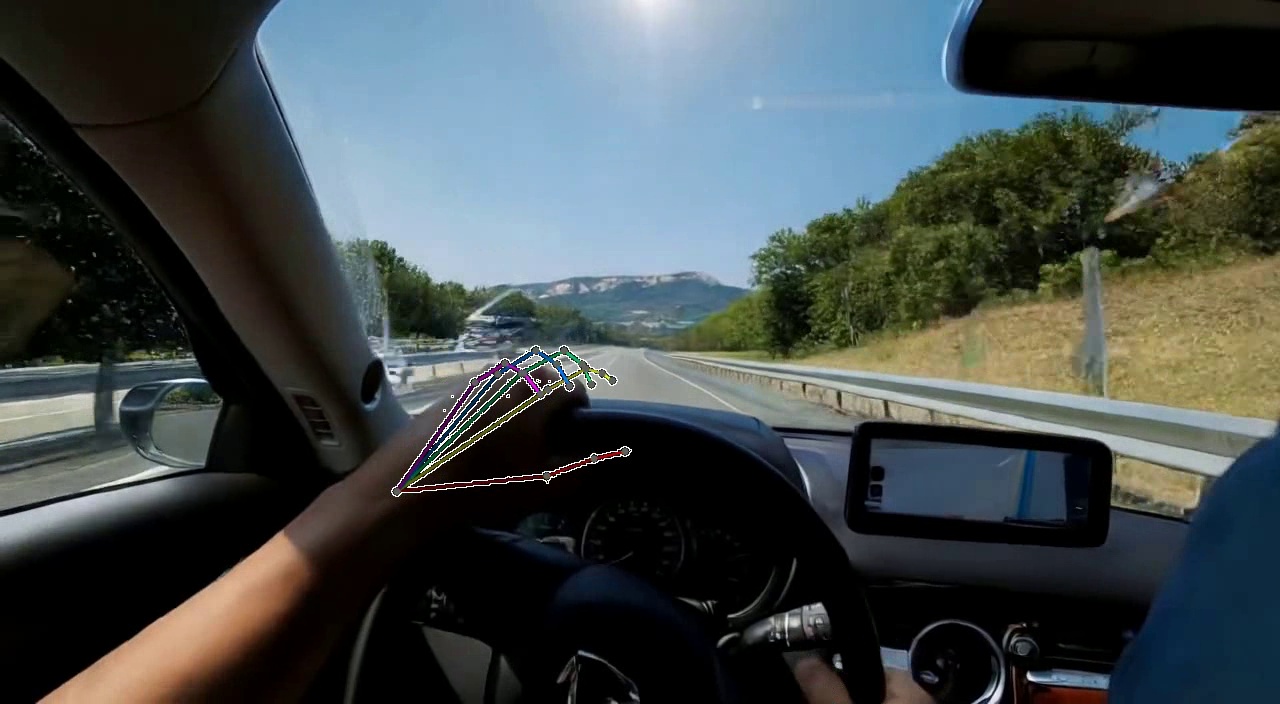}%
      \includegraphics[width=0.2\linewidth]{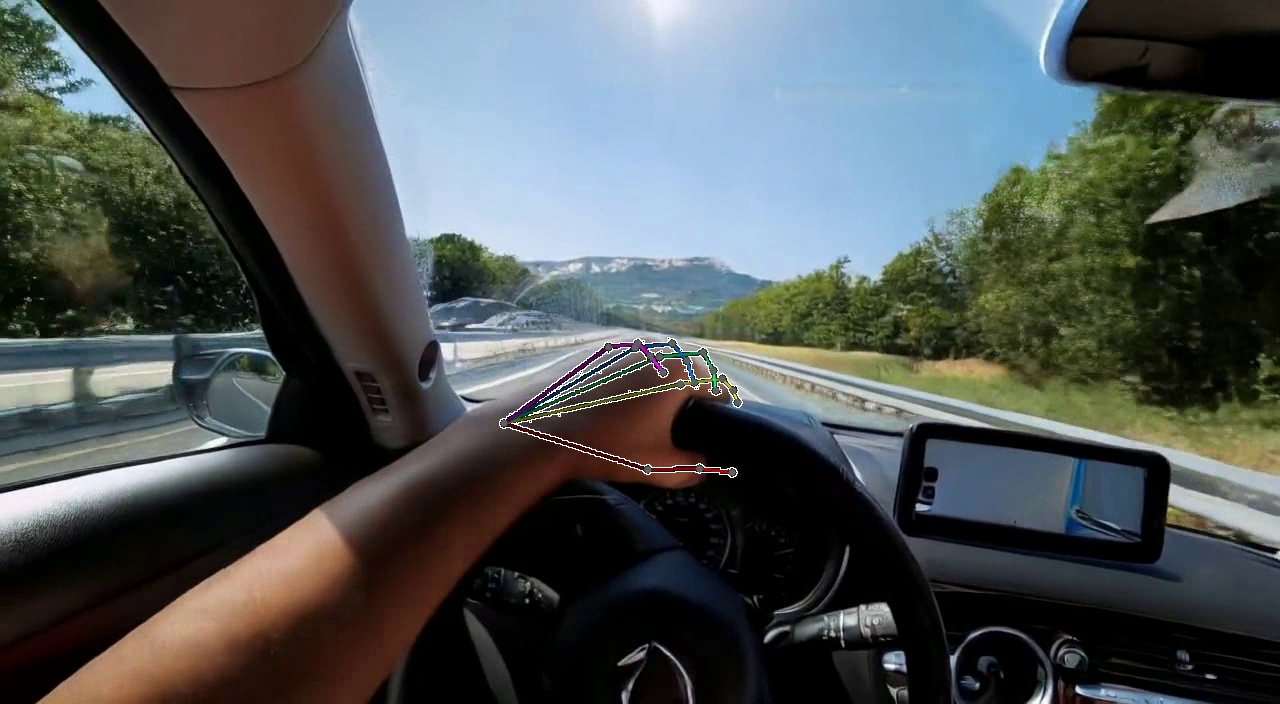}%
    \end{minipage}

     \textbf{Driving on a highway} in a modern car... a green countryside with trees on a bright clear summer day...
     \smallskip
    
    \begin{minipage}[t]{\linewidth}
      \includegraphics[width=0.2\linewidth]{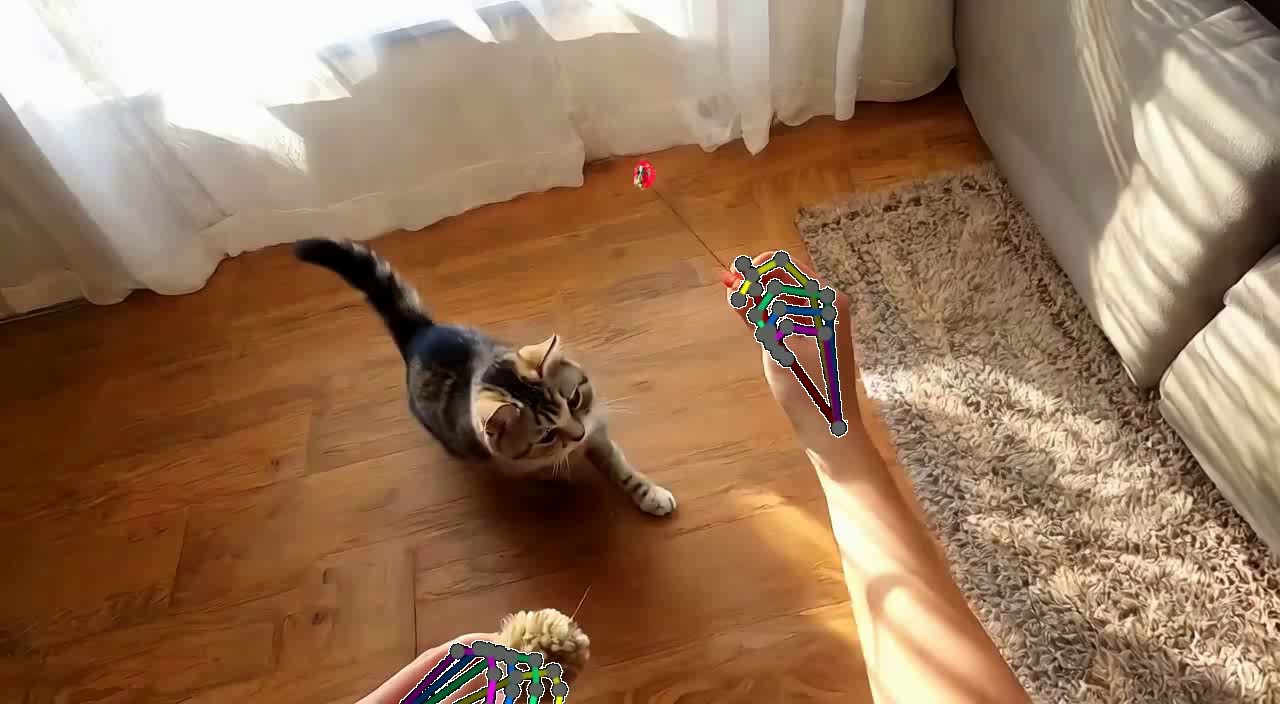}%
      \includegraphics[width=0.2\linewidth]{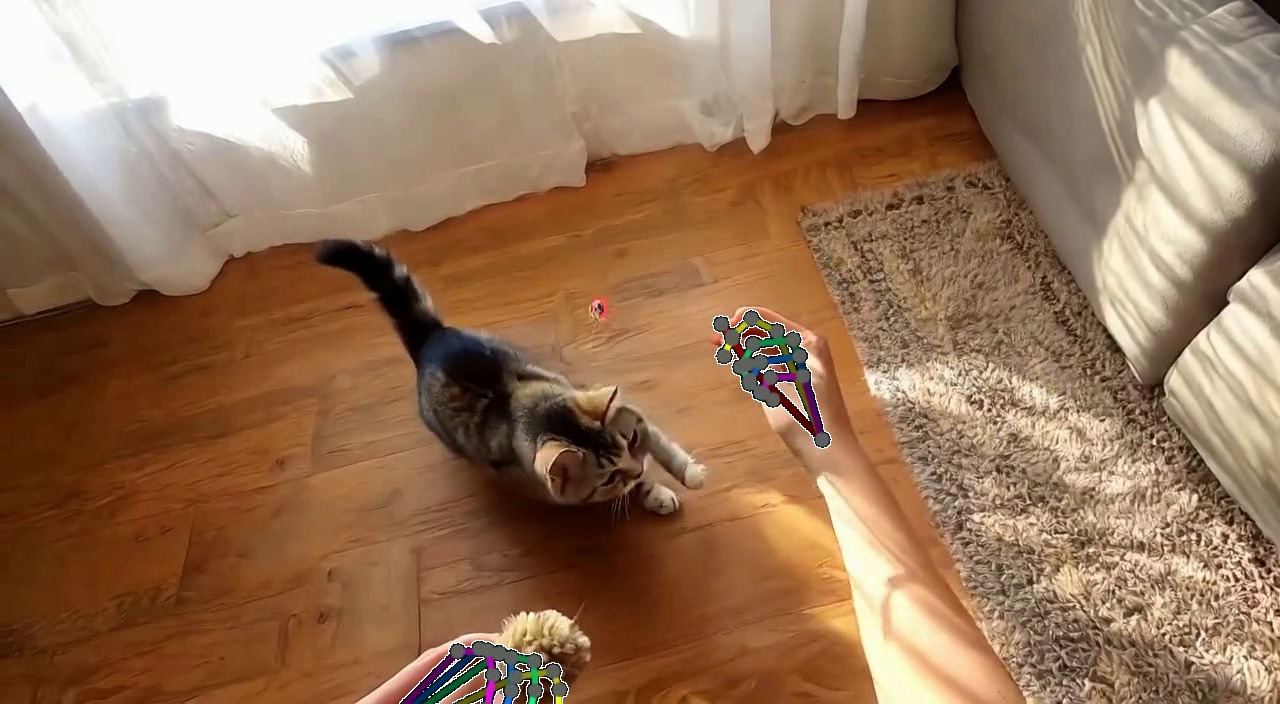}%
      \includegraphics[width=0.2\linewidth]{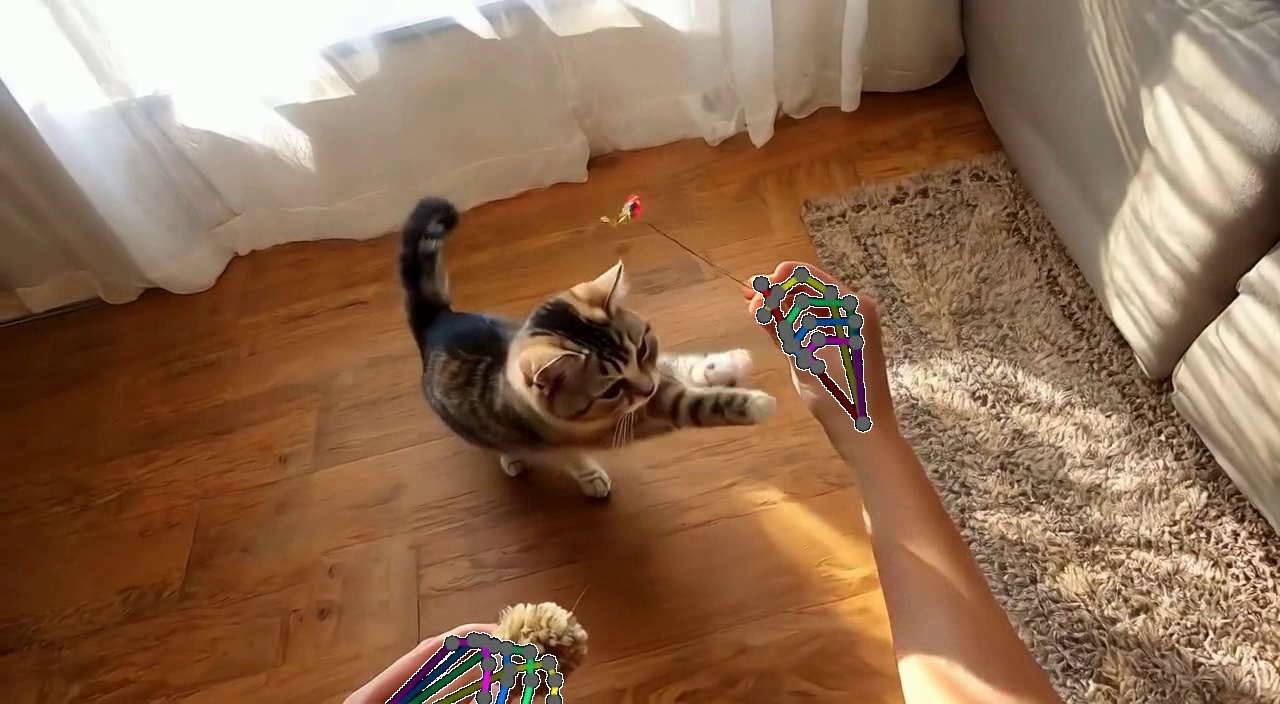}%
      \includegraphics[width=0.2\linewidth]{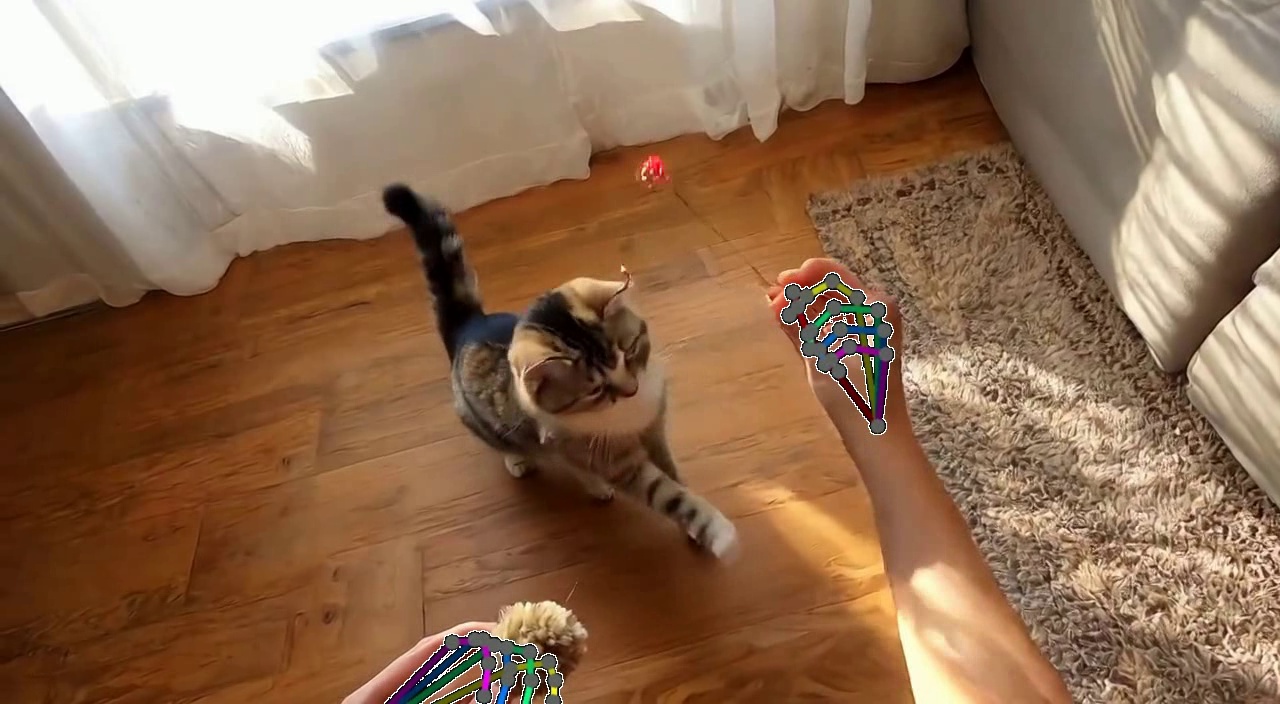}%
      \includegraphics[width=0.2\linewidth]{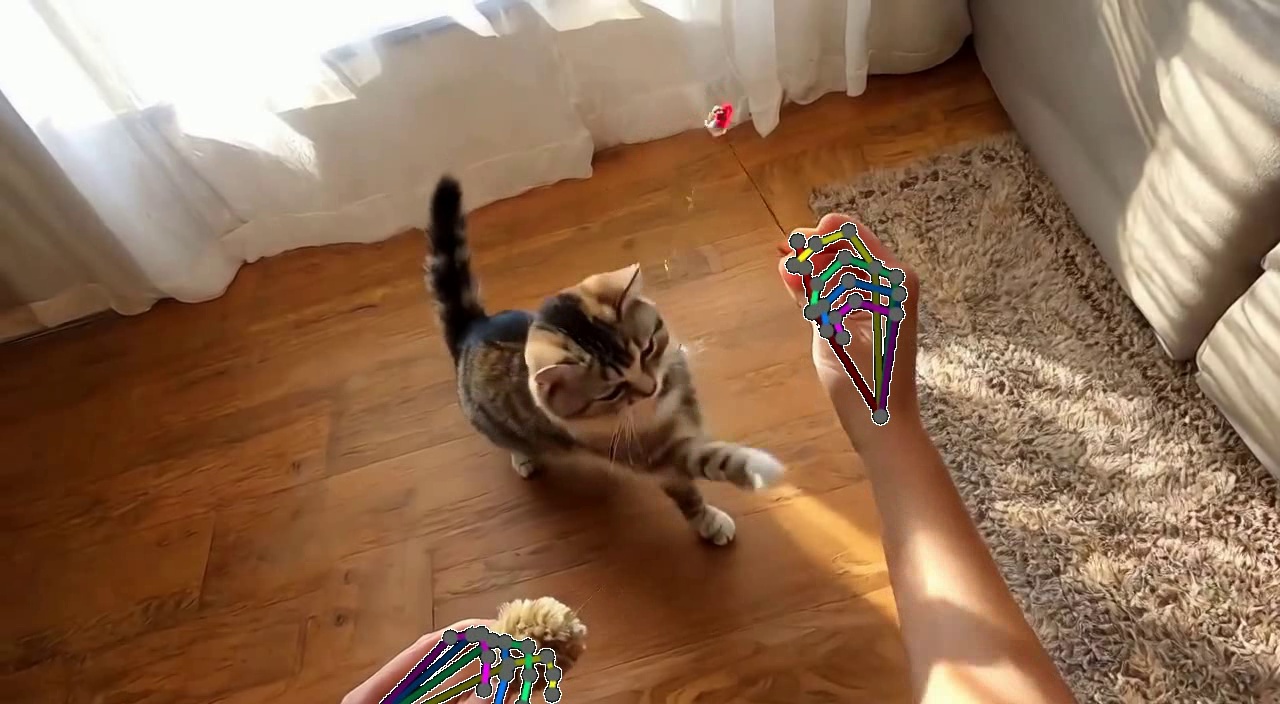}%
    \end{minipage}

     \textbf{Holding a cat wand toy} with a fuzzy pom-pom ball... a \textbf{playful domestic cat} swipes repeatedly at the fuzzball...

    \caption{\textbf{Diverse generations.} Leveraging the implicit world knowledge of foundation video models, our system generalizes to diverse scenarios with complex interactions. Generated videos (top) are visualized with input hand conditioning overlaid. Note that, consistent with the pretraining data, input text prompts (below) are augmented with an LLM before being input into the model.}

    \label{fig:t2v}
\end{figure*}

Despite being trained on videos from a controlled studio environment, our model is able to transfer its hand interaction capabilities to diverse scenes unseen in training. To demonstrate ``human-centric" generation beyond HOT3D's controlled hand-object interactions, we conduct text-to-video generation across complex, dynamic scenarios (Fig.~\ref{fig:t2v}).

\section{User Study Details}\label{sec:userstudy_appendix}

\begin{figure*}
    \centering

    \newcommand{\figpath}{figs/user_study}
    \newcommand{\sqimg}[2]{%
      \includegraphics[width=#2,clip,trim=288 0 288 0]{#1}%
    }

    \begin{tabularx}{\linewidth}{r >{\centering\arraybackslash}X >{\centering\arraybackslash}X}
        & \textbf{Baseline} & \textbf{Ours} \\
        \addlinespace[2pt]
    
        \rotatebox[origin=c]{90}{\footnotesize \textbf{Task 1}} &
        \adjustbox{valign=c}{%
          \sqimg{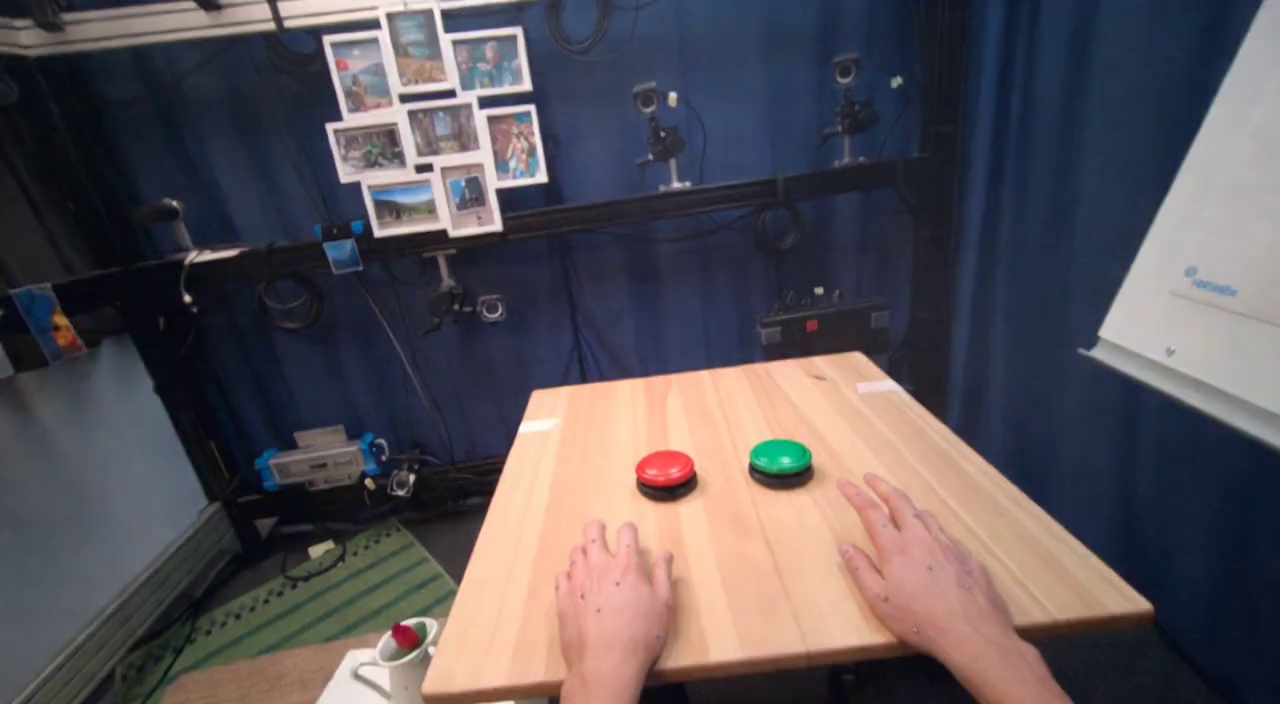}{0.25\linewidth}%
          \sqimg{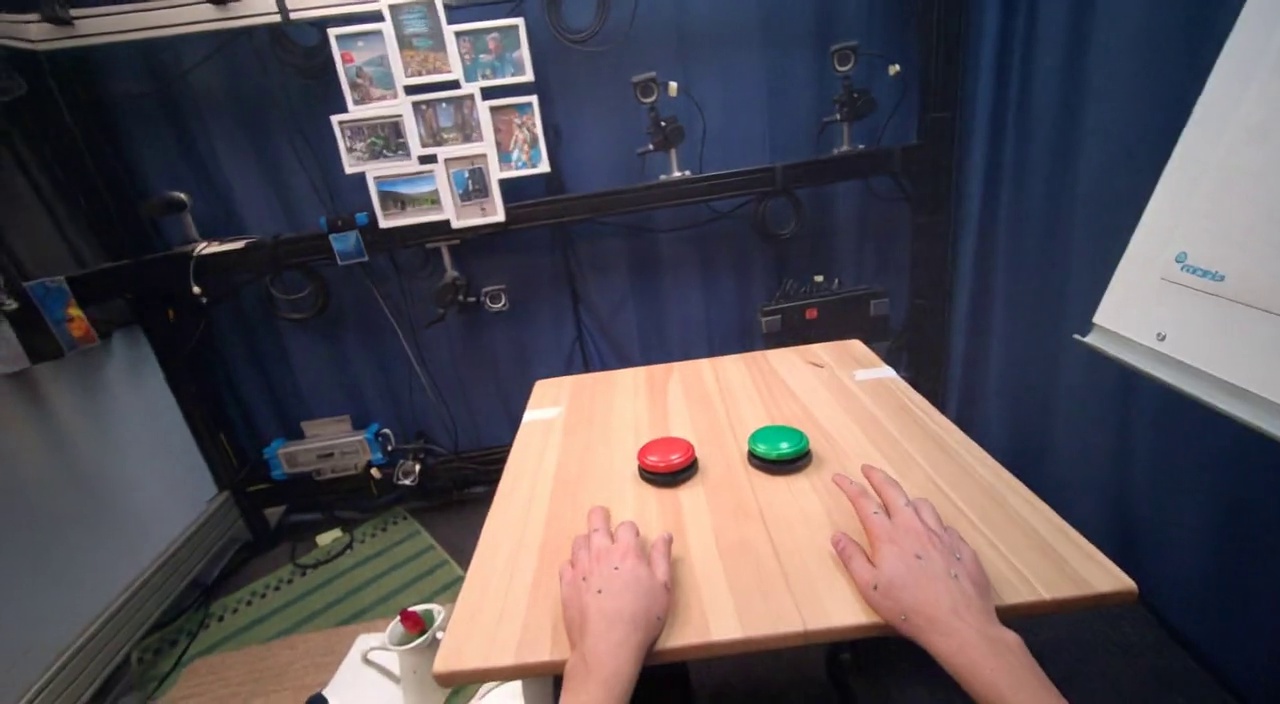}{0.25\linewidth}%
          \sqimg{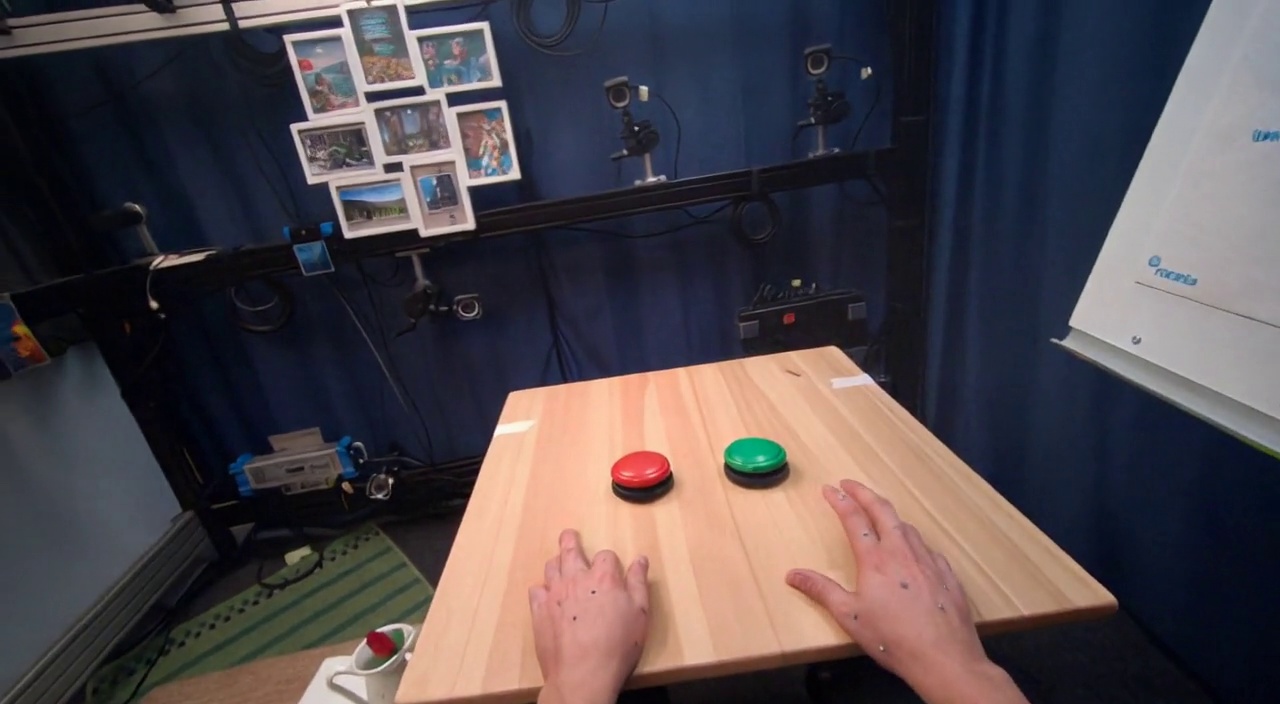}{0.25\linewidth}%
          \sqimg{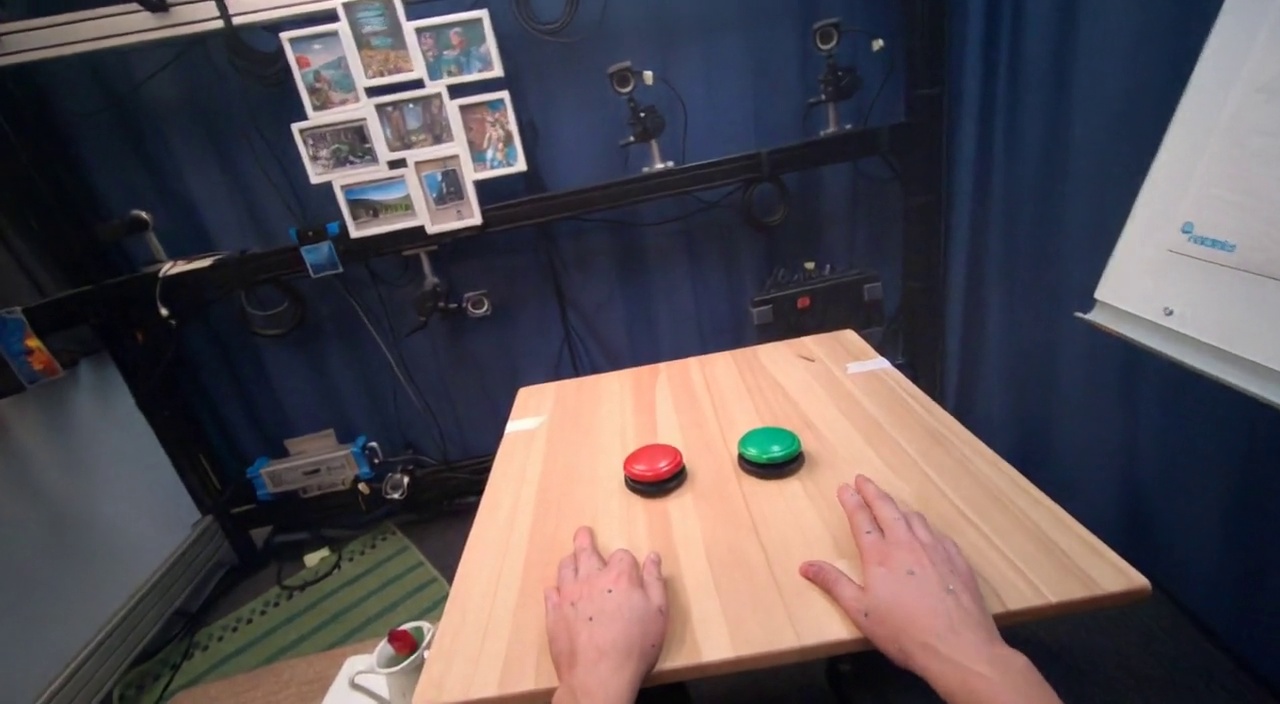}{0.25\linewidth}%
        } &
        \adjustbox{valign=c}{
          \sqimg{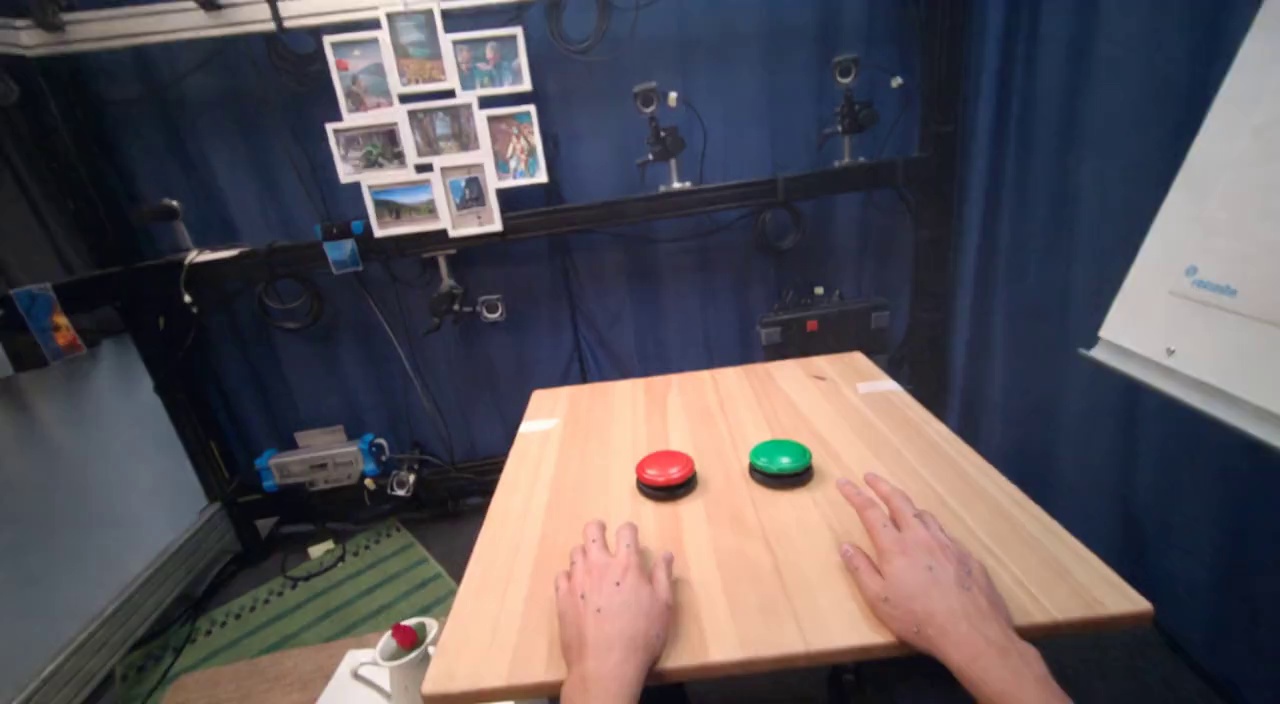}{0.25\linewidth}%
          \sqimg{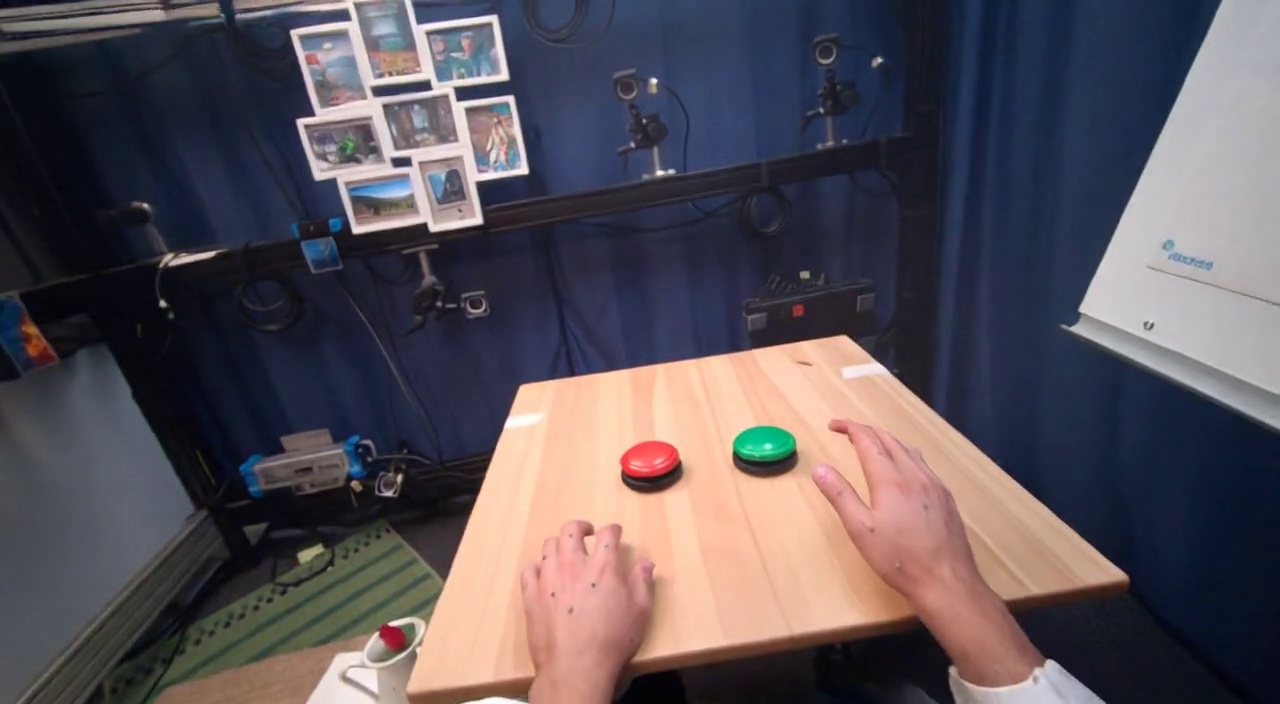}{0.25\linewidth}%
          \sqimg{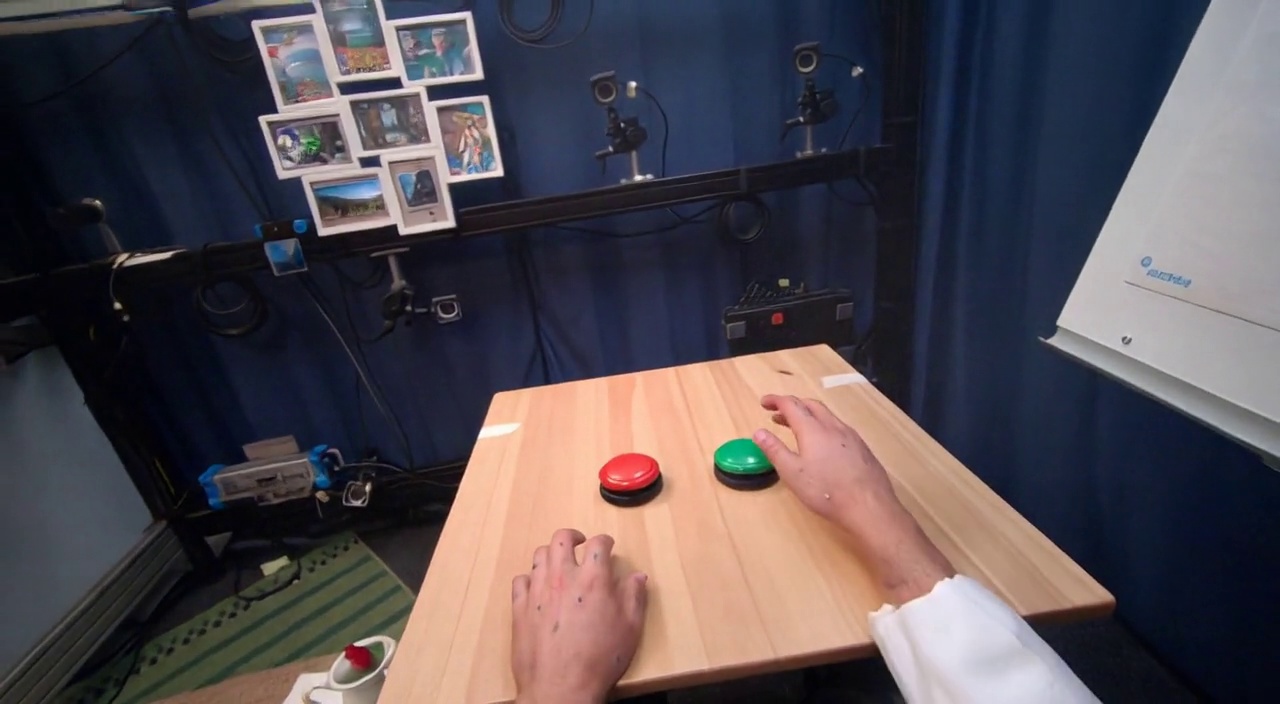}{0.25\linewidth}%
          \sqimg{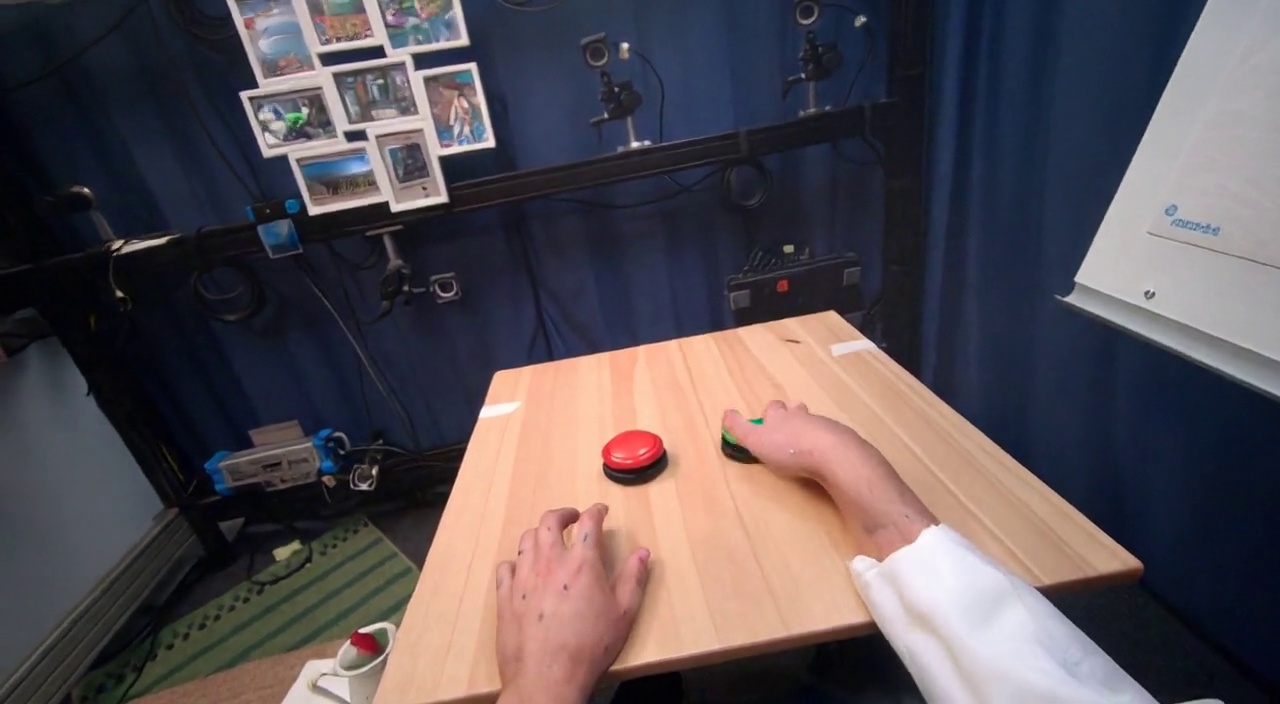}{0.25\linewidth}%
        } \\
        \addlinespace[4pt]
    
        \rotatebox[origin=c]{90}{\footnotesize \textbf{Task 2}} &
        \adjustbox{valign=c}{%
          \sqimg{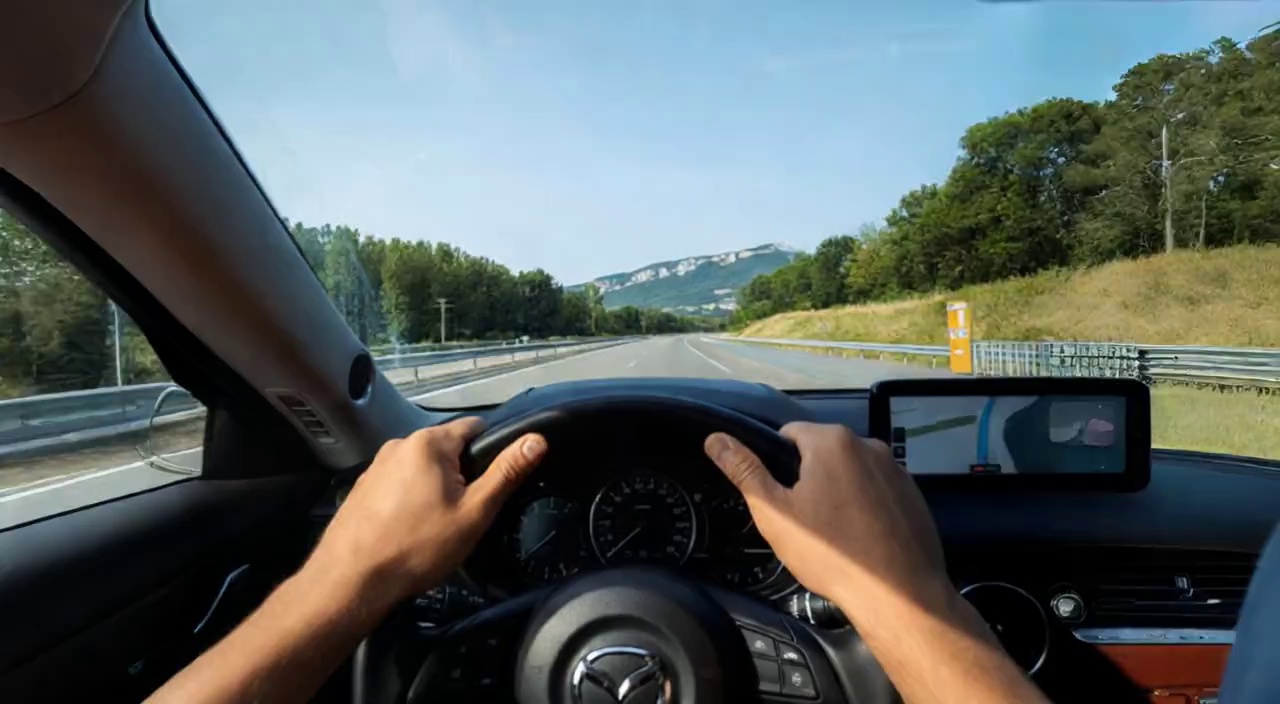}{0.25\linewidth}%
          \sqimg{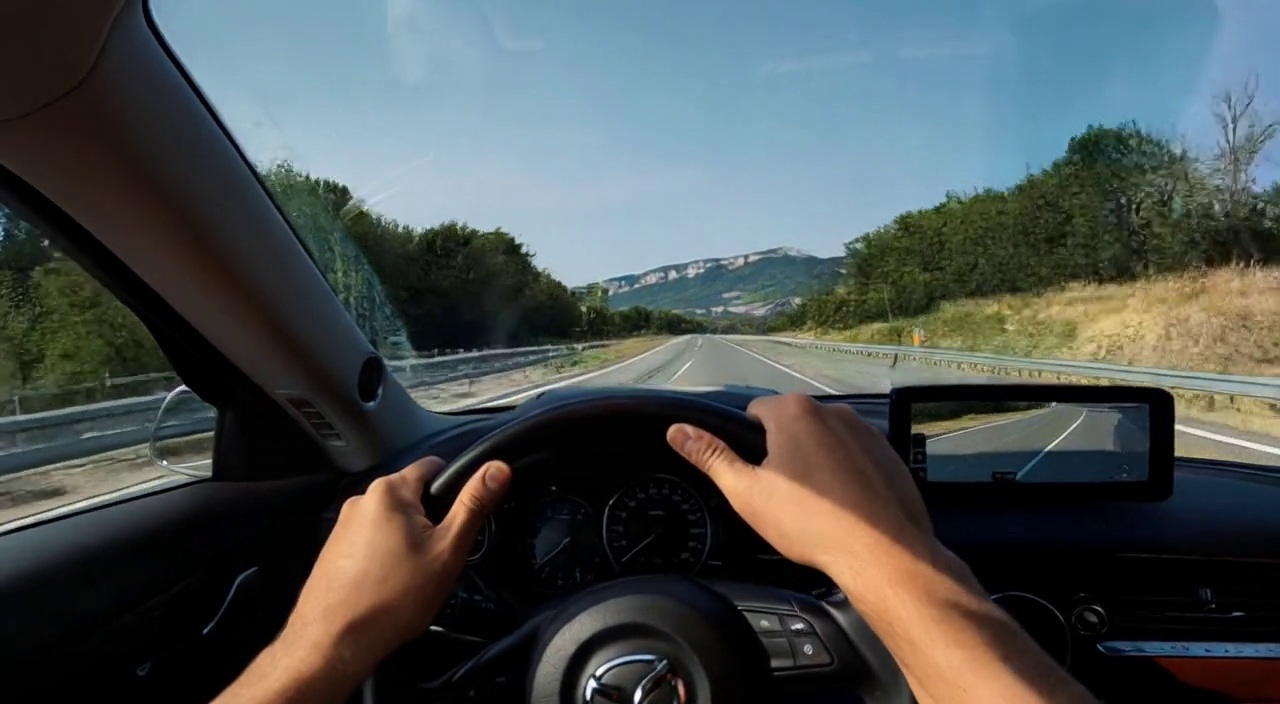}{0.25\linewidth}%
          \sqimg{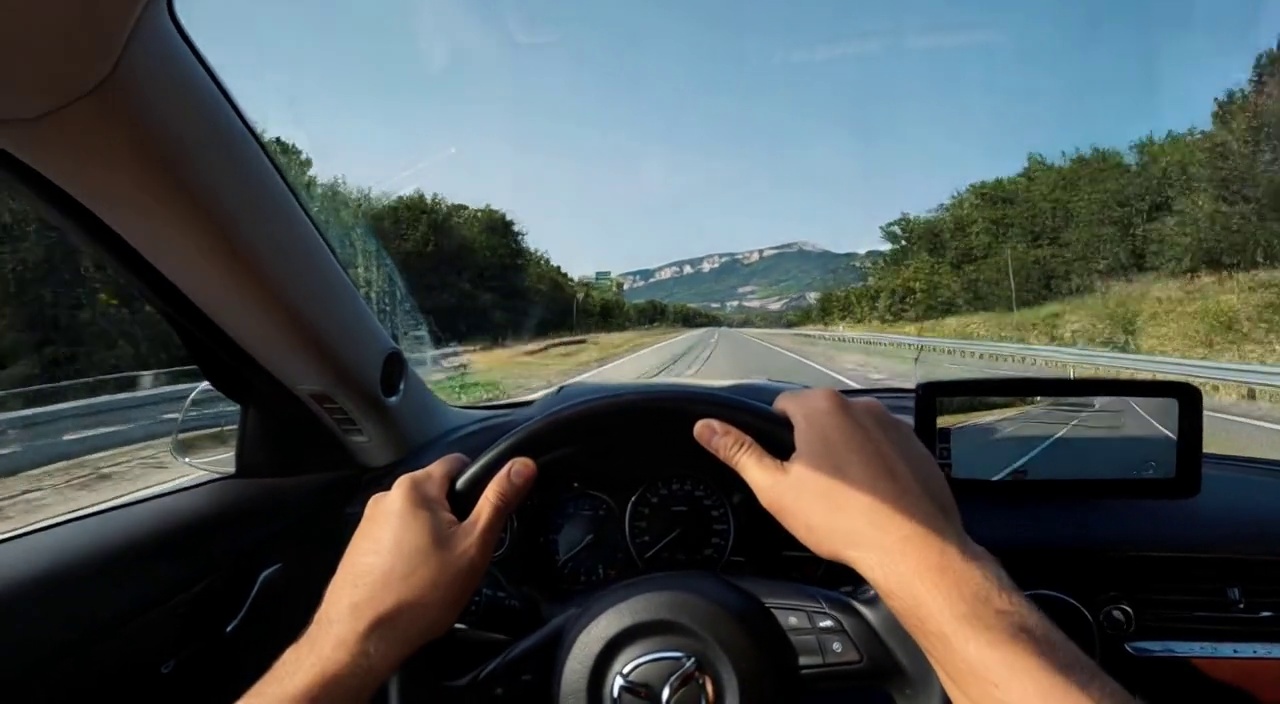}{0.25\linewidth}%
          \sqimg{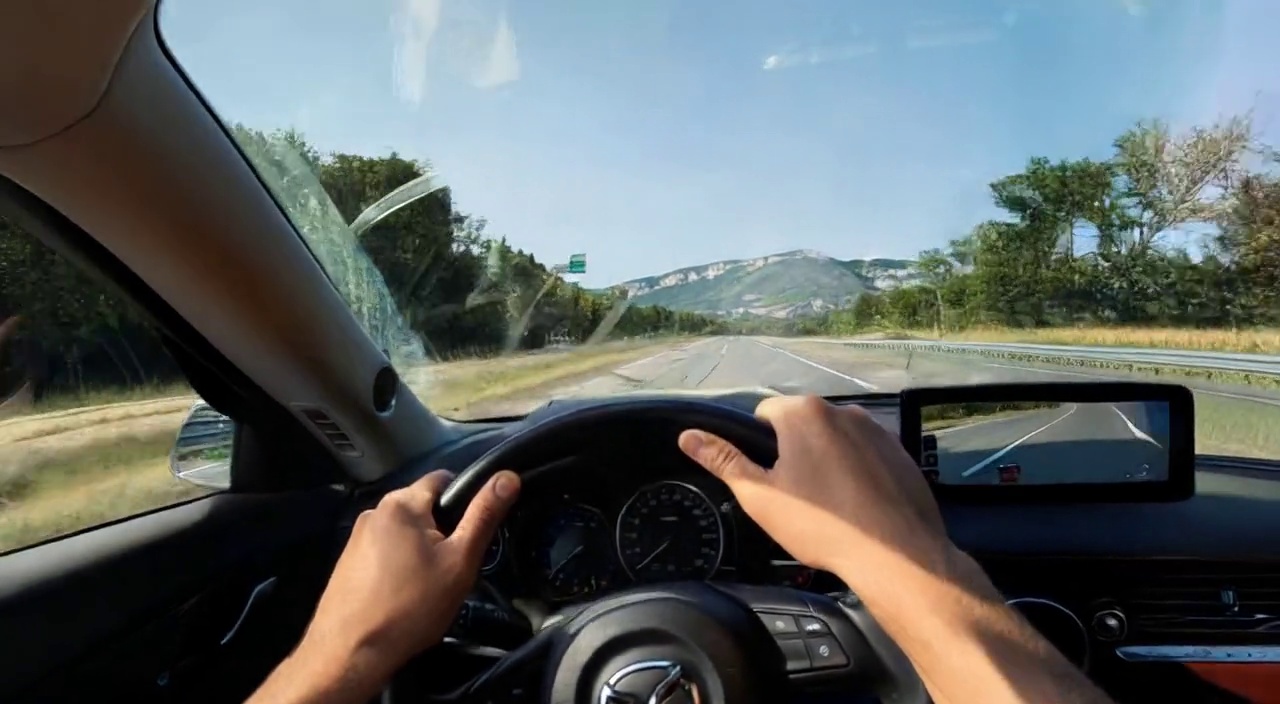}{0.25\linewidth}%
        } &
        \adjustbox{valign=c}{
          \sqimg{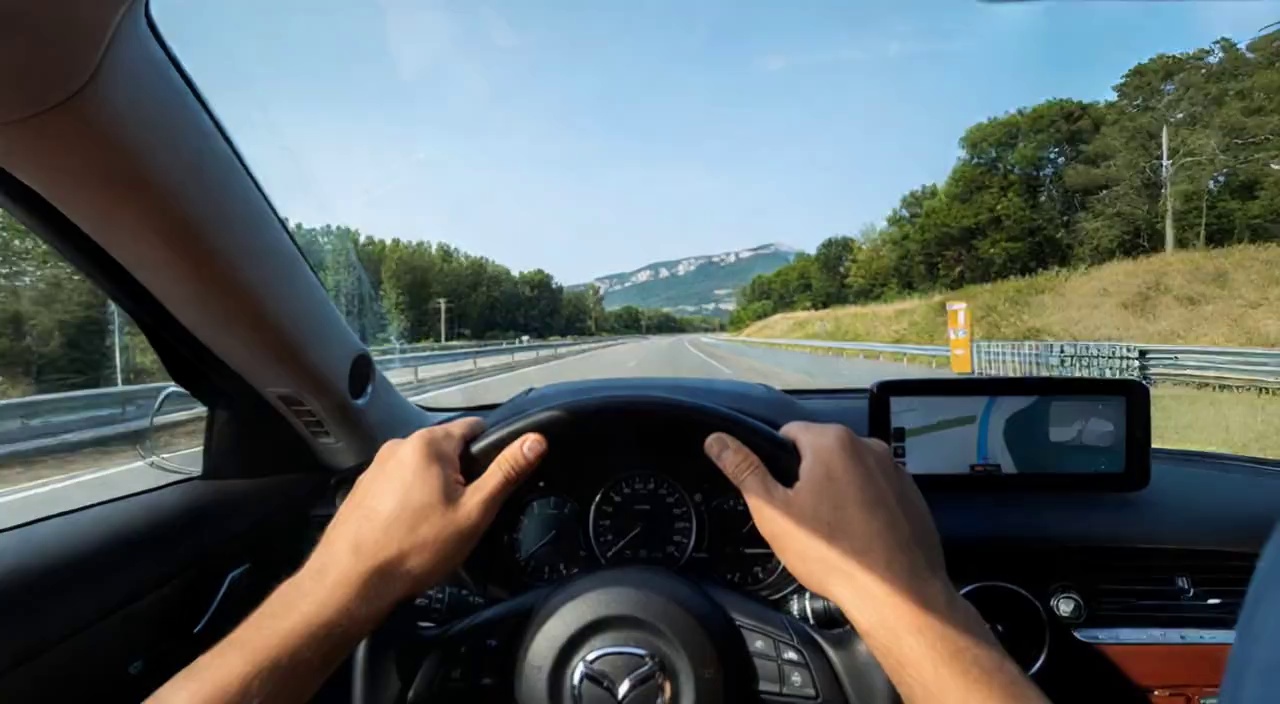}{0.25\linewidth}%
          \sqimg{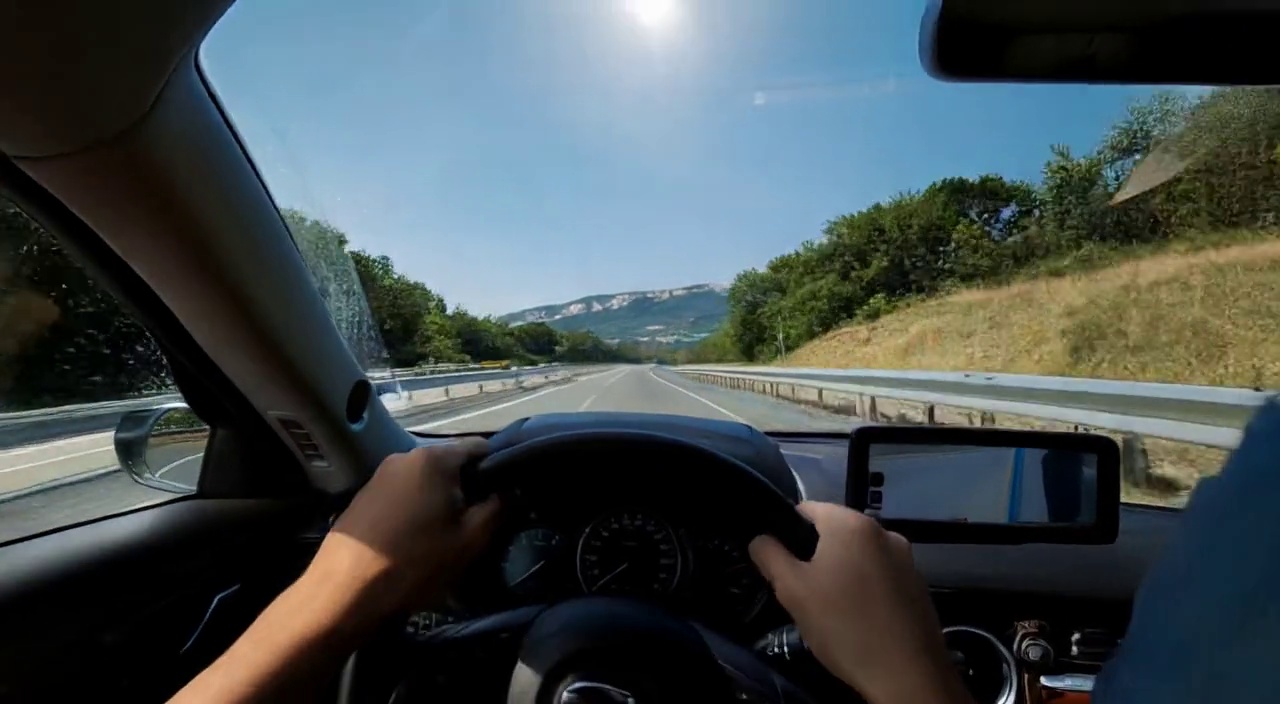}{0.25\linewidth}%
          \sqimg{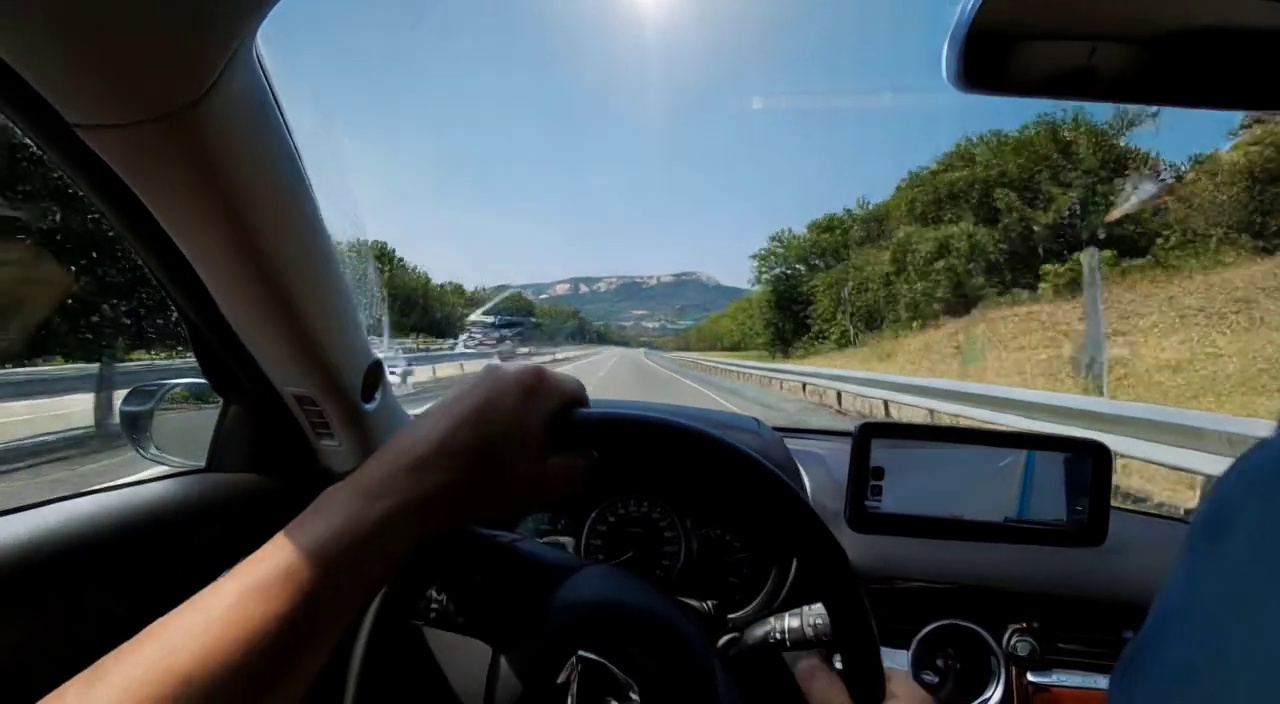}{0.25\linewidth}%
          \sqimg{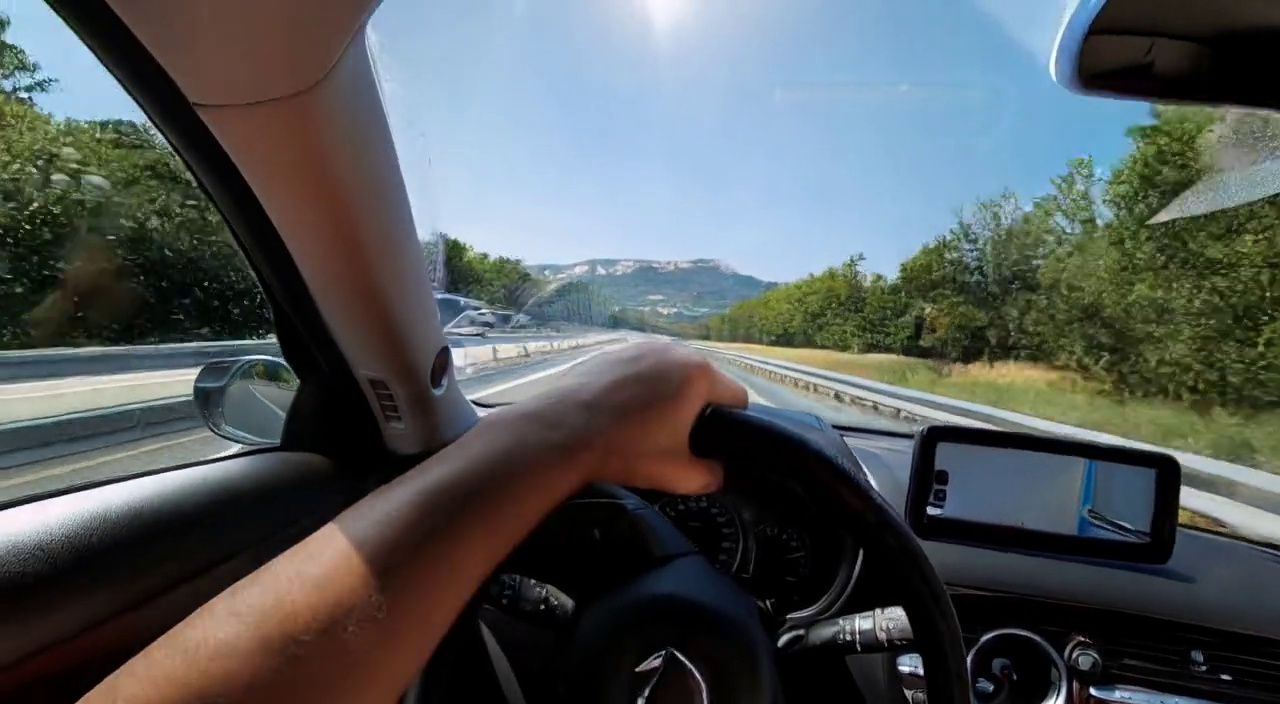}{0.25\linewidth}%
        } \\
        \addlinespace[4pt]
    
        \rotatebox[origin=c]{90}{\footnotesize \textbf{Task 3}} &
        \adjustbox{valign=c}{%
          \sqimg{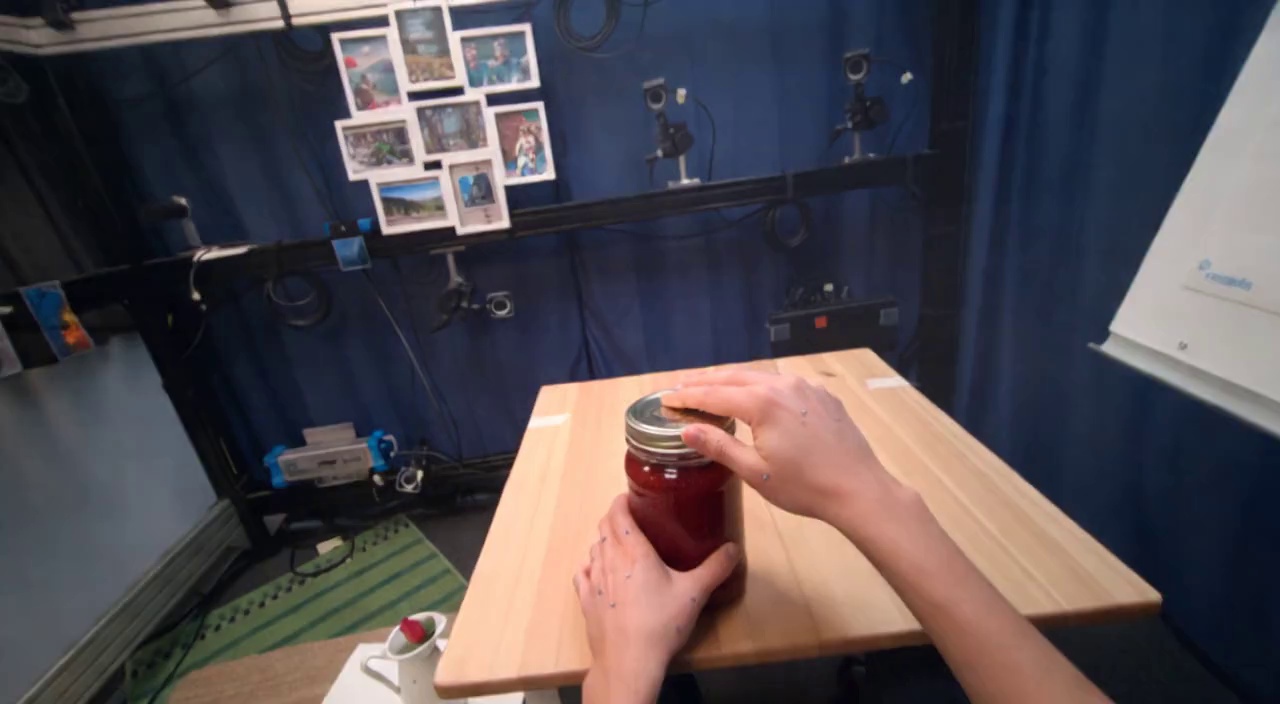}{0.25\linewidth}%
          \sqimg{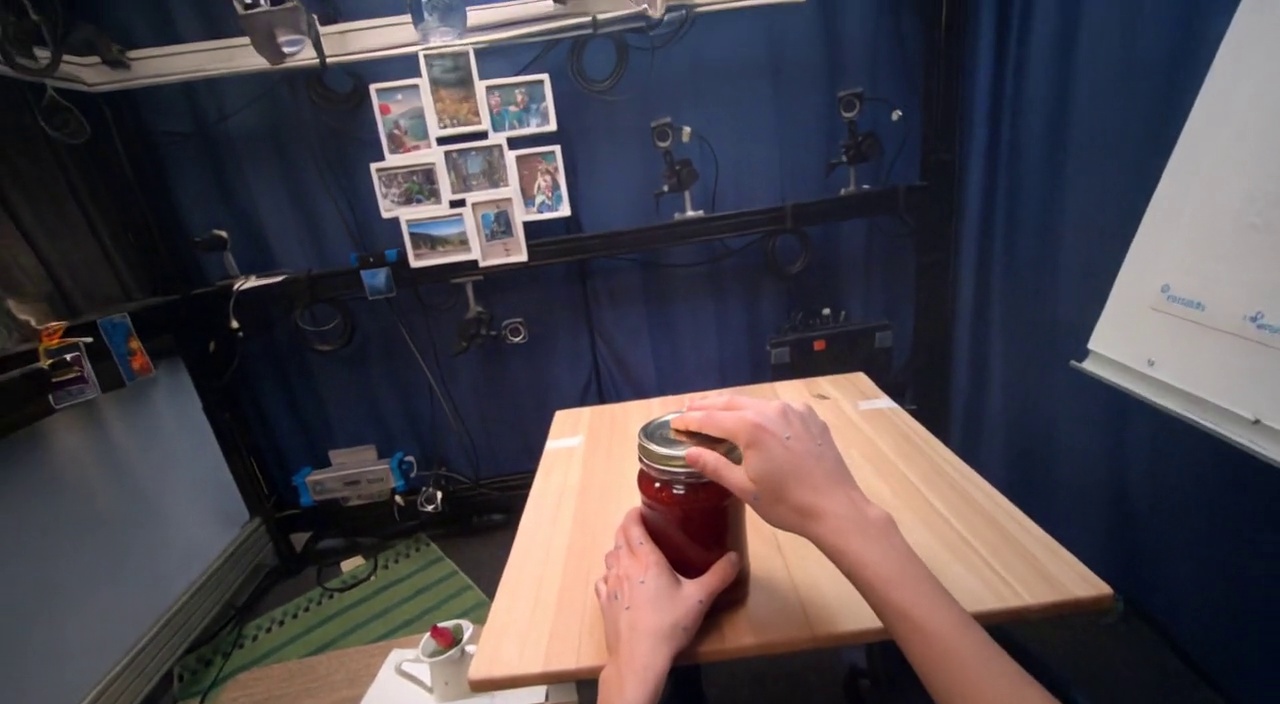}{0.25\linewidth}%
          \sqimg{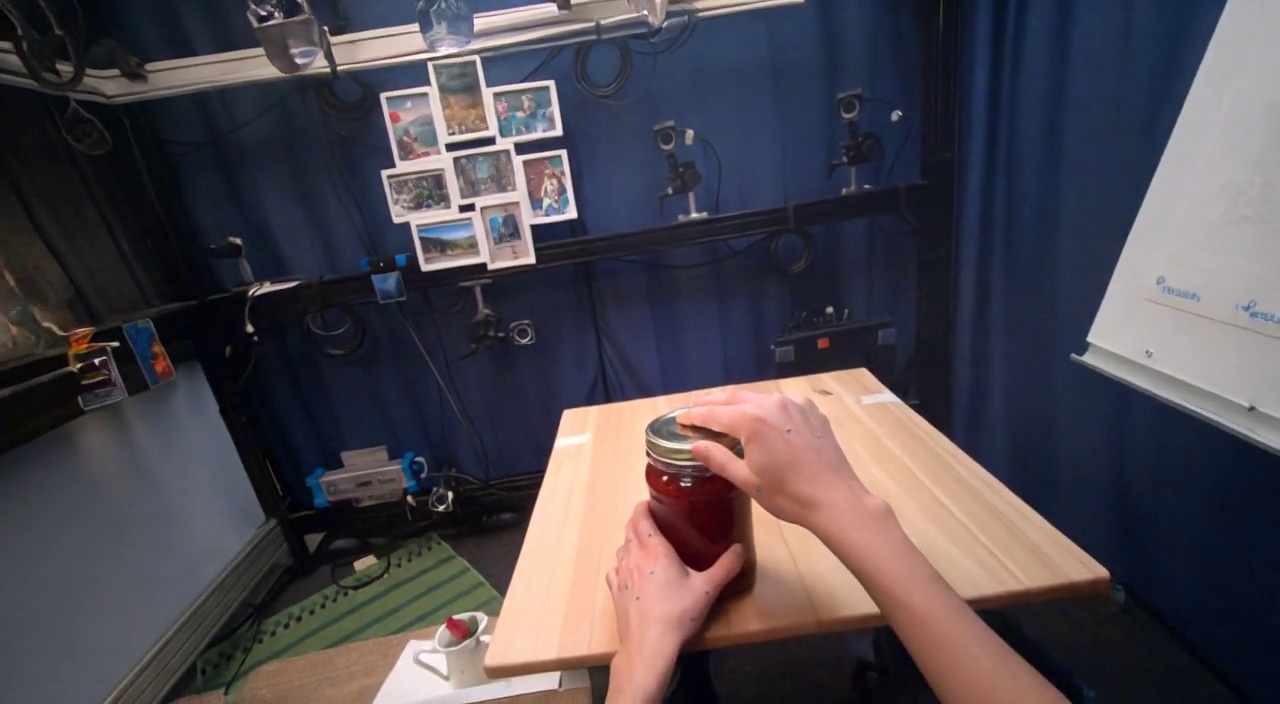}{0.25\linewidth}%
          \sqimg{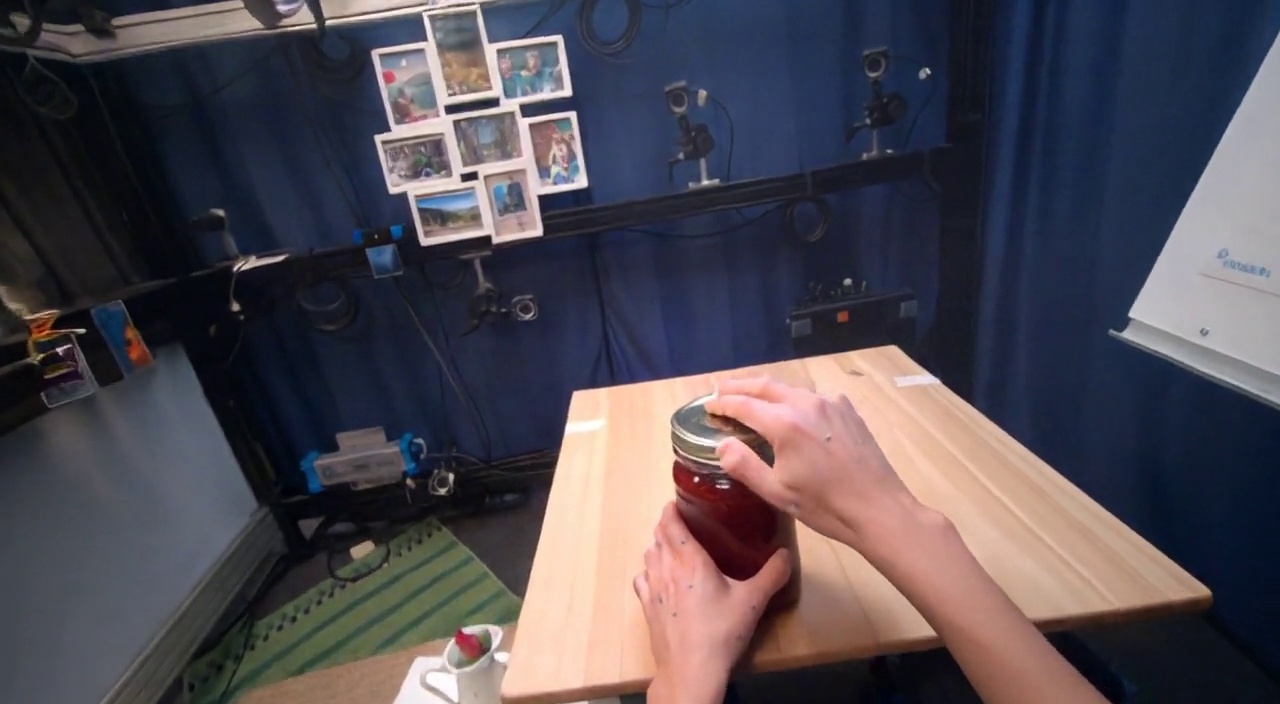}{0.25\linewidth}%
        } &
        \adjustbox{valign=c}{
          \sqimg{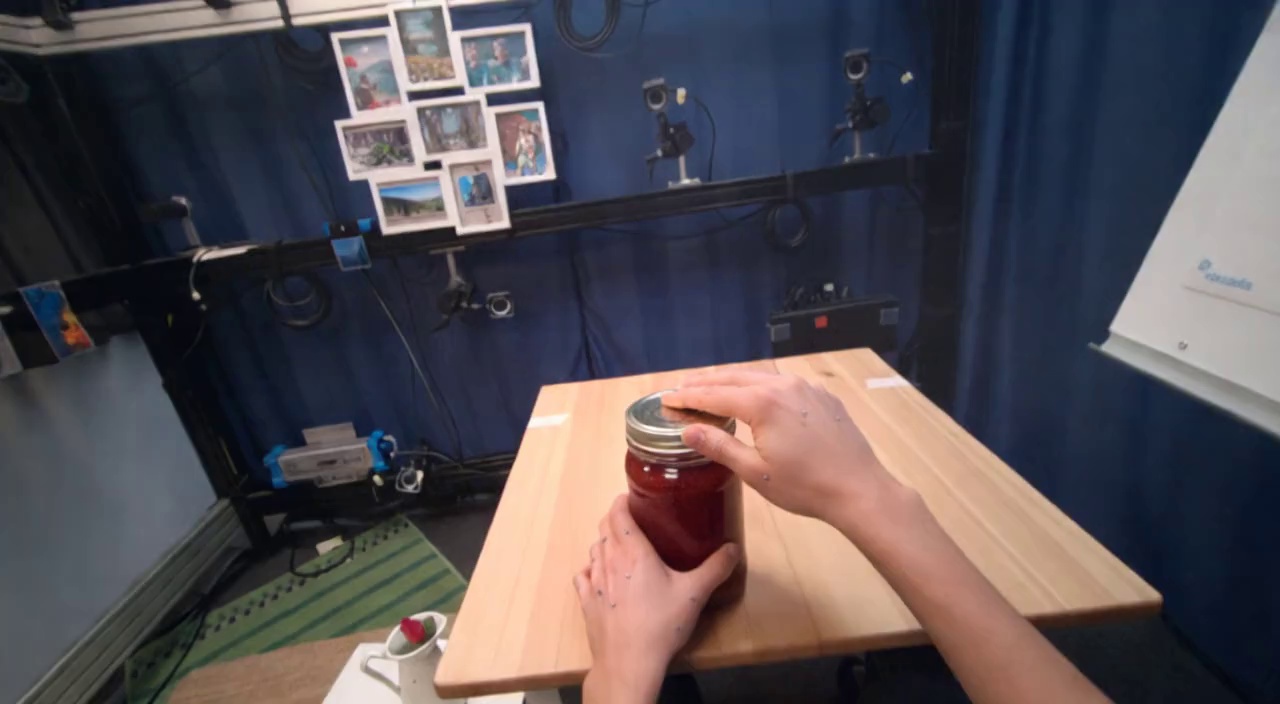}{0.25\linewidth}%
          \sqimg{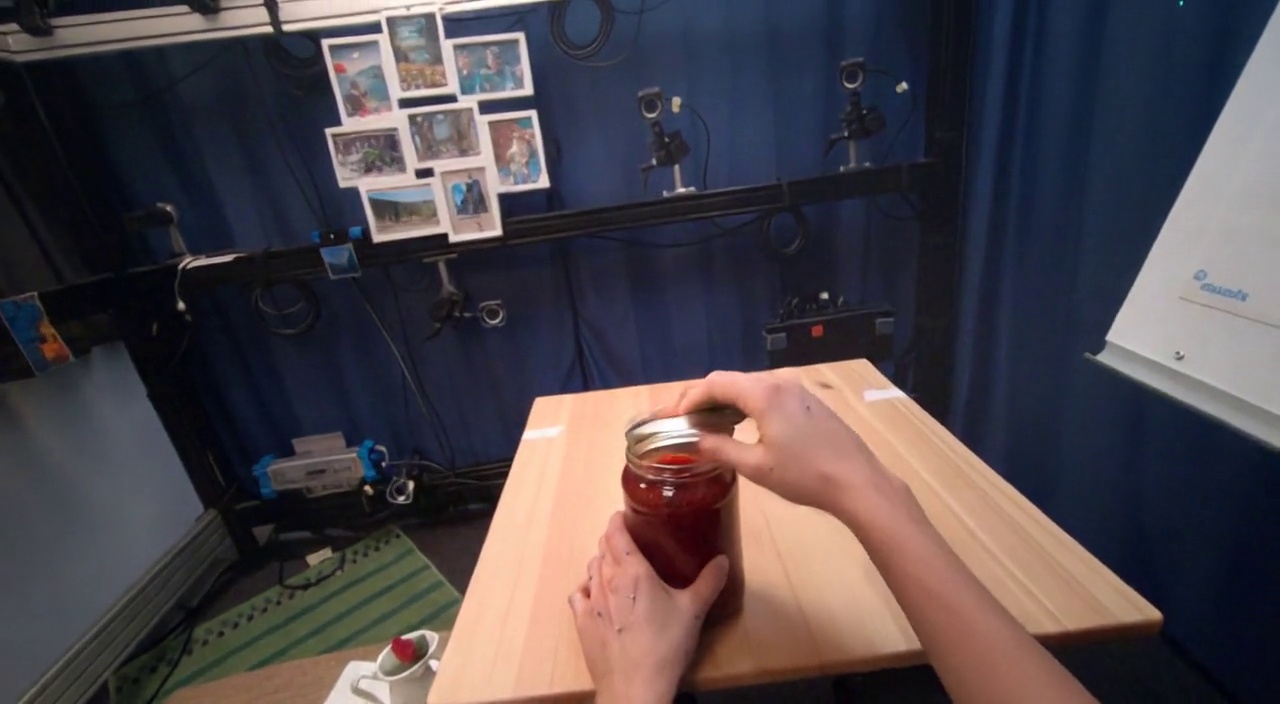}{0.25\linewidth}%
          \sqimg{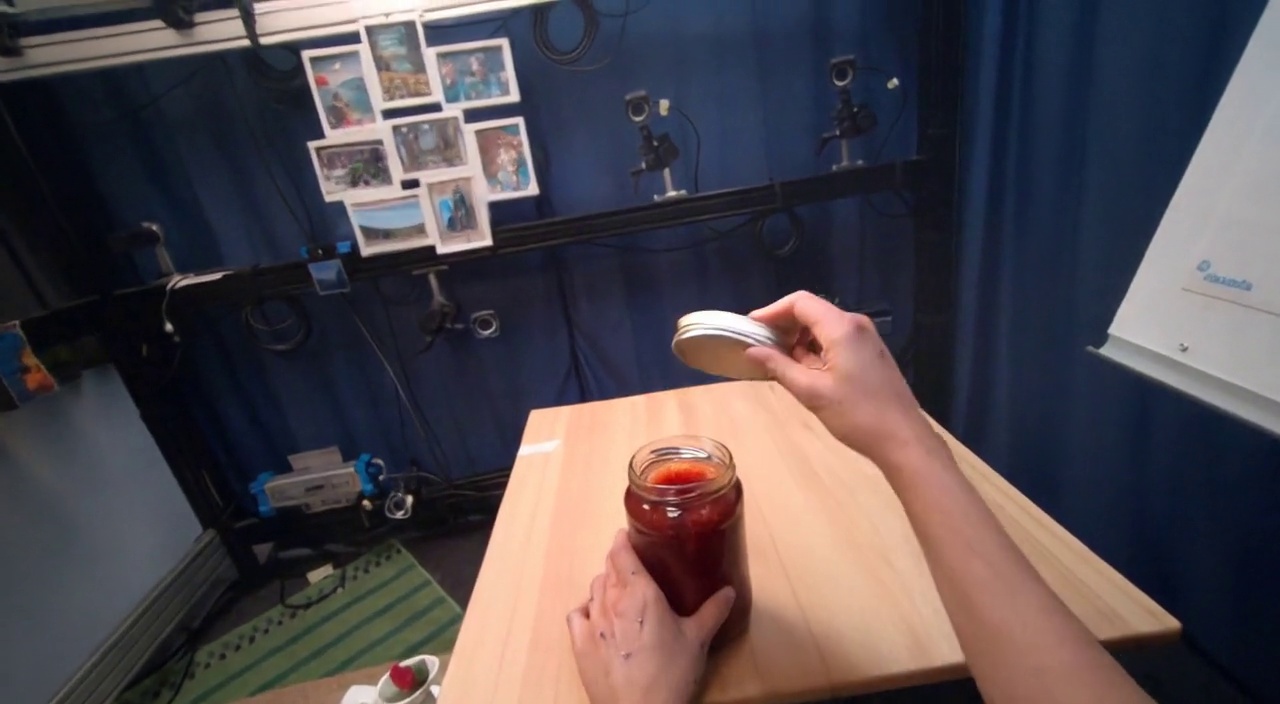}{0.25\linewidth}%
          \sqimg{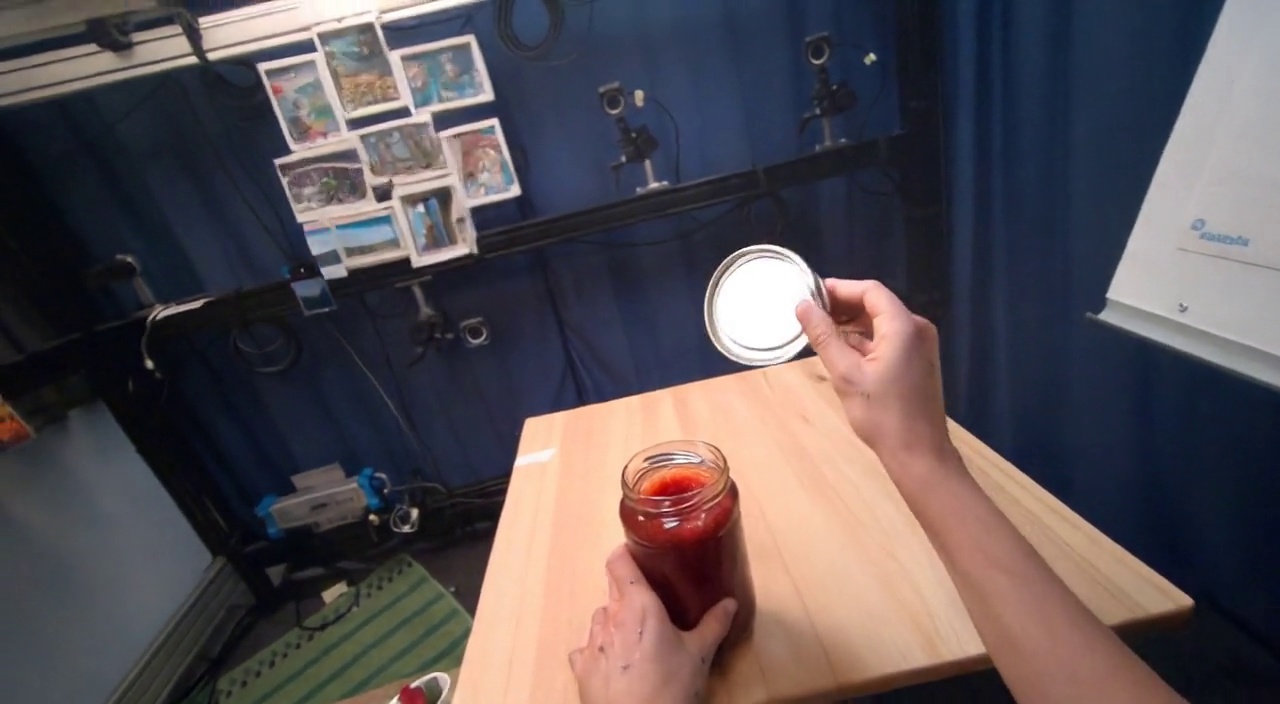}{0.25\linewidth}%
        } \\
    \end{tabularx}

    \caption{\textbf{User Study Qualitative Comparison.} Captured user study results of the baseline vs. our method.}
    \label{fig:userstudyviz}
\end{figure*}

Fig.~\ref{fig:userstudyviz} visualizes comparisons between baseline and our method, captured during the user study. We chose short, simple tasks to enable objective (binary) completion measures and to isolate controllability from generation complexity and long-horizon drift; this also reduces participant discomfort given the current latency.

After each recorded run, participants are asked the question: ``On a scale from 1-7, with 1 being no control and 7 being full control, rate the perceived controllability of the system."
To measure task completion, all generated videos from the session are blind-reviewed by a separate participant for a binary failure/success metric.

\section{Limitations}

While the system models complex hand-object interactions, it struggles with longer-range hand-object-object dependencies. The causal model suffers drawbacks typical of DMD distillation methods, i.e., mode-seeking behavior and over-saturation over long horizons.

We acknowledge that 1.4 second latency is not sufficient for fully immersive XR systems. However, this latency is not fundamental to our approach and can be improved with better hardware, alternative distillation methods, and system optimization (e.g., we communicate with a remote GPU server rather than a local one). Despite this concern, we believe the system to be a practical tool for rapid prototyping and open-ended creation.

\vspace{1cm}

\end{document}